\documentclass[11pt]{article}

\usepackage[utf8]{inputenc}
\usepackage[svgnames,x11names,dvipsnames]{xcolor}
\usepackage{natbib}
\usepackage{wustyle}
\usepackage[a4paper,margin=1in]{geometry}
\usepackage[toc,page,header]{appendix}
\usepackage[small,bf]{caption}
\allowdisplaybreaks

\usepackage[colorlinks,citecolor=blue,urlcolor=blue,linkcolor=blue,
            linktocpage=true,pagebackref=true]{hyperref}
\usepackage{minitoc}
\renewcommand*{\backref}[1]{}
\renewcommand*{\backrefalt}[4]{(\small
    \ifcase #1 not cited%
          \or cited on page~#2%
          \else cited on pages #2%
    \fi
    )}

\usepackage{url}
\usepackage{booktabs}
\usepackage{nicefrac}
\usepackage{enumitem}
\usepackage{needspace}
\usepackage{placeins}
\usepackage{multirow}
\usepackage{wrapfig}
\usepackage[most]{tcolorbox}
\usepackage{framed}
\colorlet{shadecolor}{orange!15}
\definecolor{thistle}{RGB}{216,191,216}
\usepackage{tikz}
\usepackage[textsize=tiny]{todonotes}
\usepackage[none]{hyphenat}
\DeclareRobustCommand{\figref}[2][]{%
  Figure~\hyperref[#2]{\ref*{#2}}%
  \if\relax\detokenize{#1}\relax\else~\textcolor{black}{(#1)}\fi
}
\newcommand{\vtheta}{\bm{\vartheta}}

\newcommand{\blfootnote}[1]{
  \begingroup
  \renewcommand\thefootnote{}
  \footnote{#1}
  \addtocounter{footnote}{-1}
  \endgroup
}

\makeatletter
\DeclareRobustCommand{\fslrulejoin}[3]{
  \makebox[0pt]{\smash{
    \setbox\z@=\hbox{#3}
    \ifdim\ht\z@<\ht\@arstrutbox
      \ht\z@=\ht\@arstrutbox
    \fi
    \ifdim\dp\z@<\dp\@arstrutbox
      \dp\z@=\dp\@arstrutbox
    \fi
    \vrule width\arrayrulewidth
      height\dimexpr\ht\z@+#1\relax
      depth\dimexpr\dp\z@+#2\relax
  }}
}

\makeatother

\title{When and How SignSGD Outperforms SGD: A Functional Scaling Law Perspective}

\author{
  Zilin Wang$^{1,*}$\quad
  Binghui Li$^{1,*}$\quad
  Jia-Nan Wang$^{1,*}$\\[.15em]
  Lean Wang$^{1,*}$\quad
  Jinbo Wang$^{1,*}$\quad
  Lei Wu$^{1,2,\dagger}$\\[.5em]
  $^1$Peking University\quad
  $^2$AI for Science Institute, Beijing\\[.5em]
}
\date{(Accepted at NeurIPS 2026)}

\begin{document}

\maketitle
\blfootnote{
  \begin{tabular}[t]{@{}l@{\,}l@{}}
    $^*$Equal contribution. &
    \texttt{wangzilin@stu.pku.edu.cn}, \texttt{libinghui@pku.edu.cn}, \texttt{jiananwang25@stu.pku.edu.cn},\\
    & \texttt{lean@pku.edu.cn}, \texttt{wangjinbo@stu.pku.edu.cn}.
  \end{tabular}
}
\blfootnote{$^\dagger$Corresponding author. Email: \texttt{leiwu@math.pku.edu.cn}}

\begin{abstract}
    We study how the sign operator interacts with gradient noise to shape the learning dynamics and scaling behavior of sign-based optimizers. In power-law kernel regression with diagonal features, we establish a functional scaling law (FSL) for SignSGD under general learning rate schedules. The FSL describes the learning dynamics through two processes: signal learning and noise forgetting. The key mechanism is noise-induced preconditioning: gradient noise smooths the sign operator, inducing an effective preconditioned SGD. Noise-curvature alignment enables this preconditioning to transform the effective Hessian from $\bH$ to $\bH^{1/2}$. The resulting dynamics exhibit faster signal learning but slower noise forgetting than SGD. This tradeoff yields a regime-dependent comparison with SGD: (i) SignSGD achieves improved data efficiency in the hard-task regime; (ii) both optimizers achieve the same data efficiency in the easy-task regime, but SignSGD maintains this efficiency at larger batch sizes, requiring fewer optimization steps for a fixed data budget. Power-law kernel regression experiments support the proposed mechanism and predicted scaling behavior. LLM pre-training experiments further suggest that the SignSGD FSL better captures the batch-size dependence of loss dynamics for sign-based optimizers than the original SGD FSL.
\end{abstract}

\suppressfloats[t]
\begin{figure}[t]
\centering
\captionsetup{skip=3pt}
\includegraphics[
  width=0.32\textwidth,trim=0 12bp 0 0,clip
]{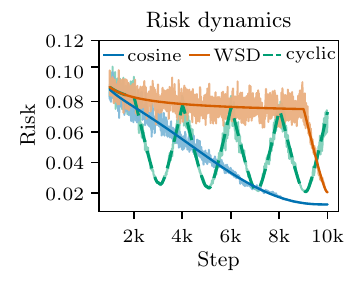}
\hfill
\includegraphics[
  width=0.32\textwidth,trim=0 12bp 0 0,clip
]{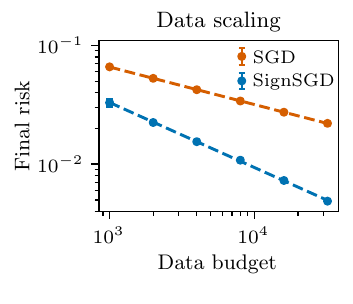}
\hfill
\includegraphics[
  width=0.32\textwidth,trim=0 12bp 0 0,clip
]{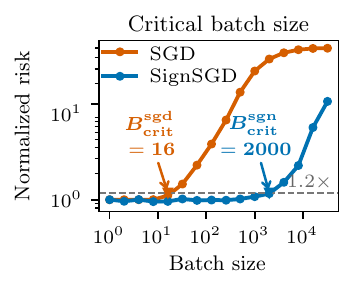}

\caption{\small
\textbf{SignSGD FSL accurately captures risk dynamics and predicts scaling behavior.}
\textbf{(Left)} FSL predictions track SignSGD risk trajectories under cosine, WSD, and cyclic LR schedules.
\textbf{(Middle)} In the hard-task regime, SignSGD achieves
a larger data-scaling exponent than SGD,
$\gamma_{\sgn}>\gamma_{\sgd}$ in $\cE(D)\eqsim D^{-\gamma}$,
indicating improved data efficiency.
Dashed lines denote power-law fits.
\textbf{(Right)} In the easy-task regime, SignSGD improves
step efficiency through a larger CBS, requiring fewer
optimization steps at a fixed data budget while preserving
small-batch performance.
Risk is normalized by each optimizer's $B=1$ baseline.
Marked CBS estimates use a $20\%$ risk-increase tolerance.}
\label{fig:story-overview}
\end{figure}

\doparttoc
\faketableofcontents

\section{Introduction}

Sign-based optimizers play an increasingly prominent role
in modern large-scale training.
SignSGD~\citep{bernstein2018signsgd} and
Lion~\citep{chen2023symbolic} apply the sign operator coordinate-wise,
while Muon~\citep{jordan2024muon,liu2025muon}
uses the matrix sign operator.
Related sign-like behavior also arises in
RMSprop~\citep{hinton2012neural} and
Adam~\citep{kingma2014adam,loshchilov2018decoupled}
through coordinate-wise second-moment
normalization~\citep{balles2018dissecting}.
Together, these methods motivate understanding the sign
operator as a basic mechanism in modern optimization.

Existing theoretical analyses of these sign-based optimizers have
largely focused on deterministic full-batch dynamics~\citep{kunstner2023noise,kunstner2024heavy,zhang2024transformers,kunstner2025scaling,li2026muon} or
convergence guarantees for convex and nonconvex objectives
under general smoothness assumptions~\citep{reddi2019convergence,chen2018convergence,
defossez2020simple,sun2023momentum,shen2025convergence}.
These analyses establish important optimization properties, but do not fully explain how the sign operator interacts with gradient noise or adequately quantify how this interaction shapes loss dynamics and scaling behavior.

In this work, we study SignSGD
in power-law kernel (PLK) regression with diagonal features.
We extend the functional scaling law (FSL) framework
of~\citet{li2025functional} to characterize how sign operator reshapes signal learning and noise forgetting
under general learning rate (LR) schedules.
Our analysis shows that SignSGD acts as an effective
preconditioned SGD method, with the preconditioner determined
by the structure of gradient noise.
Diagonal features and noise-curvature alignment are important in this square-root spectral preconditioning,
accelerating signal learning in low-curvature directions.
The accompanying change in update noise, however,
leads to slower noise forgetting.
The FSL captures these effects by expressing
the risk at step $k$ as

\[
\EE[\cE(\btheta_k)] \eqsim
S(t_k)
+
\sum_{j=0}^{k-1}K(t_k-t_j)\eta_j^2.
\]

Here $\eta_j$ is the learning rate at step $j$, and
$t_k=\sqrt B\sum_{j=0}^{k-1}\eta_j$ is the
\textbf{intrinsic time} of SignSGD with batch size $B$. Signal-learning function $S(t_k)$ describes how the target signal is learned,
and noise-forgetting kernel $K(t_k-t_j)$ measures the remaining effect of noise injected at step $j$. $S(\cdot)$ and $K(\cdot)$ follow the power laws:
$S(t) \asymp t^{-2s}, K(t) \asymp t^{-(2-2/q)},$
where $s>0$ and $q>1$ are task-dependent parameters. \figref[left]{fig:story-overview} shows that this FSL captures the risk dynamics of SignSGD. Leveraging this FSL framework, we furthermore identify two advantages of SignSGD over SGD:
\begin{itemize}[leftmargin=2em]
\item \textbf{Improved data efficiency} (Section~\ref{subsec:data-scaling}). In the hard-task regime, SignSGD achieves a faster data-scaling rate than SGD, as illustrated in \figref[middle]{fig:story-overview}.
\item \textbf{Improved step efficiency}
(Section~\ref{subsec:critical-batch-size}).
In the easy-task regime, SignSGD supports a larger critical
batch size (CBS) than SGD while preserving optimal data
efficiency, requiring fewer optimization steps for a fixed
data budget, as illustrated in
\figref[right]{fig:story-overview}.
\end{itemize}
\paragraph{Empirical validation.}
PLK regression experiments (Section~\ref{subsec:synthetic}) corroborate this
mechanism and the resulting scaling predictions.
In large language model (LLM) pre-training (Section~\ref{subsec:llm_pretrain}), the SignSGD FSL captures the dependence of sign-based optimizers' loss trajectories on batch size more accurately than the original SGD FSL in \citet{li2025functional}.

\paragraph{Related work.}
\citet{xiaoexact} and \citet{compagnoniadaptive} develop SDE descriptions of signSGD and analyze how the sign operator modifies the drift and noise, revealing effects on preconditioning, risk dynamics, and convergence. More directly related to scaling laws, \citet{kim2026scaling} study signSGD in PLK regression, primarily in the noiseless setting. Our work focuses on the label-noise-dominated regime, establishing an explicit FSL that characterizes signal learning and noise forgetting under general LR schedules, which makes the batch-size dependence explicit and enables comparisons of data and step efficiency between signSGD and SGD. \citet{paquette2026phases} study Muon and SignSGD in matrix-valued least squares and show that these sign-based optimizers can provide square-root spectral preconditioning effects. Additional related work is discussed in Appendix~\ref{sec:related_work}.

\section{Preliminaries}
\label{sec:preliminaries}

{\bf Notation.} We use $\eqsim$ to denote equivalence up to a constant factor, and $\lesssim$ (resp.~$\gtrsim$) to indicate inequality up to a constant factor. For two non-negative functions $f,g$, we write $f(x)\lesssim g(x)$ (resp. $f(x)\gtrsim g(x)$) if there exists a constant $C>0$ such that $f(x)\leq C g(x)$ (resp. $g(x)\leq C f(x)$) for all $x$ under consideration. We write $f(x)\eqsim g(x)$ if both $f(x)\lesssim g(x)$ and $f(x)\gtrsim g(x)$ hold.

\subsection{Problem setup}
\paragraph*{Power-law kernel (PLK) regression with diagonalized features.}
Consider a kernel regression task in which labels are generated by
$y=\langle\bphi(\bx),\btheta^*\rangle+\epsilon$,
where data input $\bx\sim\cD$ and label noise
$\epsilon\sim\cN(0,\sigma^2)$ are independent.
Here $\bphi:\cX\to\RR^d$ is the feature mapping and
$\btheta^*\in\RR^d$ is the unknown target parameter.
We consider the infinite-dimensional setting, i.e., $d=\infty$.
We assume Gaussian features $\bphi(\bx)\sim\cN(0,\bH)$, where
$\bH=\EE_{\bx\sim\cD}[\bphi(\bx)\bphi(\bx)^\top]$
is the feature covariance.
The prediction model $f(\bx;\btheta)=\langle\bphi(\bx),\btheta\rangle$
is trained by minimizing the population risk
$\cR(\btheta)=\EE_{\bx,y}[\ell(\btheta;\bx,y)]$
under the squared loss
$\ell(\btheta;\bx,y)=\frac12(f(\bx;\btheta)-y)^2$.
The population risk decomposes as
$\cR(\btheta)=\cE(\btheta)+\sigma^2/2$, where
$\cE(\btheta)=\frac12(\btheta-\btheta^*)^\top
\bH(\btheta-\btheta^*)$
is the excess risk.

\begin{assumption}[Constant label noise] \label{ass:label noise} Assume $\sigma\eqsim1$.
\end{assumption}
\begin{assumption}[Diagonal features] \label{ass:diagonal covariance} Assume $
\bH = \diag(\lambda_1,\cdots,\lambda_d).$

\end{assumption}
\begin{assumption}[Power-law structures] \label{ass:power-law}
    For some $q>1$ and $s>0$, $
\lambda_j \eqsim j^{-q}, \lambda_j |\theta^*_j|^2\eqsim j^{-(1+sq)}.
$
\end{assumption}

Assumption~\ref{ass:power-law} specifies the power-law source and capacity structure of the regression. Here $q$ is the \textbf{spectral-decay exponent}, which controls the decay rate of the feature spectrum, and $s$ is the \textbf{source exponent}, which describes how the target energy is distributed across spectral directions.
\paragraph{One-pass SignSGD.}
At each training step $k=0,\ldots,N-1$, we draw a fresh mini-batch
$\cB_k=\{(\bx_{k,i},y_{k,i})\}_{i=1}^B$ of size $B$,
with all samples drawn i.i.d.\ from the data distribution.
Starting from $\btheta_0=\bzero$, one-pass SignSGD performs the updates

\begin{equation}
\label{eq:signSGD-update}
\bg_k=\frac{1}{B}\sum_{i=1}^B
\nabla_{\btheta}\ell(\btheta_k;\bx_{k,i},y_{k,i}),
\qquad
\btheta_{k+1}=\btheta_k-\eta_k\sgn(\bg_k),
\end{equation}

where $\sgn(\cdot)$ acts coordinate-wise and
$\{\eta_k\}_{k=0}^{N-1}$ is the LR schedule.

\subsection{An SDE modeling for SignSGD}

We extend the intrinsic-time SDE developed for SGD
by \citet{li2025functional} to SignSGD.
\paragraph{Modeling gradient noise.}
Write $\bg_k=\bmu_k+\bxi_k$, where $\bmu_k:=\nabla\cR(\btheta_k)$ is the population gradient and $\bxi_k$ is the gradient noise satisfying $\EE[\bxi_k]=\bzero$. We model $\bxi_k\sim\cN(\bzero,\sigma_B^2\bSigma)$, where $\bSigma$ is a single-sample covariance and $\sigma_B=1/\sqrt{B}$ characterizes the dependence on batch size.  An extension beyond Gaussian noise is discussed in Appendix~\ref{app:non-gaussian-noise}.
\paragraph{Smoothed sign and update noise.}
For batch size $B$, define the scaled smoothed sign operator

\[
\bmm_B(\bx):=\sigma_B
\EE_{\bxi\sim\cN(\bzero,\bSigma)}
[\sgn(\bx+\sigma_B\bxi)]
\]

Since $\sgn(a\bx)=\sgn(\bx)$ for $a>0$, the Gaussian noise model gives $\EE[\sgn(\bg_k)] =\bmm_B(\bmu_k)/\sigma_B$. Define the centered sign noise $\bn_k:=\sgn(\bg_k)-\bmm_B(\bmu_k)/\sigma_B$ and its covariance $\bQ_B(\btheta_k):=\EE[\bn_k\bn_k^\top]$. The update then becomes $\btheta_{k+1}-\btheta_k =-\eta_k[\bmm_B(\bmu_k)/\sigma_B+\bn_k]$. Its conditional mean and covariance are $-\eta_k\bmm_B(\bmu_k)/\sigma_B$ and $\eta_k^2\bQ_B(\btheta_k)$, respectively, which specify the drift and diffusion covariance of the effective continuum model.

\paragraph{Intrinsic-time dynamics.}
To analyze the effects of the LR schedule and batch size, we introduce two key quantities: \textbf{intrinsic time} and \textbf{noise intensity}. Let $z$ denote continuous training steps and $\eta(z)$ a continuous LR schedule satisfying $\eta(k)=\eta_k$. We define the intrinsic time of SignSGD as
$
t=\Gamma_B(z):=\int_0^z\frac{\eta(u)}{\sigma_B}\,\dd u,
$
which is the continuous analogue of $\sigma_B^{-1}\sum_{j=0}^{k-1}\eta_j$. The corresponding noise intensity is defined by
$
n_B(t):=\sigma_B\eta(\Gamma_B^{-1}(t)),
$
which measures the noise strength injected per unit intrinsic time. With these two quantities, we show that SignSGD can be modeled by the SDE
\begin{equation}
\label{eq:sde-sign}
\dd\vtheta_t
=-\bmm_B(\nabla\cR(\vtheta_t))\,\dd t
+\sqrt{n_B(t)}\,\bQ_B^{1/2}(\vtheta_t)\,\dd\bW_t,
\end{equation}
where $\vtheta_t$ denotes the continuum parameter process with $\vtheta_0=\bzero$, and $\bW_t$ is a standard Brownian motion. The function $n_B(t)$ controls the noise injection per unit intrinsic time, with instantaneous diffusion covariance $n_B(t)\bQ_B(\vtheta_t)$. \citet{compagnoniadaptive} derive a related SDE for constant learning rates. Our formulation incorporates a learning rate schedule and the intrinsic-time transformation $t=\Gamma_B(z)$. Appendix~\ref{app:sde-sign} provides the derivation and the explicit coordinate-wise dynamics for the PLK regression model.

\begin{remark}[Comparison with SGD]
The intrinsic time of SGD is $t_{\sgd}(z)=\int_0^z\eta(u)\,\dd u$ \citep{li2025functional,li2026optimal}. For SignSGD, we include an additional factor $1/\sigma_B$ to account for the batch-size dependence of its effective drift. Section~\ref{sec:derivation_mechanism} explains the mechanism underlying this choice.
\end{remark}

\begin{assumption}[Curvature-aligned noise]
\label{ass:curvature-aligned-noise} Assume $\bSigma\eqsim\bH$.
\end{assumption}

Assumption~\ref{ass:curvature-aligned-noise} states that the noise variance along each direction scales with the corresponding curvature, i.e., $\bv^\top\bSigma\bv\eqsim\bv^\top\bH\bv$ for all $\bv$, with comparison constants independent of $\bv$. Such noise--curvature alignment has been studied theoretically and observed empirically in deep learning~\citep{wang2023theoretical}. In our PLK regression setting, the exact single-sample gradient-noise covariance at $\btheta_k$ is comparable to $(\cE(\btheta_k)+\sigma^2)\bH$ (see Appendix~\ref{app:noise structure}). In the label-noise-dominated regime $\cE(\btheta_k)\lesssim\sigma^2\eqsim1$, this covariance is therefore comparable to $\bH$, motivating the assumption in our Gaussian noise model.

\section{Functional scaling law for SignSGD}
\label{sec:functional_scaling_law_for_signsgd}

\begin{tcolorbox}[
    enhanced,
    colback=RoyalBlue4!4,
    colframe=RoyalBlue4!75,
    boxrule=0.9pt,
    arc=1mm,
    left=1.5mm,
    right=1.5mm,
    top=1mm,
    bottom=1mm
]
\begin{theorem}[FSL for SignSGD]
\label{thm:main-fsl-sign}
Suppose Assumptions~\ref{ass:label noise},
\ref{ass:diagonal covariance}, \ref{ass:power-law},
and~\ref{ass:curvature-aligned-noise} hold.
Consider the SDE~\eqref{eq:sde-sign} with
$\sqrt B \sup_z\eta(z)\lesssim1$.
For $t\gtrsim B^{\max\{1/2,\,1/(2s+2/q)\}}$,
it holds that
\begin{equation}
\label{eq:fsl-sign}
\EE[\cE(\vtheta_t)]\eqsim
S(t)+\int_0^t K(t-\tau)n_B(\tau)\,\dd \tau,
\end{equation}
where
\begin{equation}
\label{eq:sign-signal-kernel}
S(t)\asymp t^{-2s},
\qquad
K(t)\asymp t^{-(2-2/q)}.
\end{equation}
Here the hidden constants are independent of $B$ and the LR schedule.
\end{theorem}
\end{tcolorbox}

The proof is provided in Appendix~\ref{app:main proof}. Section~\ref{sec:derivation_mechanism} explains the underlying mechanism and derives the FSL from the effective preconditioned dynamics. The FSL separates the remaining signal error $S(t)$ from the accumulated effect of update noise. Here $n_B(\tau)$ is the noise injection intensity per unit intrinsic time at time $\tau$, while $K(t-\tau)$ measures the residual risk from that injection after elapsed time $t-\tau$. The LR schedule enters through both the intrinsic time and $n_B$, allowing the FSL to describe the risk trajectory under different schedules.

\citet{li2025functional} establish an SGD FSL with the same convolution form as~\eqref{eq:fsl-sign}. For SGD, the intrinsic time, noise intensity and FSL components are
\[
t=\Gamma_{\sgd}(z):=\int_0^z\eta(u)\,\dd u,\quad
n_B(t)=\frac{\eta(\Gamma_{\sgd}^{-1}(t))}{B},\quad
S(t)\asymp t^{-s},\quad
K(t)\asymp t^{-(2-1/q)}.
\]

\paragraph{Signal learning and noise forgetting.}
In their respective intrinsic times, SignSGD has a faster signal decay, $t^{-2s}$ compared with $t^{-s}$ for SGD, but a slower forgetting kernel, $t^{-(2-2/q)}$ compared with $t^{-(2-1/q)}$. Thus, SignSGD learns the target signal faster while retaining the effects of past noise for longer.
\paragraph{The effect of batch size.}
For the same training steps and LR schedule, the intrinsic times satisfy $t_{\sgn}=\sqrt B\,t_{\sgd}$. Unlike SGD, whose intrinsic time is independent of $B$, SignSGD has an essentially longer training time as $B$ increases. A larger batch also allows more forgetting of earlier noise injections. Moreover, the noise intensity scales as $1/\sqrt B$ for SignSGD and $1/B$ for SGD.
\begin{table}[!ht]
\centering
\small
\setlength{\tabcolsep}{5pt}
\renewcommand{\arraystretch}{1.5}
\begin{tabular}{@{}lcccc@{}}
\toprule
\textbf{Optimizer}
& \textbf{Intrinsic time}
& \textbf{Noise intensity}
& $S(t)$
& $K(t)$ \\
\midrule
SGD~\citep{li2025functional}
& $\int_0^z\eta(\tau)\,\dd\tau$
& $1/B$
& $t^{-s}$
& $t^{-(2-1/q)}$ \\
SignSGD (\textbf{ours})
& $\sqrt B\int_0^z\eta(\tau)\,\dd\tau$
& $1/\sqrt{B}$
& $t^{-2s}$
& $t^{-(2-2/q)}$ \\
\bottomrule
\end{tabular}
\end{table}

\section{When and how does SignSGD outperform SGD?}

The FSL enables the comparison of scaling efficiency between SignSGD and SGD.
For a data budget $D=NB$, we characterize the achievable
excess risk and determine how increasing the batch size $B$
can reduce the number of training steps $N$ without
sacrificing data efficiency.
Following~\citet{li2026optimal}, we consider the power-decay
LR schedule
\begin{equation}
\label{eq:power-decay}
\eta(z)
=
\eta_{\peak}\left(1-\frac{z}{N}\right)^p,
\qquad 0\le z\le N,
\end{equation}
where $\eta_{\peak}>0$ and $p>0$ denote the peak learning
rate and decay exponent, respectively.
This schedule family attains the optimal data-scaling rates characterized below.

\subsection{Data efficiency}
\label{subsec:data-scaling}

We first ask: {\it what is the optimal final-step excess risk
that SignSGD can achieve by tuning the LR schedule
at a fixed data budget $D$?}
We consider $B=1$, with the case of $B>1$ deferred to Section \ref{subsec:critical-batch-size}.
For the power-decay schedule~\eqref{eq:power-decay}, the final expected excess risk is $\cE_D:=\EE[\cE(\vtheta_{\Gamma_1(D)})]$, where $\Gamma_1(D)=\int_0^D\eta(\tau)\,\dd\tau$. The following theorem specifies choices of $\eta_{\peak}$ and $p$ and the resulting data-scaling rates.
\begin{theorem}[Data scaling of SignSGD]
  \label{thm:sign-data-scaling}
  \normalfont
  Under the assumptions of Theorem~\ref{thm:main-fsl-sign}, let $B=1$ and use the power-decay schedule~\eqref{eq:power-decay}. Then
  \begin{itemize}[leftmargin=2em]
    \item If $s\ge \tfrac12-\tfrac1q$, then choosing $p>\tfrac{q}{2}-1$ and $\eta_{\peak} \eqsim D^{-\frac{2s-1+2/q}{2s+2/q}}$
    yields
    $
        \cE_D \eqsim D^{-\frac{sq}{sq+1}} .
    $
    \item If $0<s<\tfrac12-\tfrac1q$, then choosing $p>\tfrac{2s}{1-2s}$ and $\eta_{\peak}\eqsim 1$ yields
    $
        \cE_D \eqsim D^{-2s} .
    $
  \end{itemize}
\end{theorem}

\noindent\begin{minipage}[t]{0.60\textwidth}
Its proof is given in Appendix~\ref{app:data-scaling-sign}. We show that the rates achieved in Theorem~\ref{thm:sign-data-scaling} are optimal among all admissible LR schedules (see details in Appendix~\ref{app:optimal-power-decay}). Thus, the optimal data-scaling rate of SignSGD can be written as $D^{-\min\{2s,\,sq/(sq+1)\}}$. For comparison, the optimal SGD rate under its FSL is $D^{-\min\{s,\,sq/(sq+1)\}}$~\citep{li2026optimal} (Appendix~\ref{app:data-scaling-sgd}). Thus, these data-scaling rates partition the $(s,q)$ space into three regimes: easy, hard, and super hard. The phase diagram on the right summarizes these regimes, with boundaries at $s=1-1/q$ and $s=1/2-1/q$. The corresponding SGD and SignSGD rates are $D^{-\gamma_{\sgd}}$ and $D^{-\gamma_{\sgn}}$, where
\end{minipage}\hfill
\begin{minipage}[t]{0.4\textwidth}
\centering
\raisebox{\dimexpr\ht\strutbox-\height\relax}{%
  \includegraphics[width=\linewidth]{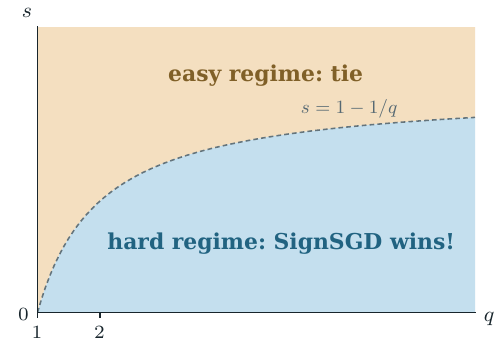}%
}
\end{minipage}
\[
(\gamma_{\sgd},\gamma_{\sgn})=
\begin{cases}
(\tfrac{sq}{sq+1},\tfrac{sq}{sq+1}), & s\ge 1-\tfrac1q,\\[3pt]
(s,\tfrac{sq}{sq+1}), & \tfrac12-\tfrac1q\le s<1-\tfrac1q,\\[3pt]
(s,2s), & 0<s<\tfrac12-\tfrac1q.
\end{cases}
\]

\begin{remark}[Connection to statistical lower bounds.]
    Classical results give the minimax rate $D^{-sq/(sq+1)}$ for the regression problem with the same spectral-decay and source conditions~\citep{caponnetto2007optimal,steinwart2009optimal}.  This rate is established in the discrete setting and it matches the optimal rate in Theorem~\ref{thm:sign-data-scaling} under our continuous SDE model when $s\ge\tfrac12-\tfrac1{q}$.
\end{remark}

\subsection{Step efficiency}
\label{subsec:critical-batch-size}

At a fixed data budget $D=NB$, increasing the batch size reduces the number of training steps. We therefore ask two questions: \emph{(i) Can we increase the batch size while preserving each optimizer's small-batch risk? (ii) If so, how large can the batch size be?} The critical batch size (CBS) characterizes the largest such batch size, with the LR schedule allowed to be retuned.

\begin{definition}[Critical batch size]
\label{def:critical-batch-size}
Let $\mathcal A_{D,B}$ denote the admissible LR
schedules for a given optimizer, and define
$
\cE_{\star}(D,B)
:=
\inf_{\eta\in\mathcal A_{D,B}}
\EE[\cE(\vtheta_{\Gamma_B(D/B)})].
$
For a fixed constant $\delta>0$, we define
\[
B_{\mathrm{crit}}(D)
:=
\sup\left\{
B\in\{1,\ldots,D\}:
\cE_{\star}(D,B)
\le (1+\delta)\,\cE_{\star}(D,1)
\right\},
\qquad
N_{\mathrm{crit}}(D)
:=
\frac{D}{B_{\mathrm{crit}}(D)}.
\]
Here $\Gamma_B$ denotes the optimizer's intrinsic-time map
for the chosen schedule $\eta$, and $N_{\mathrm{crit}}(D)$ is the
minimum number of steps needed to preserve the optimal
small-batch risk within the factor $1+\delta$.
\end{definition}

For SGD, admissibility requires $\eta_{\peak}\lesssim1$. For SignSGD, it requires $\eta_{\peak}\sqrt B\lesssim1$. We consider $0<s<3/2-1/q$, where the SignSGD training-time condition is compatible with the batch-size ranges below. The boundary $s=3/2-1/q$ and the \emph{super-easy regime} $s>3/2-1/q$ are discussed in Appendix~\ref{app:critical-batch-size}. For notational convenience, we write $\alpha=s+\frac1q$.

\begin{theorem}[Critical batch size]
\label{cor:batch-size-range}
Suppose that the assumptions of Theorem~\ref{thm:main-fsl-sign} hold and that $0<s<3/2-1/q$. Then the critical batch sizes satisfy
\begin{equation}
\label{eq:cbs-scaling}
B_{\mathrm{crit}}^{\sgd}(D)
\eqsim \max\left\{1,D^{1-\frac{1}{\alpha}}\right\},
\qquad
B_{\mathrm{crit}}^{\sgn}(D)
\eqsim \max\left\{1,D^{1-\frac{1}{2\alpha}}\right\}.
\end{equation}
\end{theorem}

\paragraph{Peak LR scaling.}
The batch-size ranges in Theorem~\ref{cor:batch-size-range} are attained by the power-decay schedule~\eqref{eq:power-decay} with $N=D/B$, any fixed $p>q-1$, and peak learning rates
\[
\eta_{\peak}^{\sgd}
\eqsim \min\left\{B\,D^{-1+\frac{1}{\alpha}},\,1\right\},
\qquad
\eta_{\peak}^{\sgn}
\eqsim \min\left\{\sqrt B\,D^{-1+\frac{1}{2\alpha}},\,B^{-1/2}\right\}.
\]
In the easy-task regime, the linear scaling rule for SGD and the square-root scaling rule for SignSGD preserve the small-batch risk as $B$ increases. Their admissibility constraints, $\eta_{\peak}^{\sgd}\lesssim1$ and $\sqrt B\,\eta_{\peak}^{\sgn}\lesssim1$, determine the batch-size ranges over which these rules apply. 

\Needspace{11\baselineskip}
\begingroup
\setlength{\intextsep}{0pt}
\begin{wrapfigure}{r}{0.30\textwidth}
    \centering
    \includegraphics[width=\linewidth]
    {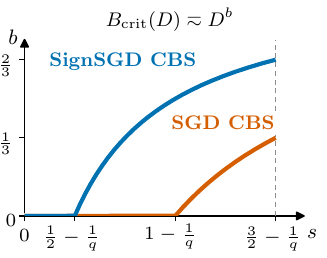}
\end{wrapfigure}

\paragraph{Efficiency gains across task regimes.}
The figure on the right compares the CBS exponents $b$ in $B_{\mathrm{crit}}(D)\eqsim D^b$. In the easy-task regime, $1-1/q\le s<3/2-1/q$, both optimizers attain excess risk of order $D^{-s/\alpha}$. At their respective CBSs, SGD requires $N_{\mathrm{crit}}^{\sgd}\eqsim D^{1/\alpha}$ updates, whereas SignSGD requires only $N_{\mathrm{crit}}^{\sgn}\eqsim D^{1/(2\alpha)}$. Thus, SignSGD uses a factor $D^{1/(2\alpha)}$ fewer sequential updates while maintaining the same data-scaling rate. For hard tasks, $\tfrac12-\tfrac1q<s<1-\tfrac1q$, SignSGD achieves both a faster data-scaling rate and a CBS that grows with $D$, whereas SGD has $B_{\mathrm{crit}}^{\sgd}(D)\eqsim1$. These CBSs preserve each optimizer's small-batch risk, so SignSGD improves both data and step efficiency. For super-hard tasks, $s\le\tfrac12-\tfrac1q$, both CBSs are of constant order, requiring $N\eqsim D$ updates to preserve their respective small-batch risks. SignSGD's advantage in this regime therefore lies in data efficiency rather than step efficiency.
\WFclear
\endgroup

\section{Experiments}
\label{sec:experiments}

We evaluate the predictive performance of FSL in PLK regression
and language model pre-training.
The FSL for SignSGD motivates the parameterized form
\begin{equation}
\label{eqn:fsl-parametrized-discrete}
L_{\Theta}(k)=L_\infty+S_k+N_k,
\qquad
S_k=C_S(A_S+t_k)^{-\alpha},
\qquad
N_k=C_K\sum_{j=0}^{k-1}
(A_K+t_k-t_j)^{-\nu}\eta_j^2,
\end{equation}
where $L_\infty$ is the irreducible loss, $S_k$ is the signal
term, and $N_k$ is the noise term.
The trainable parameters are
$\Theta=(L_\infty,C_S,A_S,\alpha,C_K,A_K,\nu)$,
and $t_k=\sqrt B\sum_{j=0}^{k-1}\eta_j$ is the intrinsic time
associated with the LR schedule $(\eta_0,\ldots,\eta_{N-1})$.
Further details are provided in Appendix~\ref{app:fsl-detail}.

\subsection{PLK regression}
\label{subsec:synthetic}

\paragraph{FSL accurately captures risk dynamics.}
Although our FSL is derived from the continuous-time model~\eqref{eq:sde-sign-app}, \figref[left]{fig:story-overview} shows that the resulting FSL also accurately describes discrete-time training. For PLK regression with $(s,q)=(0.3,4)$, we show accurate fitting and prediction of SignSGD risk curves across cosine, WSD, and cyclic schedules.

\begin{figure}[h]
    \centering
\captionsetup{skip=3pt}

    \includegraphics[width=\linewidth]
    {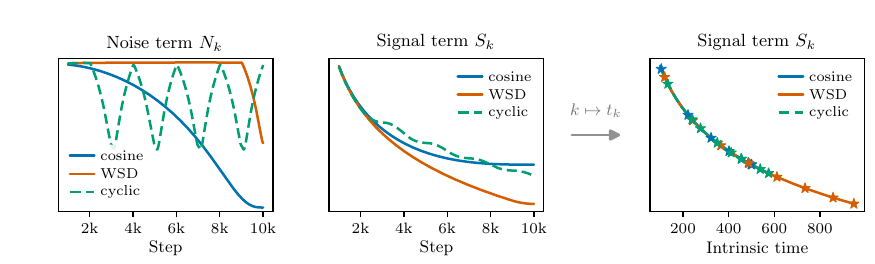}

    \caption{\small \textbf{FSL decomposition and intrinsic-time collapse for SignSGD.}
    We use the same configurations and shared FSL as in
    \figref[left]{fig:story-overview}.
    Solid curves correspond to the cosine and WSD schedules used
    for fitting; dashed curves show predictions for the held-out
    cyclic schedule.
    \textbf{(Left)} Noise term $N_k$ versus training steps.
    \textbf{(Middle)} Signal term $S_k$ versus training steps.
    \textbf{(Right)} The signal terms collapse onto a common curve
    when plotted against intrinsic time.}
    \label{fig:intrinsic-time-collapse}
\end{figure}

\paragraph{Signal--noise decomposition and intrinsic-time collapse.}
Using the experiments in \figref[left]{fig:story-overview},
we decompose the risk trajectories under the three LR schedules
into the noise term $N_k$ and signal term $S_k$,
shown in the left and middle panels of
\figref{fig:intrinsic-time-collapse}, respectively.
Although the signal trajectories differ in training steps,
they collapse onto a common curve when plotted against
intrinsic time in the right panel.
The FSL thus separates schedule-dependent noise effects
from a shared signal-learning profile governed by intrinsic time.

\Needspace{11\baselineskip}
\begingroup
\setlength{\intextsep}{0pt}
\begin{wrapfigure}{r}{0.34\textwidth}
    \centering
    \includegraphics[width=\linewidth,trim=0 0 0 58.5bp,clip]{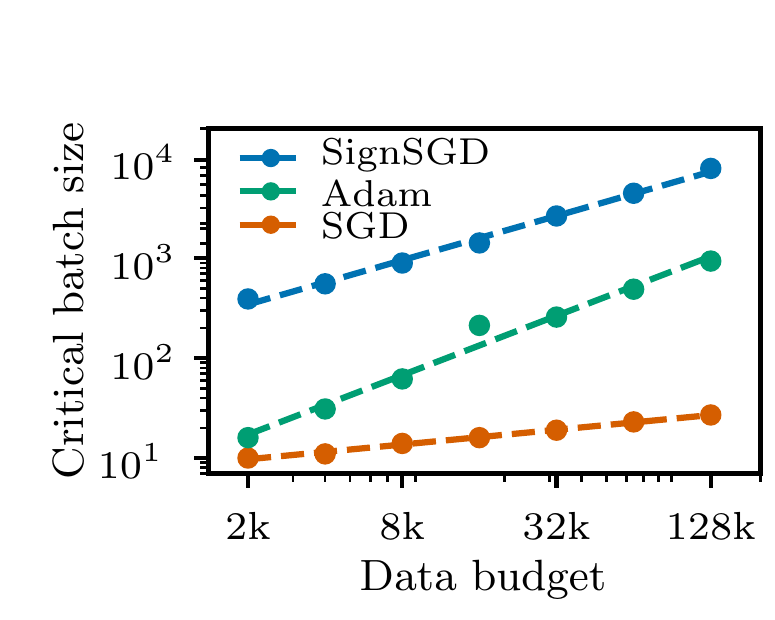}
\end{wrapfigure}

\paragraph{Data scaling and critical batch size.}
We validate the data-scaling and batch-size predictions in Sections~\ref{subsec:data-scaling} and~\ref{subsec:critical-batch-size}. On the hard task with $(s,q)=(0.3,4)$, we tune the peak learning rate for each optimizer and data budget and compare the best final excess risk. \figref[middle]{fig:story-overview} shows that SignSGD achieves a faster data-scaling rate than SGD, consistent with Theorem~\ref{thm:sign-data-scaling}. On the easy task with $(s,q)=(1,4)$, we sweep batch sizes at each data budget and tune the peak learning rate separately for every optimizer and batch size. We estimate the CBS as the largest batch size that approximately preserves each optimizer's small-batch risk. The right panel shows that SignSGD has the largest CBS, followed by Adam and SGD, throughout the tested range. The power-law fits further indicate that the CBS grows faster with the data budget for SignSGD and Adam than for SGD. These findings support the predicted step efficiency of SignSGD and its heuristic extension to Adam. Similar trends appear in language-model training: Muon exhibits greater large-batch robustness than Adam, which in turn outperforms SGD~\citep{marek2025small,sreckovic2025batch,therien2026muloco}. Since Muon applies matrix sign operator to gradient updates, these observations suggest a broader connection between sign-based normalization and batch-size efficiency.
\WFclear
\endgroup

\subsection{Language model pre-training}
\label{subsec:llm_pretrain}
We train a Llama-style language model~\citep{touvron2023llama} with approximately $0.1$B parameters on $10$B tokens, using a realistic token-to-parameter ratio of approximately $100$. We evaluate the FSL for SignSGD and Adam under different LR schedules, including cosine, WSD~\citep{hu2024minicpm}, and the $8\text{-}1\text{-}1$ schedule~\citep{bi2024deepseek}. Additional experimental details are provided in Appendix~\ref{app:additional-experiments}.

{\bf FSL accurately captures loss dynamics across optimizers.} We evaluate the predictive power of FSL across different LR schedules and optimizers. Specifically, we extend the Adam experiments of \citet{li2025functional} to SignSGD. For each optimizer, we fit the FSL parameters using the loss curves of cosine and 8-1-1 schedules, and then use the fitted FSL to predict the loss curves under the WSD schedule. The results for SignSGD and Adam are reported in \figref[a]{fig:llm-fit-predict}. We observe that FSL accurately captures the loss curves for both optimizers and yields accurate predictions for the held-out schedule.

\begin{figure}[!ht]
\centering
\captionsetup{skip=3pt}
\begin{minipage}[t]{0.51\linewidth}
    \centering
    \begin{minipage}[t]{0.49\linewidth}
        \centering
        \includegraphics[width=\linewidth]{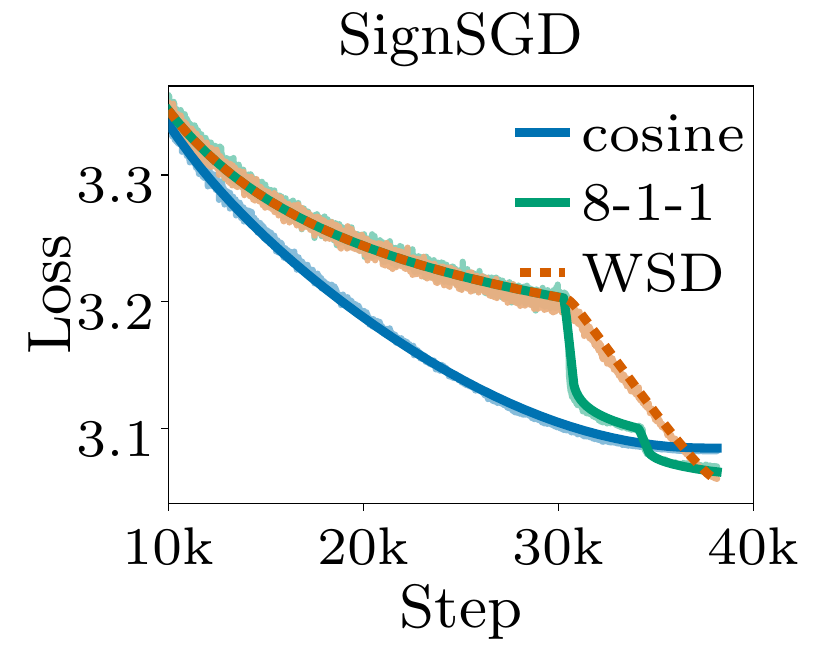}
    \end{minipage}
    \hfill
    \begin{minipage}[t]{0.49\linewidth}
        \centering
        \includegraphics[width=\linewidth]{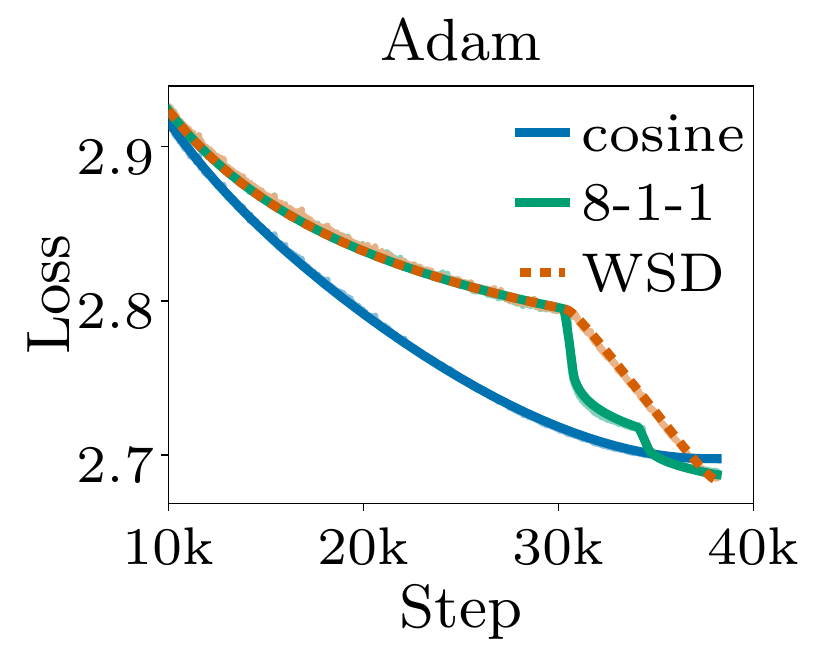}
    \end{minipage}
    \par
    {\small (a) Cross-schedule prediction}
\end{minipage}%
\hfill
\begin{minipage}[t]{0.49\linewidth}
    \centering
    \begin{minipage}[t]{0.49\linewidth}
        \centering
        \includegraphics[width=\linewidth]{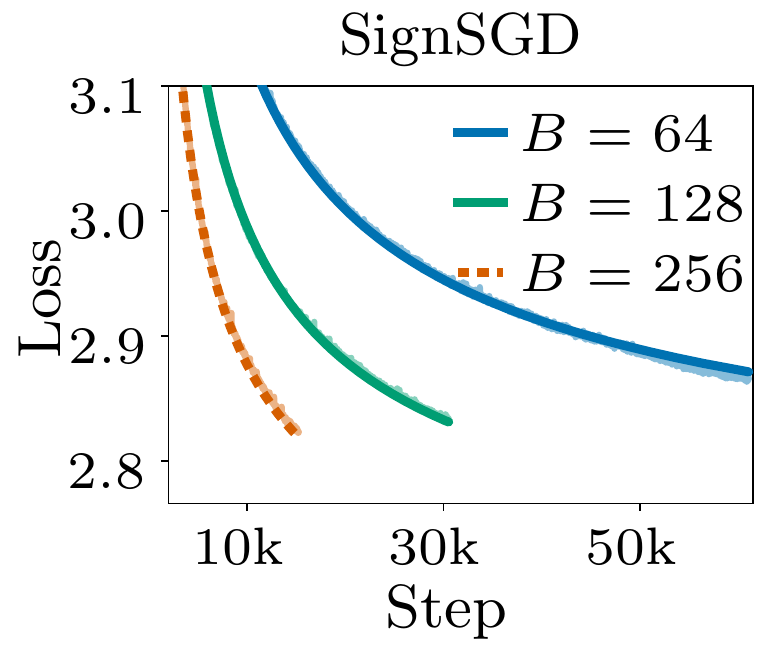}
    \end{minipage}
    \hfill
    \begin{minipage}[t]{0.49\linewidth}
        \centering
        \includegraphics[width=\linewidth]{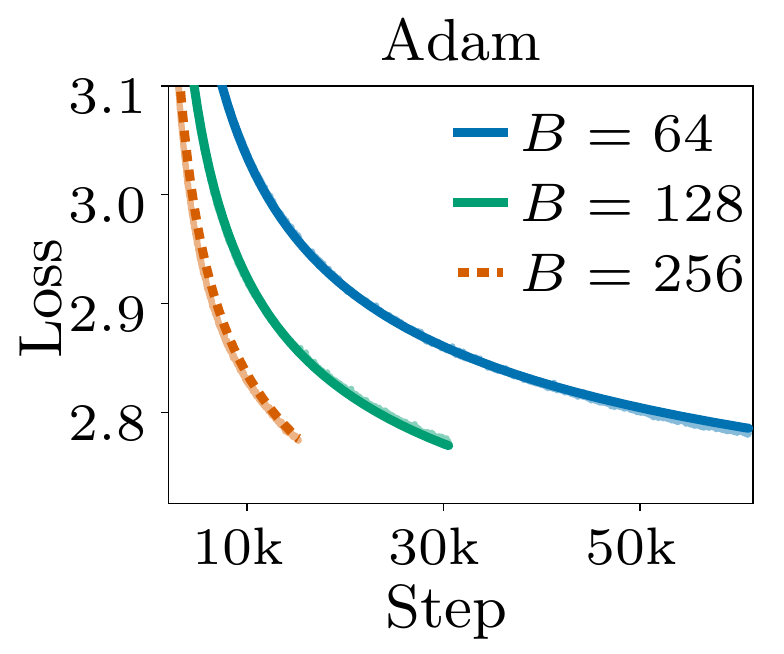}
    \end{minipage}
    \par
    {\small (b) Batch-size extrapolation}
\end{minipage}
\caption{\small \textbf{FSL captures loss dynamics and batch-size dependence in practical language model pre-training.} \textbf{(a)} Cross-schedule loss prediction for SignSGD (left) and Adam (right). For each optimizer, we fit the FSL parameters $\Theta$ to the loss curves under the cosine and $8\text{-}1\text{-}1$ schedules and predict the loss curve under the held-out WSD schedule. \textbf{(b)} Batch-size extrapolation for SignSGD (left) and Adam (right). With a constant learning rate, we fit $\Theta$ to the loss curves at $B=64$ and $B=128$ and predict the loss curve at $B=256$. Solid and dashed FSL curves indicate fits and predictions, respectively.}
\label{fig:llm-fit-predict}
\end{figure}

{\bf FSL accurately captures batch-size dependence and enables extrapolation.} A key difference between the FSL in~\eqref{eqn:fsl-parametrized-discrete} and the SGD-based FSL of~\citet{li2025functional} lies in the batch-size scaling. We test this batch-size dependence in practical training. We train with batch sizes $B\in\{64,128,256\}$ and a constant learning rate. For each optimizer, we fit the FSL using the curves for $B=64$ and $B=128$, and then predict the held-out curve for $B=256$. As shown in \figref[b]{fig:llm-fit-predict}, the FSL jointly fits the two batch sizes and extrapolates well to the larger batch size for both SignSGD and Adam. An additional experiment with cosine schedule for Adam is reported in Appendix~\ref{app:llm}, \figref[left]{fig:llm-batch-additional}.

\section{Why does SignSGD outperform SGD?}
\label{sec:derivation_mechanism}

\subsection{Smoothed sign and one-dimensional dynamics}
\label{subsec:why-fsl-form}

\paragraph{Smoothed sign.}
We first examine the drift and diffusion in~\eqref{eq:sde-sign} in one dimension. Recall that $\sigma_B=1/\sqrt{B}$. Let $a>0$ denote the single-sample gradient-noise standard deviation, so the mini-batch noise standard deviation is $\sigma_B a$. The scaled smoothed sign is $m_B(x)=\sigma_B\EE_{\xi\sim\cN(0,a^2)} [\sgn(x+\sigma_B\xi)]$. Writing $c=\sqrt{2/\pi}$, its two limiting responses are
\begin{equation}
\label{eq:smoothed-sign-regimes}
m_B(x)\approx
\begin{cases}
c\,x/a,
& |x|/(\sigma_B a)\ll1,\\
\sigma_B\sgn(x),
& |x|/(\sigma_B a)\gg1.
\end{cases}
\end{equation}
The variance of the corresponding sign update is $Q_B(x)=1-m_B(x)^2/\sigma_B^2$. Thus, the noise-dominated region has approximately linear drift and $Q_B(x)\approx1$, whereas the signal-dominated region has saturated drift and $Q_B(x)\approx0$.

\Needspace{12\baselineskip}
\begin{wrapfigure}[10]{r}{0.4\textwidth}
    \centering
    \includegraphics[width=\linewidth]
    {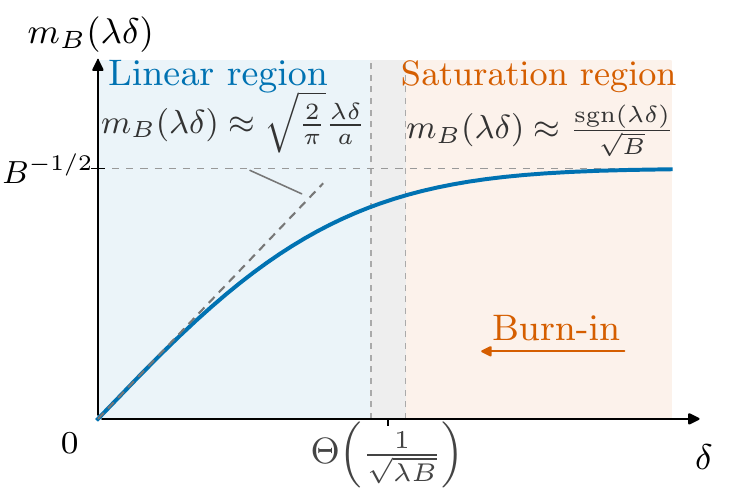}
\end{wrapfigure}

\paragraph{One-dimensional dynamics.}
For a coordinate with curvature $\lambda$, let
$\delta_t=\vartheta_t-\theta^*$, so the population gradient is
$\lambda\delta_t$. Applying~\eqref{eq:smoothed-sign-regimes} to $x=\lambda\delta$ reveals two dynamical regimes. For $\lambda|\delta|/(\sigma_B a)\gg1$, the response saturates at $\sigma_B\sgn(\delta)$ and $Q_B(\lambda\delta)\approx0$, giving approximately deterministic motion toward zero. For $\lambda|\delta|/(\sigma_B a)\ll1$, the response is linear, $m_B(\lambda\delta)\approx c\lambda\delta/a$, and $Q_B(\lambda\delta)\approx1$. The local dynamics therefore have the drift of SGD preconditioned by $c/a$, with approximately unit sign-noise variance. The crossover occurs at $|\delta|\eqsim\sigma_B a/\lambda$. Under noise-curvature alignment $a^2\eqsim\lambda$, this scale becomes $\sigma_B/\sqrt{\lambda} =(\lambda B)^{-1/2}$, as illustrated on the right for $\delta\ge0$.

These two limits describe the saturated dynamics during
burn-in and the local linear dynamics underlying the
scaling regime, respectively.
Substituting them into~\eqref{eq:sde-sign} gives
\WFclear
\begin{equation}
\label{eq:discussion-scalar-dynamics}
\dd\delta_t\approx
\begin{cases}
-\sigma_B\sgn(\delta_t)\,\dd t,
& \dfrac{\lambda|\delta_t|}{\sigma_B a}\gg1
\quad \text{(burn-in phase)},\\[6pt]
-c\,\dfrac{\lambda}{a}\delta_t\,\dd t
+\sqrt{n_B(t)}\,\dd W_t,
& \dfrac{\lambda|\delta_t|}{\sigma_B a}\ll1
\quad \text{(local phase)}.
\end{cases}
\end{equation}

\paragraph{From burn-in to local phase.}
We refer to the initial transient before entering the linear region as \emph{burn-in}. The first branch of~\eqref{eq:discussion-scalar-dynamics} describes its saturated part. Under zero initialization, this approximation gives $|\delta_t|\approx|\theta^*|-\sigma_B t$ until the transition region is reached. The source condition implies $|\theta_j^*|\eqsim j^{(q-1-sq)/2}$. For easy tasks, $s+1/q\ge1$, the initial coordinate errors are uniformly bounded, giving a saturated burn-in time of at most order $\sigma_B^{-1}=\sqrt B$. For hard tasks, $s+1/q<1$, these errors can grow with the coordinate index, and controlling burn-in across the relevant spectral modes requires $t\gtrsim\sigma_B^{-1/(s+1/q)} =B^{1/(2s+2/q)}$. Together, these motivate the training-time condition $t\gtrsim B^{\max\{1/2,\,1/(2s+2/q)\}}$ in Theorem~\ref{thm:main-fsl-sign}. This local analysis provides intuition for the scaling regime. The rigorous argument in Appendix~\ref{app:main proof} uses a Lyapunov function to establish global moment bounds for the nonlinear dynamics, allowing noise-driven excursions outside the local linear region.

\subsection{Noise-induced preconditioning}
\label{subsec:noise-preconditioning}

\paragraph{From coordinate-wise normalization to preconditioning.}
The local linear response in the preceding subsection extends
coordinate-wise to the multidimensional dynamics.
Let $\bdelta_t=\vtheta_t-\btheta^*$ and
$\bD=\diag(\Sigma_{11},\Sigma_{22},\ldots)$.
Linearizing the smoothed sign around the origin gives
$
\bmm_B(\bH\bdelta_t)
\approx c\,\bD^{-1/2}\bH\bdelta_t.
$
Thus, the local drift takes the form
$-\bP\nabla\cR(\vtheta_t)$, where $\bP:=c\,\bD^{-1/2}$.
This is the drift of preconditioned SGD:
the gradient is normalized by its single-sample noise scale on each coordinate.
Such normalization does not by itself guarantee acceleration;
its effectiveness depends on the relation between
the noise scales and the curvature.

\paragraph{Structural conditions for effective preconditioning.}
Noise-curvature alignment and diagonal features play important roles in SignSGD's scaling advantage. The former ensures that $\Sigma_{jj}\eqsim H_{jj}$, so the preconditioner compensates for coordinate-wise curvature. Diagonal features further ensure that these coordinates coincide with the eigendirections of $\bH$. Together, they give $a_j:=\sqrt{\Sigma_{jj}}\eqsim\sqrt{\lambda_j}$ and effective decay rates
$
\rho_j:=\frac{c\lambda_j}{a_j}\eqsim\sqrt{\lambda_j}.
$
The effective spectrum therefore decays as $j^{-q/2}$, rather than $j^{-q}$, reducing the disparity between fast and slow spectral directions. Without noise-curvature alignment, noise normalization need not compensate for curvature. Without diagonal features, coordinate-wise preconditioning need not produce a square-root transformation of the Hessian spectrum.

\paragraph{Effective preconditioned SGD.}
Under our diagonal Gaussian noise model, the sign-noise
coordinates are conditionally independent, and their variances
approach one in the local regime.
Hence $\bQ_B\approx\bI$, yielding
\begin{equation}
\label{eq:effective-preconditioned-dynamics}
\dd\bdelta_t
\approx-\bP\bH\bdelta_t\,\dd t
+\sqrt{n_B(t)}\,\dd\bW_t,
\qquad
\bP=c\,\bD^{-1/2}.
\end{equation}
This has the form of preconditioned SGD with normalized gradient noise. Under noise-curvature alignment and diagonal features, $\bP\bH=\diag(\rho_j)$ with $\rho_j\eqsim\sqrt{\lambda_j}$. The local dynamics therefore combine square-root spectral decay rates with approximately isotropic noise injection. These two changes determine the signal-learning and noise-forgetting profiles derived next.

\begin{figure}[!htbp]
\centering
\captionsetup{skip=3pt}
\includegraphics[width=\linewidth]{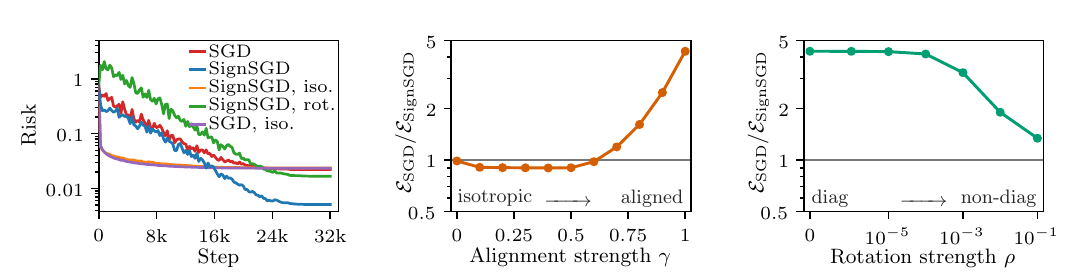}
\caption{\small \textbf{Structural conditions for the SignSGD advantage.} \textbf{(Left)} Risk dynamics of SGD and SignSGD with curvature-aligned noise, isotropic-noise controls, and a rotated feature covariance. \textbf{(Middle)} We vary noise-curvature alignment by setting the noise covariance to $\bSigma\propto\bH^\gamma$ with $\gamma\in[0,1]$. SignSGD shows an advantage only when $\gamma$ is close to $1$. \textbf{(Right)} We vary feature diagonality by rotating the feature covariance, with $\rho$ controlling the rotation strength. As $\rho$ increases, the covariance becomes less diagonal and the SignSGD advantage weakens. All experiments use optimally tuned learning rates.}
\label{fig:structural-conditions}
\end{figure}

\paragraph{Empirical evidence for the structural conditions.}
\figref{fig:structural-conditions} illustrates the role of these two structural ingredients at a fixed data budget. With matched total gradient noise energy and separately tuned learning rates, SignSGD attains a substantially lower final risk in the setting with diagonal features and curvature-aligned noise, while rotations (breaking diagonal structure) or isotropic noise (breaking noise-curvature alignment) reduce this advantage.

\subsection{Deriving the functional scaling law}
\label{subsec:deriving-fsl}

\paragraph{Spectral representation.}
The effective dynamics~\eqref{eq:effective-preconditioned-dynamics}
have diagonal drift $\bP\bH=\diag(\rho_j)$, where
$\rho_j=c\lambda_j/a_j\eqsim\sqrt{\lambda_j}$.
For this linear model initialized at
$\bdelta_0=-\btheta^*$, the expected excess risk decomposes as
\begin{equation}
\label{eq:linearized-risk-decomposition}
\EE[\cE(\vtheta_t)]
\approx
S(t)+\int_0^t K(t-\tau)n_B(\tau)\,\dd\tau,
\end{equation}
where
\[
S(t)=\frac12\sum_j\lambda_j|\theta_j^*|^2e^{-2\rho_jt},\qquad
K(t)=\frac12\sum_j\lambda_j e^{-2\rho_jt}.
\]
The signal profile measures the remaining initial error,
while the forgetting kernel measures the residual risk
from a unit noise injection after elapsed intrinsic time $t$.
Both are mixtures of exponential decays with rates $\rho_j$,
but their spectral weights differ.
These expressions describe the effective linear model;
they do not assume that the original dynamics are linear
throughout burn-in.

\paragraph{From exponential modes to power laws.}
Each spectral mode has characteristic time
$\tau_j:=\rho_j^{-1}\eqsim j^{q/2}$.
At time $t$, modes with $\tau_j\ll t$ have largely decayed,
whereas those with $\tau_j\gg t$ retain substantial contributions. Therefore, we introduce the {\bf moving frontier $j_t$} to characterize which modes are learned at time $t$. It is determined by
$\tau_{j_t}\eqsim t$, and thus satisfies $j_t\eqsim t^{2/q}$.
To estimate the signal profile, split the spectrum at this cutoff:
\begin{equation}
\label{eq:signal-spectral-split}
2S(t)
=
\underbrace{\sum_{j\le j_t}
\lambda_j|\theta_j^*|^2e^{-2\rho_jt}}_{\text{decaying modes}}
+
\underbrace{\sum_{j>j_t}
\lambda_j|\theta_j^*|^2e^{-2\rho_jt}}_{\text{remaining spectral tail}}.
\end{equation}
For $j>j_t$, we have $\rho_jt\lesssim1$,
so the exponential factor is bounded above and below
by positive constants.
The tail contribution is therefore
$\eqsim\sum_{j>j_t}j^{-(1+sq)}\eqsim j_t^{-sq}$.
For $j\le j_t$, the bound $e^{-u}\lesssim u^{-p}$,
with any fixed $p>2s$, gives
\[
\sum_{j\le j_t}\lambda_j|\theta_j^*|^2e^{-2\rho_jt}
\lesssim
t^{-p}\sum_{j\le j_t}j^{-1-sq+pq/2}
\eqsim t^{-p}j_t^{pq/2-sq}
\eqsim j_t^{-sq}.
\]
Thus, the scaling behavior is determined by $j_t^{-sq}$.
Applying the same argument to $K$ yields
\begin{equation}
\label{eq:spectral-tail-scaling}
S(t)\eqsim j_t^{-sq}\eqsim t^{-2s},
\qquad
K(t)\eqsim j_t^{1-q}\eqsim t^{-(2-2/q)}
\end{equation}
for sufficiently large $t$.
Substitution into~\eqref{eq:linearized-risk-decomposition}
recovers the FSL structure.
Appendix~\ref{app:main proof} controls the nonlinear dynamics
to establish the FSL of
Theorem~\ref{thm:main-fsl-sign}.

\paragraph{Faster signal learning, slower noise forgetting.}
The same spectral-cutoff argument explains the difference
between SGD and SignSGD.
In their respective intrinsic times, SGD has decay rates
of order $\lambda_j$ and cutoff $j_t\eqsim t^{1/q}$,
whereas SignSGD has rates of order $\sqrt{\lambda_j}$
and cutoff $j_t\eqsim t^{2/q}$.
The faster advance through the spectrum changes the signal-learning function from $t^{-s}$ to $t^{-2s}$.
Applying the same argument to the forgetting kernels gives decay rates $t^{-(2-1/q)}$ for SGD and $t^{-(2-2/q)}$ for SignSGD, showing that SignSGD forgets noise more slowly.
Thus, the preconditioning effect accelerates signal learning while slows the noise forgetting.

\paragraph{The effect of batch size.}
The comparison above uses each method's own intrinsic time.
For the same LR schedule $\eta(z)$, SGD uses $t_{\sgd}(z)=\int_0^z\eta(\tau)\,\dd\tau$, whereas SignSGD uses $t_{\sgn}(z)=\sqrt B\,t_{\sgd}(z)$.
Consequently, their signal-learning functions expressed in training steps satisfy $S_{\sgn}(z)\eqsim(\sqrt B\,t_{\sgd}(z))^{-2s}$ and $S_{\sgd}(z)\eqsim t_{\sgd}(z)^{-s}$.
The additional $\sqrt B$ factor is from the local response $\EE[\sgn(g_j)]\approx c\sqrt B\,\lambda_j\delta_j/a_j$: a larger batch size reduces gradient noise and increases the mean sign response in the local regime. 
Batch size also enters the noise injection intensity through $n_B(t)=\eta(\Gamma_B^{-1}(t))/\sqrt B$. Together, the intrinsic-time transformation and the noise injection intensity determine how batch size affects signal learning and noise accumulation in the FSL.

\section{Concluding remarks}

Our results show that the advantages of SignSGD depend on
both task difficulty and the structure of gradient noise.
The FSL makes this dependence explicit by tracking the decay
of the initial error and the persistence of injected noise.
Its mechanistic interpretation links the local response
of the sign operator to noise-dependent preconditioning,
whose effectiveness relies on diagonal features and
noise-curvature alignment.
The resulting balance between faster signal learning
and slower noise forgetting distinguishes two benefits:
better use of data on hard tasks, and fewer sequential updates
at comparable data efficiency in the covered easy-task regime.
Experiments corroborate these conclusions in kernel regression
and show that the SignSGD FSL better describes the effect
of batch size on sign-based optimizer trajectories in LLM
pre-training than the original SGD FSL.

Our results motivate two directions for future work:
\begin{itemize}[leftmargin=2em]
\item \textbf{Extensions to RMSprop and Adam.}
Our derivations predict the same FSL form
and batch-size dependence, with supporting numerical evidence
(Appendix~\ref{app:adaptive-extension}
and Appendix~\ref{app:experiments-toy}).
A rigorous treatment would need to account for momentum,
evolving second-moment estimates, and their coupling
with gradient noise.

\item \textbf{From coordinate-wise sign to matrix sign.}
Extending the analysis to the matrix sign operator underlying
Muon would clarify how matrix structure and gradient noise
jointly determine effective preconditioning, signal learning,
noise forgetting, and batch-size dependence.
\end{itemize}

\section*{Acknowledgement}
Lei Wu is supported in part by the National Natural Science Foundation of China (NSFC12522120, NSFC92470122, and NSFC12288101). Binghui Li is supported by the Elite Ph.D. Program in Applied Mathematics at Peking University.

\setlength{\bibsep}{0.25\baselineskip plus 1pt minus 1pt}
\bibliography{scaling-law-ref}
\bibliographystyle{arxiv-bibstyle}

\clearpage
\appendix
\part{Appendix}
\parttoc

% !TEX root = main-arxiv-v0.tex
\newpage
\section{Additional related work}
\label{sec:related_work}

{\bf Theoretical analysis of SignSGD.} Existing  studies  have primarily focused on convergence guarantees of SignSGD~\citep{bernstein2018signsgd,sun2023momentum,jiang2024efficient,jiang2025improved,chen2026and}. Another line of work derives continuous-time stochastic differential equations (SDEs) for modeling SignSGD dynamics~\citep{malladi2022sdes,xiaoexact,compagnoniadaptive}. We build on this continuous-time approach, and use it to derive a trajectory-level functional scaling law for SignSGD.

{\bf Theoretical studies of scaling laws.} A growing literature seeks theoretical explanations for scaling laws~\citep{sharma2020neural,maloney2022solvable,wei2022more,jain2024scaling,michaud2024quantization,nam2024exactly,atanasov2024scaling,dohmatob2024tale,bahri2024explaining,bordelon2024dynamical,lin2024scaling,paquette2024fourplus,bordelon2024feature,li2025functional,kunstner2025scaling,ding2025scaling,yan2025larger,lin2025improved,wang2026fast,li2026optimal,li2026muon,zhang2026precision,li2026signsgdscaling}, but predominantly analyzes vanilla SGD. In contrast, we study SignSGD, which is closer to  optimizers used in large-scale practice, and show that it exhibits qualitatively different scaling behavior. Another related work is \citet{kunstner2025scaling}, which studies full-batch sign gradient descent (signGD) in a linear bigram model and identifies scaling properties of deterministic sign dynamics. We instead focus on the stochastic regime, where gradient noise plays a central role and interacts nontrivially with sign normalization to shape the scaling behavior.

{\bf Comparison with \citet{xiaoexact}.} \citet{xiaoexact} derive a high-dimensional SDE and a deterministic risk ODE for SignSGD, quantifying diagonal preconditioning and noise reshaping. Their formulation allows time-dependent learning rates, and they also discuss a minibatch extension. Their main high-dimensional approximation theorem assumes covariance eigenvalues bounded above and away from zero uniformly in dimension. We instead study a diagonal power-law spectrum with eigenvalues tending to zero, focusing on the label-noise-dominated SDE model. In this setting, we obtain an explicit FSL with power-law signal-learning and noise-forgetting profiles under general LR schedules. This yields optimal data-scaling rates over all admissible schedules and comparisons of critical batch sizes.

{\bf Comparison with \citet{kim2026scaling}.} The most closely related work is \citet{kim2026scaling}, which also provides a scaling-law analysis of SignSGD in PLK regression. Our work differs in three main aspects. First, we focus on the label-noise-dominated regime and establish an explicit FSL for general LR schedules in our diagonal SDE model, whereas their general trajectory formulation is implicit and depends on the unknown risk. Second, we characterize the batch-size dependence through the intrinsic clock $t=\int_0^z\tfrac{\eta(u)}{\sigma_B}\,\dd u$, whereas their theory is restricted to $B=1$. Third, we derive the optimal data-scaling rate of SignSGD over all admissible LR schedules and obtain a complete phase diagram comparing the rates of SignSGD and SGD under optimally tuned LR schedules. In contrast, their comparison is restricted to specific schedules.

\section{Background on SDEs}
\label{app:sde-background}

This appendix reviews two standard tools for analyzing Itô stochastic differential equations (SDEs): the generator for expectation estimates and a time-change formula. These tools are used in the SDE modeling of Appendix~\ref{app:sde-modeling} and in the proof in Appendix~\ref{app:main proof}.

\subsection{The generator for expectation estimates}
\label{app:sde-generator}

We briefly recall how the generator of an Itô SDE can be used to derive expectation and moment estimates. For the details of this method, one can refer to Sec.7 of \citet{oksendal2003stochastic}. We now state a more general version (time-inhomogeneous case) and derive the corresponding results here.

Consider the following SDE
\begin{equation}
\label{eq:general-inhomogeneous-sde}
\dd \bY_t = \bb(t,\bY_t)\,\dd t + \bG(t,\bY_t)\,\dd \bW_t, \qquad \bY_t\in\mathbb R^n,
\end{equation}
where $\bW_t$ is a $d$-dimensional Brownian motion, $\bb(t,\by)\in\mathbb R^n$, and $\bG(t,\by)\in\mathbb R^{n\times d}$.

We first recall the time-homogeneous case, in which $\bb$ and $\bG$ do not depend explicitly on $t$. Let $\bY_t^{\by}$ denote the solution initialized at $\bY_0^{\by}=\by$. The infinitesimal generator is defined by
\begin{equation}
\label{eq:generator-short-time-definition}
(\mathcal L f)(\by) \coloneqq \lim_{h\downarrow0} \frac{ \EE[f(\bY_h^{\by})]-f(\by) }{h},
\end{equation}
whenever the limit exists. For $f\in C^2(\mathbb R^n)$, Itô's formula gives the explicit representation
\begin{equation}
\label{eq:time-homogeneous-generator}
(\mathcal L f)(\by) = \bb(\by)^\top\nabla f(\by) + \frac12 \operatorname{tr}\!\left( \bG(\by)\bG(\by)^\top\nabla^2f(\by) \right).
\end{equation}

For the time-inhomogeneous SDE \eqref{eq:general-inhomogeneous-sde}, we define the instantaneous generator at time $t$ by
\begin{equation}
\label{eq:sde-generator-definition}
(\mathcal L_t f)(\by) \coloneqq \bb(t,\by)^\top\nabla f(\by) + \frac12 \operatorname{tr}\!\left( \bG(t,\by)\bG(t,\by)^\top\nabla^2f(\by) \right).
\end{equation}
Under the standard Markov regularity conditions, this operator agrees with the infinitesimal conditional expectation
\begin{equation}
\label{eq:time-inhomogeneous-generator-limit}
(\mathcal L_t f)(\by) = \lim_{h\downarrow0} \frac{ \EE[f(\bY_{t+h})\mid \bY_t=\by]-f(\by) }{h},
\end{equation}
whenever the limit exists. Thus, $\mathcal L_tf$ describes the instantaneous rate of change of $f(\bY_t)$ after the stochastic fluctuation has been averaged out.

Applying Itô's formula to \eqref{eq:general-inhomogeneous-sde} gives
\begin{equation}
\label{eq:ito-generator}
f(\bY_t) = f(\bY_0) + \int_0^t (\mathcal L_sf)(\bY_s)\,\dd s + \int_0^t \nabla f(\bY_s)^\top \bG(s,\bY_s)\,\dd \bW_s.
\end{equation}
Under suitable integrability conditions, the stochastic integral has zero expectation. Therefore, the generator $\mathcal L_t f$ determines the evolution of expectation $\EE[f(\bY_t)]$.

The following lemma states Dynkin's formula and the upper and lower expectation bounds obtained from generator inequalities. Its proof combines Itô's formula with an integrating factor and also applies to time-inhomogeneous coefficients.

\begin{lemma}[Dynkin's formula and expectation bounds]
\label{lem:generator-dynkin}
Let $\bY_t$ satisfy \eqref{eq:general-inhomogeneous-sde}, and let $V\in C^2(\mathbb R^n;[0,\infty))$. Assume that $\sup_{t\le T}V(\bY_t)$ and $\int_0^T|(\mathcal L_tV)(\bY_t)|\,\dd t$ are integrable. Then
\begin{equation}
\label{eq:dynkin-formula}
\EE[V(\bY_t)] = \EE[V(\bY_0)] + \int_0^t \EE[(\mathcal L_sV)(\bY_s)]\,\dd s.
\end{equation}
Suppose, in addition, that there exist $c_V>0$ and a nonnegative function $r:[0,T]\times\mathbb R^n\to[0,\infty)$ such that
\begin{equation}
\label{eq:abstract-lyapunov-drift}
(\mathcal L_tV)(\by) \le -c_V V(\by)+r(t,\by)
\end{equation}
for every $t\in[0,T]$ and $\by\in\mathbb R^n$. Then
\begin{equation}
\label{eq:abstract-lyapunov-bound}
\EE[V(\bY_t)] \le e^{-c_V t}\EE[V(\bY_0)] + \int_0^t e^{-c_V(t-s)} \EE[r(s,\bY_s)]\,\dd s.
\end{equation}
If instead the generator satisfies
\begin{equation}
\label{eq:abstract-lyapunov-lower-drift}
(\mathcal L_tV)(\by) \ge -c_V V(\by)+r(t,\by)
\end{equation}
for every $t\in[0,T]$ and $\by\in\mathbb R^n$, then
\begin{equation}
\label{eq:abstract-lyapunov-lower-bound}
\EE[V(\bY_t)] \ge e^{-c_V t}\EE[V(\bY_0)] + \int_0^t e^{-c_V(t-s)} \EE[r(s,\bY_s)]\,\dd s.
\end{equation}
\end{lemma}

\begin{proof}
For $R>0$, define the stopping time
\[
\rho_R \coloneqq \inf\left\{t\ge0:\int_0^t\left\|\nabla V(\bY_s)^\top\bG(s,\bY_s)\right\|^2\,\dd s\ge R\right\}.
\]
Applying Itô's formula to $V(\bY_{t\wedge\rho_R})$ gives
\begin{align*}
V(\bY_{t\wedge\rho_R})
={}&
V(\bY_0)
+
\int_0^{t\wedge\rho_R}
(\mathcal L_sV)(\bY_s)\,\dd s
\\
&+
\int_0^{t\wedge\rho_R}
\nabla V(\bY_s)^\top \bG(s,\bY_s)\,\dd \bW_s.
\end{align*}
The stopped stochastic integral is a martingale and therefore has zero expectation. Hence,
\[
\EE[V(\bY_{t\wedge\rho_R})] = \EE[V(\bY_0)] + \EE\left[ \int_0^{t\wedge\rho_R} (\mathcal L_sV)(\bY_s)\,\dd s \right].
\]
Letting $R\to\infty$ under the stated integrability assumptions and applying Fubini's theorem gives \eqref{eq:dynkin-formula}.
If we further assume \eqref{eq:abstract-lyapunov-drift}, differentiating \eqref{eq:dynkin-formula} gives, for almost every $t$,
\[
\frac{\dd}{\dd t}\EE[V(\bY_t)] \le -c_V\EE[V(\bY_t)] + \EE[r(t,\bY_t)].
\]
Multiplying by $e^{c_V t}$ and integrating from $0$ to $t$ proves \eqref{eq:abstract-lyapunov-bound}.
Under \eqref{eq:abstract-lyapunov-lower-drift}, the same argument with both inequalities reversed proves \eqref{eq:abstract-lyapunov-lower-bound}.
\end{proof}

To obtain a moment bound, we choose a test function $V$, establish a pointwise inequality for $\mathcal L_tV$, and apply the corresponding expectation bound in Lemma~\ref{lem:generator-dynkin}. This strategy is used in Appendix~\ref{app:main proof}. Specifically, Corollary~\ref{cor:uniform-second-moment-lower} combines the quadratic drift estimate in Lemma~\ref{lem:uniform-second-moment-lower} with the lower bound \eqref{eq:abstract-lyapunov-lower-bound}. Corollary~\ref{cor:uniform-second-moment-upper} combines the Lyapunov drift estimate in Lemma~\ref{lem:uniform-bounded-noise} with the upper bound \eqref{eq:abstract-lyapunov-bound}.

\subsection{Time-change formula for SDEs}
\label{app:time-change}

We next recall how an SDE transforms under a deterministic time change. This follows from the standard time-change formula for Itô integrals; see Sec.8.5 of \citet{oksendal2003stochastic}. To make the presentation self-contained, we state and prove the version needed here.

\begin{lemma}[Deterministic time change for SDEs]
\label{lem:time-change-general}
Let $\Gamma:[0,Z]\to[0,T]$ be a continuously differentiable bijection satisfying $\Gamma(0)=0$ and $\Gamma'(z)>0$ for all $z\in[0,Z]$. Suppose that $(\widetilde{\bY}_z)_{z\in[0,Z]}$ satisfies
\begin{equation}
\label{eq:timechange-original-sde}
\dd\widetilde{\bY}_z = \ba(z,\widetilde{\bY}_z)\,\dd z + \bC(z,\widetilde{\bY}_z)\,\dd \bW_z,
\end{equation}
where $\bW_z$ is a $d$-dimensional Brownian motion. Define
\[
\overline{\bY}_t \coloneqq \widetilde{\bY}_{\Gamma^{-1}(t)}, \qquad t\in[0,T].
\]
Then there exists a $d$-dimensional Brownian motion $\overline{\bW}_t$ such that
\begin{equation}
\label{eq:timechanged-sde}
\dd\overline{\bY}_t = (\Gamma^{-1})'(t) \ba(\Gamma^{-1}(t),\overline{\bY}_t)\,\dd t + \sqrt{(\Gamma^{-1})'(t)} \bC(\Gamma^{-1}(t),\overline{\bY}_t)\,\dd\overline{\bW}_t.
\end{equation}
\end{lemma}

\begin{proof}
Write \eqref{eq:timechange-original-sde} in integral form. For $t\in[0,T]$,
\begin{equation}
\label{eq:timechange-integral0}
\overline{\bY}_t = \widetilde{\bY}_0 + \int_0^{\Gamma^{-1}(t)} \ba(u,\widetilde{\bY}_u)\,\dd u + \int_0^{\Gamma^{-1}(t)} \bC(u,\widetilde{\bY}_u)\,\dd \bW_u.
\end{equation}
For the drift term, using the change of variables $u=\Gamma^{-1}(s)$, we have
\begin{equation}
\label{eq:timechange-integral1}
\int_0^{\Gamma^{-1}(t)} \ba(u,\widetilde{\bY}_u)\,\dd u = \int_0^t (\Gamma^{-1})'(s) \ba(\Gamma^{-1}(s),\overline{\bY}_s)\,\dd s.
\end{equation}

To make the change of the stochastic integral explicit, define
\[
\overline{\bW}_t \coloneqq \int_0^{\Gamma^{-1}(t)} \sqrt{\Gamma'(u)}\,\dd \bW_u.
\]
We claim that $\overline{\bW}$ is a standard Brownian motion in the new time variable $t$. Indeed, for $0\le s<t$,
\[
\overline{\bW}_t-\overline{\bW}_s = \int_{\Gamma^{-1}(s)}^{\Gamma^{-1}(t)} \sqrt{\Gamma'(u)}\,\dd \bW_u.
\]
This increment is Gaussian with mean zero and covariance
\[
\int_{\Gamma^{-1}(s)}^{\Gamma^{-1}(t)} \Gamma'(u)\,\dd u\,\bI_{d} = \left( \Gamma(\Gamma^{-1}(t))-\Gamma(\Gamma^{-1}(s)) \right)\bI_{d} = (t-s)\bI_{d}.
\]
Moreover, increments over disjoint intervals in $t$ are independent because they are stochastic integrals over disjoint intervals in the original Brownian time $u$. Thus,
\[
\overline{\bW}_t-\overline{\bW}_s \sim \mathcal N(0,(t-s)\bI_{d})
\]
and has independent increments relative to the time-changed filtration. Therefore, $\overline{\bW}$ is a standard $d$-dimensional Brownian motion in the new time variable.

Using $ (\Gamma^{-1})'(t) = \frac{1}{\Gamma'(\Gamma^{-1}(t))}, $ the deterministic time-change formula for Itô integrals gives
\begin{equation}
\label{eq:timechange-integral2}
\int_0^{\Gamma^{-1}(t)} \bC(u,\widetilde{\bY}_u)\,\dd \bW_u = \int_0^t \sqrt{(\Gamma^{-1})'(s)} \bC(\Gamma^{-1}(s),\overline{\bY}_s)\,\dd\overline{\bW}_s.
\end{equation}
Combining \eqref{eq:timechange-integral0}, \eqref{eq:timechange-integral1}, and \eqref{eq:timechange-integral2}, we obtain
\[
\overline{\bY}_t = \overline{\bY}_0 + \int_0^t (\Gamma^{-1})'(s) \ba(\Gamma^{-1}(s),\overline{\bY}_s)\,\dd s + \int_0^t \sqrt{(\Gamma^{-1})'(s)} \bC(\Gamma^{-1}(s),\overline{\bY}_s)\,\dd\overline{\bW}_s.
\]
This is precisely the integral form of \eqref{eq:timechanged-sde}.
\end{proof}

\section{The SDE modeling for SGD and SignSGD}
\label{app:sde-modeling}

Throughout this appendix, $z$ denotes physical time measured in training steps, and $\eta(z)$ is a continuous learning rate schedule satisfying $\eta(k)=\eta_k$.

\subsection{The structure of gradient noise}
\label{app:noise structure}

We first characterize the covariance structure of the gradient noise. The Hessian-aligned structure of SGD noise has been observed and analyzed in prior works on SGD noise geometry \citep{wu2022alignment,wang2023theoretical}. For completeness, we provide a self-contained derivation in our kernel regression model.

We use the batch noise scale $\sigma_B=1/\sqrt B$ from Section~\ref{sec:preliminaries}. Let $\cB$ be a fresh mini-batch of $B$ i.i.d.\ samples and define its gradient by
\[
\bg(\btheta)=\frac1B\sum_{(\bx,y)\in\cB}\nabla_{\btheta}\ell(\btheta;\bx,y).
\]
Define the gradient noise by $\bxi(\btheta)\coloneqq \bg(\btheta)-\nabla\cR(\btheta)$. Then $\EE_{\cB}[\bxi(\btheta)]=\bzero$. At step $k$, evaluating $\bg(\btheta)$ and $\bxi(\btheta)$ at $\btheta_k$ using the fresh batch $\cB_k$ gives $\bg_k$ and $\bxi_k$, respectively. The following theorem shows that the covariance of this noise is aligned with the Hessian.

\begin{theorem}[Gradient-noise covariance]
\label{thm:gradient-noise-covariance}
Consider the kernel regression model in Section \ref{sec:preliminaries}. The mini-batch and single-sample gradient-noise covariances satisfy
\[
\EE_{\cB}[\bxi(\btheta)\bxi(\btheta)^\top]=\sigma_B^2\bSigma(\btheta),\qquad
\bSigma(\btheta)\eqsim\bigl(\cE(\btheta)+\sigma^2\bigr)\bH.
\]
\end{theorem}

Here and below, for two positive semidefinite matrices $\bA$ and $\bB$, we write $\bA\eqsim\bB$ if their quadratic forms are equivalent up to universal constants, i.e., there exist constants $c_1,c_2>0$, independent of $\bv$, such that
$c_1\bv^\top\bB\bv\leq \bv^\top\bA\bv\leq c_2\bv^\top\bB\bv$.

Therefore, for any unit direction $\bv$,
\[
\EE_{\cB}[|\bxi(\btheta)^\top\bv|^2]=\sigma_B^2\bv^\top\bSigma(\btheta)\bv\eqsim\sigma_B^2\bigl(\cE(\btheta)+\sigma^2\bigr)\bv^\top\bH\bv.
\]
Since $\bv^\top\bH\bv$ is the curvature along $\bv$, the gradient noise variance in any direction is proportional to the curvature in that direction (up to constants). In the label-noise-dominated regime $\cE(\btheta) \lesssim \sigma^2$, this simplifies to
\[
\bSigma(\btheta)\eqsim\bH,\qquad \EE_{\cB}[|\bxi(\btheta)^\top\bv|^2]\eqsim\sigma_B^2\bv^\top\bH\bv.
\]
This is due to our assumption $\sigma^2\eqsim 1$. Thus, in coordinate $j$, the mini-batch noise has standard deviation of order $\sigma_B\sqrt{\lambda_j}$. This coordinate-wise noise scale motivates the normalized noise model used below for the SignSGD scaling analysis.

\begin{proof}[Proof of Theorem \ref{thm:gradient-noise-covariance}]
For a mini-batch of size $B$, write $\bxi(\btheta)=\frac1B\sum_{i=1}^B\bzeta_i(\btheta)$, where each $\bzeta_i(\btheta)$ is the gradient noise over a single data point. Then
\[
\begin{aligned}
    \EE[\bxi(\btheta)\bxi(\btheta)^\top]  
    =
    \frac1{B^2}
    \sum_{i,j=1}^B
    \EE[
        \bzeta_i(\btheta)
        \bzeta_j(\btheta)^\top
    ]
    =
    \frac1{B^2}
    \sum_{i=1}^B
    \EE[
        \bzeta_i(\btheta)
        \bzeta_i(\btheta)^\top
    ]
    =
    \frac1B \bSigma(\btheta)
    =\sigma_B^2\bSigma(\btheta).
\end{aligned}
\]
It remains to compute $\bSigma(\btheta)$. Consider a single data point $(\bx,y)$ and let $\bg_{\bx,y}(\btheta)$ be its gradient. Let $ \bdelta=\btheta-\btheta^*, \bphi=\bphi(\bx). $ Since $y=\langle\bphi,\btheta^*\rangle+\epsilon$, we have
\[
\bg_{\bx,y}(\btheta) = \nabla_{\btheta} \left(\frac{1}{2}(f(\bx;\btheta)-y)^2\right) =\bigl(\langle\bphi,\btheta\rangle-y\bigr)\bphi = \bigl(\langle\bphi,\bdelta\rangle-\epsilon\bigr)\bphi.
\]
The noise $\epsilon$ and feature $\bphi$ are independent and satisfy $\EE[\epsilon]=0$, $\EE[\bphi\bphi^\top]=\bH$. Taking expectations over $(\bx,y)$, we obtain
\[
\nabla\cR(\btheta) = \EE_{\bx,y} [\bg_{\bx,y}(\btheta)] = \bH\bdelta.
\]
Thus, the single-sample gradient noise is
\[
\bzeta(\btheta) = \bg_{\bx,y}(\btheta)-\nabla\cR(\btheta) = (\bphi\bphi^\top-\bH)\bdelta-\epsilon\bphi.
\]
We next compute its covariance:
\[
\bSigma(\btheta) = \EE_{\bx,y} [\bzeta(\btheta)\bzeta(\btheta)^\top] = \EE\left[ (\bphi\bphi^\top-\bH) \bdelta\bdelta^\top (\bphi\bphi^\top-\bH) \right] + \sigma^2\EE[\bphi\bphi^\top],
\]
where the cross terms vanish because $\epsilon$ is independent of $\bphi$ and $\EE[\epsilon]=0$. The second term equals $\sigma^2\bH$. For the first term, we have
\[
\begin{aligned}
\EE\left[
(\bphi\bphi^\top-\bH)
\bdelta\bdelta^\top
(\bphi\bphi^\top-\bH)
\right] 
=
\EE\left[
(\bdelta^\top\bphi)^2\bphi\bphi^\top
\right]
-
\bH\bdelta\bdelta^\top\bH
=
(\bdelta^\top\bH\bdelta)\bH
+
\bH\bdelta\bdelta^\top\bH,
\end{aligned}
\]
where we use the Gaussian fourth-moment identity $ \EE\left[ (\bdelta^\top\bphi)^2\bphi\bphi^\top \right] = (\bdelta^\top\bH\bdelta)\bH + 2\bH\bdelta\bdelta^\top\bH . $ Thus, we obtain the noise covariance of single-sample gradient
\[
\bSigma(\btheta) = \bigl(\sigma^2+\bdelta^\top\bH\bdelta\bigr)\bH + \bH\bdelta\bdelta^\top\bH .
\]
We claim that, uniformly over all $\bv\in\RR^d$,
\[
\bv^\top\bSigma(\btheta)\bv \eqsim \bigl(\cE(\btheta)+\sigma^2\bigr)\bv^\top\bH\bv .
\]
Indeed, using the covariance formula above,
\[
\begin{aligned}
\bv^\top\bSigma(\btheta)\bv
&=
\bigl(\sigma^2+\bdelta^\top\bH\bdelta\bigr)\bv^\top\bH\bv
+
(\bdelta^\top\bH\bv)^2 .
\end{aligned}
\]
The lower bound follows immediately by dropping the nonnegative second term: $ \bv^\top\bSigma(\btheta)\bv \ge \bigl(\sigma^2+\bdelta^\top\bH\bdelta\bigr)\bv^\top\bH\bv . $ For the upper bound, the Cauchy--Schwarz inequality gives $ (\bdelta^\top\bH\bv)^2 \le (\bdelta^\top\bH\bdelta)(\bv^\top\bH\bv). $ Therefore,
\[
\bv^\top\bSigma(\btheta)\bv \le \bigl(\sigma^2+2\bdelta^\top\bH\bdelta\bigr)\bv^\top\bH\bv .
\]
Using $\cE(\btheta)=\frac12\bdelta^\top\bH\bdelta$ and combining this with the lower bound, we obtain
\[
\bigl(\cE(\btheta)+\sigma^2\bigr)\bv^\top\bH\bv \le \bv^\top\bSigma(\btheta)\bv \le 4\bigl(\cE(\btheta)+\sigma^2\bigr)\bv^\top\bH\bv .
\]
This proves
\[
\bSigma(\btheta)\eqsim(\cE(\btheta)+\sigma^2)\bH.
\]
\end{proof}

\begin{remark}[Beyond Gaussian features]
Gaussianity is not needed for curvature alignment in the label-noise-dominated regime. Suppose the hypercontractivity condition $\EE[(\bv^\top\bphi)^4]\le C(\bv^\top\bH\bv)^2$ holds for every $\bv\in\RR^d$, with a constant $C$ independent of $\bv$ and $d$. A similar argument gives
\[
\sigma_B^2\sigma^2\bH\lesssim\sigma_B^2\bSigma(\btheta)\lesssim\sigma_B^2\bigl(\sigma^2+\cE(\btheta)\bigr)\bH.
\]
This leads to the same result $\sigma_B^2\bSigma(\btheta)\eqsim\sigma_B^2\bH$ in the label-noise-dominated regime $\cE(\btheta) \lesssim \sigma^2 \eqsim 1$.
\end{remark}

\subsection{The SDE modeling for SGD}
\label{app:sgd-sde}

We briefly review the SDE modeling for SGD in \citet{li2025functional}. The SGD update is
\begin{equation}
\btheta_{k+1} = \btheta_k - \eta_k \bg_k.
\label{eqn:app sgd update}
\end{equation}
We decompose the mini-batch gradient as $ \bg_k = \nabla \cR(\btheta_k)+\bxi_k$. By Theorem~\ref{thm:gradient-noise-covariance}, its conditional moments satisfy
\[
\EE[\bxi_k\mid\btheta_k]=\bzero,\qquad \EE[\bxi_k\bxi_k^\top\mid\btheta_k]=\sigma_B^2\bSigma(\btheta_k).
\]
Here $\bSigma(\btheta_k)$ denotes the covariance of a single-sample stochastic gradient.

Therefore, one SGD step can be written as $ \btheta_{k+1}-\btheta_k = -\eta_k \nabla \cR(\btheta_k) - \eta_k \bxi_k, $ consisting of a drift term $-\eta_k\nabla \cR(\btheta_k)$ and a fluctuation term with covariance $\eta_k^2\bSigma(\btheta_k)/B$. This motivates the physical-time SDE:
\begin{equation}\label{eq:sde-sgd-physical}
\dd \widetilde{\btheta}_z = -\eta(z)\nabla \cR(\widetilde{\btheta}_z)\,\dd z + \frac{\eta(z)}{\sqrt B} \bSigma(\widetilde{\btheta}_z)^{1/2}\,\dd \bW_z,
\end{equation}
where $\eta(z)$ is the continuous-time LRS satisfying $\eta(k) = \eta_k$ for $k\in\NN$, and $z$ denotes the physical time, which is a continuous analogue of the training step. To account for the influence of LRS, we define the intrinsic time $ t=\Gamma_{\sgd}(z)\coloneqq\int_0^z \eta(u)\,\dd u, $ which can be viewed as the continuous-time analogue of $\sum_{j=0}^{k-1}\eta_j$. Let $ \vtheta_t=\widetilde{\btheta}_{\Gamma_{\sgd}^{-1}(t)} $. Since $ \tfrac{\dd}{\dd t}\Gamma_{\sgd}^{-1}(t) = \tfrac{1}{\eta(\Gamma_{\sgd}^{-1}(t))}, $ applying the time-change formula in Lemma~\ref{lem:time-change-general} with $ \ba(z,\btheta)=-\eta(z)\nabla\cR(\btheta), \bC(z,\btheta)=\tfrac{\eta(z)}{\sqrt B}\bSigma(\btheta)^{1/2}, $ yields the intrinsic-time SDE
\begin{equation}
\label{eq:sde-sgd}
\dd \vtheta_t = -\nabla \cR(\vtheta_t)\,\dd t + \sqrt{\frac{\varphi_{\sgd}(t)}{B}}\, \bSigma(\vtheta_t)^{1/2}\,\dd \bW_t,
\end{equation}
where $\varphi_{\sgd}(t)\coloneqq \eta(\Gamma_{\sgd}^{-1}(t))$ is the learning rate schedule in intrinsic time. By the time change, we decouple the effect of the learning rate from the drift term.

Here $t=t_{\sgd}(z)=\Gamma_{\sgd}(z)$, and the noise injection intensity is $n_{\sgd}(t)=\varphi_{\sgd}(t)/B$.

\subsection{The SDE modeling for SignSGD}
\label{app:sde-sign}

We now derive the SDE modeling for SignSGD. The SignSGD update is
\begin{equation}
\btheta_{k+1} = \btheta_k - \eta_k \sgn(\bg_k).
\label{eqn:app sign update}
\end{equation}

As in Section~\ref{sec:preliminaries}, write $\bg_k=\bmu_k+\bxi_k$, where $\bmu_k=\nabla\cR(\btheta_k)$, and use the fixed-covariance Gaussian model $\bxi_k\sim\cN(\bzero,\sigma_B^2\bSigma)$. Recall the scaled smoothed sign operator for batch size $B$:
\[
\bmm_B(\bx)\coloneqq\sigma_B
\EE_{\bxi\sim\cN(\bzero,\bSigma)}
\bigl[\sgn(\bx+\sigma_B\bxi)\bigr].
\]
Then
\[
\EE[\sgn(\bg_k)\mid\btheta_k]
=\bmm_B(\bmu_k)/\sigma_B.
\]
Define the centered sign noise and its covariance by
\[
\bn_k\coloneqq\sgn(\bg_k)-\bmm_B(\bmu_k)/\sigma_B,
\qquad
\bQ_B(\btheta_k)\coloneqq
\EE[\bn_k\bn_k^\top\mid\btheta_k].
\]
Thus the update reads
\[
\btheta_{k+1}-\btheta_k
=-\frac{\eta_k}{\sigma_B}\,\bmm_B(\nabla\cR(\btheta_k))-\eta_k\bn_k.
\]
Matching its conditional mean and covariance gives the physical-time model
\begin{equation}\label{eqn:sde-sign-physical-app}
\dd\widetilde{\btheta}_z
=-\frac{\eta(z)}{\sigma_B}\,\bmm_B(\nabla\cR(\widetilde{\btheta}_z))\,\dd z
+\eta(z)\bQ_B(\widetilde{\btheta}_z)^{1/2}\,\dd\bW_z.
\end{equation}
Define the SignSGD intrinsic clock, its intrinsic-time learning rate, and its noise injection intensity by
\[
\begin{aligned}
t=\Gamma_B(z)&\coloneqq\int_0^z\frac{\eta(u)}{\sigma_B}\,\dd u,\\
\varphi(t)=\varphi_{\sgn}(t)&\coloneqq\eta(\Gamma_B^{-1}(t)),
\qquad n_B(t)\coloneqq \sigma_B\varphi(t).
\end{aligned}
\]
Set $\vtheta_t=\widetilde{\btheta}_{\Gamma_B^{-1}(t)}$. Applying Lemma~\ref{lem:time-change-general} with
\[
\ba(z,\btheta)=-\frac{\eta(z)}{\sigma_B}\,\bmm_B(\nabla\cR(\btheta)),
\qquad
\bC(z,\btheta)=\eta(z)\bQ_B(\btheta)^{1/2},
\]
and $(\Gamma_B^{-1})'(t)=\sigma_B/\varphi(t)$ gives
\begin{equation}\label{eq:sde-sign-app}
\dd\vtheta_t
=-\bmm_B(\nabla\cR(\vtheta_t))\,\dd t
+\sqrt{n_B(t)}\,\bQ_B(\vtheta_t)^{1/2}\,\dd\bW_t.
\end{equation}
This is the same SDE as \eqref{eq:sde-sign}. Its instantaneous diffusion covariance is $n_B(t)\bQ_B(\vtheta_t)$, and $\EE[\cE(\vtheta_t)]$ denotes the expected excess risk. At zeros of the learning rate the dynamics are understood by continuity. On any interval where $\eta=0$, both the parameter process and its clock are constant.

Under Assumption~\ref{ass:curvature-aligned-noise}, the marginal standard deviations satisfy $a_j\coloneqq\sqrt{\Sigma_{jj}}\eqsim\sqrt{\lambda_j}$. We use these marginals to derive the coordinate equations for Appendix~\ref{app:main proof}.

\begin{lemma}[Smoothed sign under Gaussian noise]
\label{lem:smoothed-sign-gaussian}
Let $Z\sim\cN(0,1)$, let $\tau>0$, and define the smoothed sign function $m(u)\coloneqq 2\Phi(u)-1$. Then, for any $\mu\in\RR$,
\[
\EE[\sgn(\mu+\tau Z)] = m\left(\frac{\mu}{\tau}\right), \qquad \var(\sgn(\mu+\tau Z)) = 1- m\left(\frac{\mu}{\tau}\right)^2 .
\]
\end{lemma}

\begin{proof}
Since $\sgn(\mu+\tau Z)=1$ on the event $Z>-\mu/\tau$ and equals $-1$ on its complement, we have
\[
\EE[\sgn(\mu+\tau Z)] = \PP\left(Z>-\frac{\mu}{\tau}\right) - \PP\left(Z\le -\frac{\mu}{\tau}\right) = 2\Phi\left(\frac{\mu}{\tau}\right)-1 = m\left(\frac{\mu}{\tau}\right).
\]
Moreover, $\sgn(\mu+\tau Z)^2=1$ almost surely, so
\[
\var(\sgn(\mu+\tau Z)) = \EE[\sgn(\mu+\tau Z)^2] - \EE[\sgn(\mu+\tau Z)]^2 = 1- m\left(\frac{\mu}{\tau}\right)^2 .
\]
\end{proof}

Applying Lemma~\ref{lem:smoothed-sign-gaussian} coordinate-wise to the drift and sign-noise covariance in \eqref{eq:sde-sign-app} yields the following explicit dynamics.

\begin{theorem}[Coordinate-wise dynamics]
\label{thm:coordinatewise-sde-sign}
Suppose $\bH=\diag(\lambda_j)$ and $\bxi\sim\cN(\bzero,\sigma_B^2\bSigma)$ with $\bSigma\eqsim\bH$. Set $k_j=\lambda_j/(\sigma_Ba_j)$. For each coordinate $j$, we have
\[
[\bmm_B(\nabla\cR(\btheta))]_j
=\sigma_B m\!\left(k_j(\theta_j-\theta_j^*)\right),
\qquad
[\bQ_B(\btheta)]_{jj}
=1-m\!\left(k_j(\theta_j-\theta_j^*)\right)^2.
\]
Accordingly, the $j$-th coordinate of \eqref{eq:sde-sign-app} satisfies
\begin{equation}\label{eq:1D-sde-original}
\dd X_j(t)
=-\sigma_B m\!\left(k_jX_j(t)\right)\dd t
+\sqrt{n_B(t)
\left[1-m\!\left(k_jX_j(t)\right)^2\right]}
\,\dd\widetilde W_{j,t},
\end{equation}
where $X_j(t)=(\vtheta_t)_j-\theta_j^*$ and $X_j(0)=-\theta_j^*$ under the initialization $\vtheta_0=\bzero$.
\end{theorem}
\begin{proof}
The $j$-th gradient coordinate has mean $\lambda_j(\theta_j-\theta_j^*)$ and noise standard deviation $\sigma_Ba_j$. Lemma~\ref{lem:smoothed-sign-gaussian} therefore gives its mean sign $m(k_j(\theta_j-\theta_j^*))$ and variance $1-m(k_j(\theta_j-\theta_j^*))^2$. Multiplying the mean by $\sigma_B$ gives the scaled drift. The $j$-th coordinate martingale in~\eqref{eq:sde-sign-app} has instantaneous variance $n_B(t)[\bQ_B(\vtheta_t)]_{jj}$, so it admits the scalar Brownian representation in~\eqref{eq:1D-sde-original}. The Brownian motions for different coordinates may be correlated.
\end{proof}

\subsection{Properties of the smoothed sign function}
\label{app:smoothed-sign-properties}

We collect the properties of the smoothed sign function $m(z)=2\Phi(z)-1$ introduced in Appendix~\ref{app:sde-sign}. These estimates will be used in Appendix~\ref{app:main proof} to obtain lower and upper bounds for the modal second moments. Throughout the following analysis, we write
\[
c\coloneqq m'(0)=\sqrt{2/\pi}.
\]

\begin{lemma}[Properties of the smoothed sign function]
\label{lem:smoothed-sign-properties}
The smoothed sign function $m$ has the following properties.
\begin{enumerate}[leftmargin=2em]
\item The function $m$ is smooth, odd, and strictly increasing, with $|m(z)|<1$ for any $z\in\mathbb R$.
\item The following bounds hold globally:
\begin{equation}
\label{eq:smoothed-sign-global-bounds}
0\le zm(z)\le cz^2,
\qquad m(z)^2\le c^2z^2,
\qquad z\in\mathbb R.
\end{equation}
\item Near the origin, we have
\begin{equation}
\label{eq:smoothed-sign-local-bounds}
\begin{aligned}
m(z)&=cz+O(z^3), && z\to0,\\
zm(z)&\ge\frac c2z^2, && |z|\le1.
\end{aligned}
\end{equation}
\item In the tails,
\begin{equation}
\label{eq:smoothed-sign-tail-limits}
m(z)\operatorname{sgn}(z)\to1,
\qquad 1-m(z)^2\to0,
\qquad |z|\to\infty.
\end{equation}
\end{enumerate}
\end{lemma}

\begin{proof}
The definition of $m$ gives
\[
m(z)=c\int_0^z e^{-u^2/2}\,\dd u,
\qquad m'(z)=ce^{-z^2/2}.
\]
Thus $m$ is smooth, odd, and strictly increasing. Since $0<\Phi(z)<1$ for any finite $z$, we also have $|m(z)|<1$. The integral representation gives $|m(z)|\le c|z|$, and $m(z)$ has the same sign as $z$. These observations prove \eqref{eq:smoothed-sign-global-bounds}.

Expanding the integrand at zero gives $m(z)=cz+O(z^3)$. For $|z|\le1$, the same integral representation yields
\[
zm(z)=c|z|\int_0^{|z|}e^{-u^2/2}\,\dd u
\ge ce^{-1/2}z^2
\ge\frac c2z^2,
\]
which proves \eqref{eq:smoothed-sign-local-bounds}. Finally, $\Phi(z)\to1$ as $z\to\infty$ and $\Phi(z)\to0$ as $z\to-\infty$, giving \eqref{eq:smoothed-sign-tail-limits}.
\end{proof}

\subsection{An extension to non-Gaussian gradient noise}
\label{app:non-gaussian-noise}

This extension is supplementary and may be skipped when reading the main proof in Appendix~\ref{app:main proof}, which uses the Gaussian SDE from Appendix~\ref{app:sde-sign}. We show that the original mini-batch noise also yields the relevant smoothed-sign properties for integer $B\ge2$.

\paragraph{Mini-batch gradient and density at zero.}
Fix $\bdelta=\btheta-\btheta^*$ and a coordinate $j$. For independent samples from the Gaussian regression model,
\[
g_{B,j}=\frac1B\sum_{i=1}^B\phi_{ij}\bigl(\bphi_i^\top\bdelta-\epsilon_i\bigr)
=\lambda_j\delta_j+\xi_j.
\]
Set
\[
\sigma_{\mathrm{res},j}^2=\sigma^2+\sum_{\ell\ne j}\lambda_\ell\delta_\ell^2,
\qquad
R_B=\frac1{B\lambda_j}\sum_{i=1}^B\phi_{ij}^2\sim\frac{\chi_B^2}{B}.
\]
The residual $\sum_{\ell\ne j}\phi_{i\ell}\delta_\ell-\epsilon_i$ is $\cN(0,\sigma_{\mathrm{res},j}^2)$ and independent of $\phi_{ij}$. Conditioning on the coordinate features gives
\[
g_{B,j}\overset{\rm d}=\lambda_j\delta_jR_B
+\sigma_{\mathrm{res},j}\sqrt{\frac{\lambda_j}{B}}\sqrt{R_B}\,Z,
\qquad Z\perp R_B.
\]
Here $Z\sim\cN(0,1)$. At zero coordinate error, $\delta_j=0$, the noise is a symmetric Gaussian scale mixture with $\widehat\sigma_j=\sigma_{\mathrm{res},j}\sqrt{\lambda_j/B}$. Its density $f_j$ satisfies
\[
f_j(0)=\frac{\EE[R_B^{-1/2}]}{\sqrt{2\pi}\,\widehat\sigma_j}
=\frac{\sqrt B}{2\sqrt\pi\,\widehat\sigma_j}
\frac{\Gamma((B-1)/2)}{\Gamma(B/2)}
\eqsim\frac1{\widehat\sigma_j}.
\]
For $B\ge3$, $\EE[R_B^{-1}]=B/(B-2)\le3$ gives $1\le\EE[R_B^{-1/2}]\le\sqrt3$. For $B=2$, the mixture has the Laplace density
\[
f_j(x)=\frac1{\sqrt2\,\widehat\sigma_j}
\exp\left(-\frac{\sqrt2\,|x|}{\widehat\sigma_j}\right).
\]
Thus the density is continuous, finite, and strictly positive at zero, with $f_j(0)\eqsim\widehat\sigma_j^{-1}$ uniformly over $B\ge2$.

\paragraph{Smoothed sign and its role in the proof.}
To account for the dependence of the noise on $\bdelta$, compute the mean sign directly. With $m(u)=2\Phi(u)-1$, conditioning gives
\begin{equation}
\label{eq:full-minibatch-sign-response}
\EE[\sgn(g_{B,j})\mid\bdelta]
=\widetilde m_B\left(\frac{\sqrt{B\lambda_j}\,\delta_j}{\sigma_{\mathrm{res},j}}\right),
\qquad
\widetilde m_B(u)\coloneqq\EE[m(u\sqrt{R_B})].
\end{equation}
Integrating against the $\chi_B^2$ density yields
\[
\widetilde m_B'(u)
=\frac2{\sqrt{\pi B}}\frac{\Gamma((B+1)/2)}{\Gamma(B/2)}
\left(1+\frac{u^2}{B}\right)^{-(B+1)/2}.
\]
Hence $\widetilde m_B'(0)\eqsim1$, with a uniform upper bound on the derivative and a positive lower bound on $|u|\le1$. For $B\ge3$, $\widetilde m_B(u)=\EE[\sgn(u+Z/\sqrt{R_B})]$ and $\EE[Z^2/R_B]\le3$, so Chebyshev's inequality gives $1-\widetilde m_B(u)^2\le12/u^2$ for $|u|\ge1$. For $B=2$, direct integration gives
\[
\widetilde m_2(u)=\frac{u}{\sqrt{u^2+2}},
\qquad
1-\widetilde m_2(u)^2=\frac2{u^2+2},
\]
which yields the same tail bound. These estimates provide the local linear response, global linear upper bound, and restoring drift used in Appendix~\ref{app:main proof}, uniformly over $B\ge2$.

If $\sigma^2\eqsim1$ and $\bdelta^\top\bH\bdelta\lesssim1$, then $\sigma_{\mathrm{res},j}\eqsim1$ uniformly in $j$. Consequently, the response in \eqref{eq:full-minibatch-sign-response} has the same coordinate-wise scale $\sqrt{B\lambda_j}$ as the Gaussian model of Appendix~\ref{app:sde-sign}, with comparable local response and restoring drift bounds uniformly over $B\ge2$. These estimates support the Gaussian modeling choice. Extending the full FSL proof to the original mini-batch noise would additionally require controlling the state-dependent residual scales along the trajectory.

\section{FSL of SignSGD (Proof of Theorem~\ref{thm:main-fsl-sign})}
\label{app:main proof}

Throughout this appendix, $t=\Gamma_B(z)=\sqrt B\int_0^z\eta(u)\,\dd u$ denotes SignSGD intrinsic time. We work under the assumptions and the learning-rate and training-time conditions of Theorem~\ref{thm:main-fsl-sign}, using the Gaussian model with $\bSigma\eqsim\bH$ specified in Appendix~\ref{app:sde-sign}.

We use $\varphi(t)=\varphi_{\sgn}(t)=\eta(\Gamma_B^{-1}(t))$ and $n_B(t)=\varphi(t)/\sqrt B$, consistently with Section~\ref{sec:preliminaries}. The scalar $m(u)=2\Phi(u)-1$ is the unscaled Gaussian sign response; the scaled vector drift in the main text is $\bmm_B$.

We prove the functional scaling law by obtaining upper and lower bounds for every modal second moment and then summing over the spectrum. The lower bound follows directly from the exact second-moment equation. For the upper bound, a Foster--Lyapunov estimate controls the dynamics, while the training-time requirement ensures burn-in control for modes with large normalized initial values. Throughout the proof, constants may depend on $q,s$, the fixed bound $C_0$ in $\sqrt B\,\eta_{\peak}\le C_0$, and the comparison constants in Assumptions~\ref{ass:power-law} and~\ref{ass:curvature-aligned-noise}, but not on $B,t$ or the particular learning rate schedule $\varphi$.

\subsection{A technical lemma: power-law spectral sums}
\label{app:uniform-spectral-summation}

The modal estimates below reduce the signal learning and noise accumulation terms to sums of exponentials over the spectrum. The following lemma converts both sums into power laws and will be used in Appendix~\ref{app:uniform-proof-completion} to complete the proof of Theorem~\ref{thm:main-fsl-sign}.

\begin{lemma}[Power-law spectral sums]
\label{lem:uniform-characteristic-summation}
Let $r>1$ and $\theta>0$, and suppose $b_j\eqsim j^{-r}$ and $\rho_j\eqsim j^{-\theta}$. There is a constant $C\ge1$ such that, for every $Q\ge0$,
\begin{equation}
\label{eq:uniform-characteristic-summation}
C^{-1}(1+Q)^{-(r-1)/\theta}
\le\sum_{j=1}^{\infty} b_j e^{-Q\rho_j}
\le C(1+Q)^{-(r-1)/\theta}.
\end{equation}
\end{lemma}

\begin{proof}
It suffices to consider $b_j=j^{-r}$ and $\rho_j=j^{-\theta}$. First consider $0\le Q\le1$. Then the $j=1$ term gives a positive constant lower bound, while
\[
\sum_{j=1}^{\infty} j^{-r}e^{-Qj^{-\theta}}
\le\sum_{j=1}^\infty j^{-r}<\infty.
\]
Also $(1+Q)^{-(r-1)/\theta}\eqsim1$ on this interval.

Now consider $Q>1$. Let $N=Q^{1/\theta}$. For the upper bound, split the sum at $N$. The tail satisfies
\[
\sum_{j>N} j^{-r}e^{-Qj^{-\theta}} \le \sum_{j>N}j^{-r} \le CN^{1-r}.
\]
For the part $j\le N$, divide the indices into dyadic blocks on which $j$ is comparable to $2^{-k}N$. The contribution of the $k$-th block is bounded by
\[
CN^{1-r} 2^{k(r-1)} e^{-2^{k\theta}}.
\]
The sum over $k\ge0$ is finite, so the contribution of $j\le N$ is also bounded by $CN^{1-r}=CQ^{-(r-1)/\theta}$.

For the lower bound, consider the integer interval $\lceil N\rceil\le j\le\lfloor3N\rfloor$, which contains at least $N$ indices. On this interval, $e^{-Qj^{-\theta}}\ge e^{-1}$ and $j^{-r}\ge(3N)^{-r}$. Therefore,
\[
\sum_{j=1}^{\infty} j^{-r}e^{-Qj^{-\theta}}
\ge e^{-1}N(3N)^{-r}
\gtrsim Q^{-(r-1)/\theta}.
\]
Since $Q\ge1$, this is equivalent to $(1+Q)^{-(r-1)/\theta}$.
\end{proof}

\subsection{Time rescaling and normalized modal dynamics}
\label{app:modal-reduction-uniform}

Let $a_j=\sqrt{\Sigma_{jj}}$ be the single-sample noise standard deviation, as in Section~\ref{sec:derivation_mechanism}. With $k_j=\sqrt B\,\lambda_j/a_j$, Theorem~\ref{thm:coordinatewise-sde-sign} gives
\begin{equation}
\label{eq:modal-sde-B-uniform}
\dd X_j(t) = -\frac1{\sqrt B}m\!\left(k_jX_j(t)\right)\dd t + \sqrt{n_B(t) \left[ 1-m\!\left(k_jX_j(t)\right)^2 \right]}\,\dd \widetilde W_{j,t},
\end{equation}
where $X_j(t)=(\vtheta_t)_j-\theta_j^*$, $X_j(0)=-\theta_j^*$, and $m(z)=2\Phi(z)-1$ is the smoothed sign function.

The initial signal energy in mode $j$ is
\[
w_j \coloneqq \lambda_jX_j(0)^2 = \lambda_j|\theta_j^*|^2.
\]
By Assumption~\ref{ass:power-law}, we have $ \lambda_j\eqsim j^{-q} $ and $ w_j\eqsim j^{-1-q s}. $ Tonelli's theorem gives
\begin{equation}
\label{eq:loss-modal-decomposition-uniform}
\EE[\cE(\vtheta_t)] = \frac12\sum_{j=1}^{\infty} \lambda_j\EE[X_j(t)^2].
\end{equation}

The Brownian motions in different coordinates may be correlated. Since $\bH$ is diagonal, \eqref{eq:loss-modal-decomposition-uniform} requires only the marginal second moments, and the scalar drift and diffusion coefficients in \eqref{eq:modal-sde-B-uniform} depend only on $X_j(t)$.

At intrinsic time $t$, the elapsed time for mode $j$ in rescaled units is
\[
\widehat t_j=\frac{k_jt}{\sqrt B}=\frac{\lambda_j}{a_j}\,t.
\]
Writing $\tau$ for the time variable in these units, define
\begin{equation}
\label{eq:Zj-B-def-uniform}
Z_j(\tau) \coloneqq k_j\, X_j\!\left(\frac{\sqrt B\,\tau}{k_j}\right).
\end{equation}
Then $Z_j(0)^2=B(\lambda_j/a_j^2)w_j$. Applying Lemma~\ref{lem:time-change-general} with the time change $\tau=k_jt/\sqrt B$, and then using the rescaling \eqref{eq:Zj-B-def-uniform}, we obtain
\begin{equation}
\label{eq:normalized-modal-sde-uniform}
\dd Z_j(\tau) = -m(Z_j(\tau))\,\dd\tau + \varepsilon_j(\tau) \sqrt{1-m(Z_j(\tau))^2}\, \dd\widetilde W_{j,\tau},
\end{equation}
where $\widetilde W_{j,\tau}$ is a standard Brownian motion in the rescaled time $\tau$, and
\begin{equation}
\label{eq:epsilon-j-def-uniform}
\varepsilon_j(\tau)^2 \coloneqq k_j\, \varphi\!\left( \frac{\sqrt B\,\tau}{k_j} \right)=\sqrt B\,k_j\,n_B\!\left(\frac{\sqrt B\,\tau}{k_j}\right).
\end{equation}

Choose fixed constants $\lambda_+,w_+,h_-,h_+>0$ such that
\[
\lambda_j\le\lambda_+j^{-q},
\qquad
w_j\le w_+j^{-1-q s},\qquad h_-\lambda_j\le a_j^2\le h_+\lambda_j.
\]
By the learning-rate condition in Theorem~\ref{thm:main-fsl-sign}, fix $C_0>0$ such that $\sqrt B\,\|\varphi\|_\infty=\sqrt B\,\eta_{\peak}\le C_0$. This gives the uniform normalized noise bound
\begin{equation}
\label{eq:uniform-noise-cap}
\varepsilon_j(\tau)^2
\le k_j\|\varphi\|_\infty
\le C_0\sqrt{\lambda_+/h_-}
\eqqcolon \varepsilon_{\max}^2.
\end{equation}
Here we used $k_j/\sqrt B=\lambda_j/a_j\le\sqrt{\lambda_+/h_-}$.

Returning to the original risk weights in \eqref{eq:loss-modal-decomposition-uniform} gives
\begin{equation}
\label{eq:excess-risk-Z-uniform}
\EE[\cE(\vtheta_t)]=\frac12\sum_{j=1}^{\infty}\frac{\lambda_j}{k_j^2}\EE[Z_j(\widehat t_j)^2],\qquad \frac{\lambda_j}{k_j^2}=\frac{a_j^2}{B\lambda_j}.
\end{equation}

\subsection{The generator and a global lower bound}
\label{app:uniform-moment-lower}

For $f\in C^2(\mathbb R)$, we can write the instantaneous generator of the SDE \eqref{eq:normalized-modal-sde-uniform} as
\begin{equation}
\label{eq:normalized-modal-generator-uniform}
(\mathcal L_{j,\tau}f)(z) = -m(z)f'(z) + \frac{\varepsilon_j(\tau)^2}{2} \left(1-m(z)^2\right)f''(z).
\end{equation}

Note that both $j$ and $\tau$ enter the instantaneous generator only through the scalar $\varepsilon_j(\tau)$. We therefore introduce the following operator, which provides a common representation of all the instantaneous generators:
\begin{equation}
\label{eq:normalized-generator-family-uniform}
(L_\varepsilon f)(z) \coloneqq -m(z)f'(z) + \frac{\varepsilon^2}{2} \left(1-m(z)^2\right)f''(z).
\end{equation}
We have
\[
\mathcal L_{j,\tau} = L_{\varepsilon_j(\tau)}.
\]

After the time rescaling, the drift $-m(z)$ is identical for every mode, while all dependence on the mode $j$ and on the learning-rate schedule is encoded in the single scalar $\varepsilon_j(\tau)$. It therefore suffices to establish drift estimates for a single generator $L_\varepsilon$, rather than treating each mode separately.

We use the properties of $m$ collected in Appendix~\ref{app:smoothed-sign-properties}, with $c=m'(0)$ as defined there. The local expansion \eqref{eq:smoothed-sign-local-bounds} identifies $c$ as the strength of the local linear drift, which enters both the lower and upper moment bounds. This local behavior motivates the quadratic test function $V(z)=z^2$. Since the drift saturates in the tails and the process can leave the neighborhood of the origin, the global upper bound uses a Lyapunov function that is quadratic near the origin and exponential outside a bounded interval.

\begin{lemma}[Quadratic drift lower bound]
\label{lem:uniform-second-moment-lower}
For $V(z)=z^2$, set
\begin{equation}
\label{eq:uniform-lower-rate}
\omega\coloneqq2c+c^2\varepsilon_{\max}^2.
\end{equation}
Then, for any $z\in\mathbb R$ and $0\le\varepsilon^2\le \varepsilon_{\max}^2$,
\begin{equation}
\label{eq:uniform-lower-drift}
L_\varepsilon V(z)\ge-\omega V(z)+\varepsilon^2.
\end{equation}
\end{lemma}

\begin{proof}
For the test function $V(z)=z^2$, the global bounds \eqref{eq:smoothed-sign-global-bounds} in Lemma~\ref{lem:smoothed-sign-properties} give
\[
\begin{aligned}
L_{\varepsilon}V(z)
&=-2zm(z)+\varepsilon^2\left(1-m(z)^2\right)\\
&\ge-\left(2c+c^2\varepsilon^2\right)z^2
+\varepsilon^2\\
&\ge-\omega V(z)+\varepsilon^2.
\end{aligned}
\]
\end{proof}

Applying the lower-bound conclusion of Lemma~\ref{lem:generator-dynkin} to this drift estimate gives the following second-moment bound.

\begin{corollary}[Global second-moment lower bound]
\label{cor:uniform-second-moment-lower}
Let $Z$ solve
\[
\dd Z(\tau)=-m(Z(\tau))\,\dd\tau
+\varepsilon(\tau)\sqrt{1-m(Z(\tau))^2}\,\dd W_\tau,
\qquad Z(0)=z_0,
\]
where $0\le\varepsilon(\tau)^2\le \varepsilon_{\max}^2$. With $\omega$ as in \eqref{eq:uniform-lower-rate}, for any $\tau\ge0$,
\begin{equation}
\label{eq:uniform-second-moment-lower}
\EE[Z(\tau)^2]\ge z_0^2e^{-\omega\tau}
+\int_0^\tau e^{-\omega(\tau-r)}\varepsilon(r)^2\,\dd r.
\end{equation}
\end{corollary}

\begin{proof}
Apply \eqref{eq:abstract-lyapunov-lower-bound} in Lemma~\ref{lem:generator-dynkin} with $V(z)=z^2$, $c_V=\omega$, and $r(\tau,z)=\varepsilon(\tau)^2$, using Lemma~\ref{lem:uniform-second-moment-lower}.
\end{proof}

\subsection{A Foster--Lyapunov upper bound}
\label{app:uniform-lyapunov}

We first construct a Lyapunov function, then use it to prove the second-moment upper bound.

\begin{lemma}[Auxiliary Lyapunov function]
\label{lem:uniform-lyapunov-function}
For any $R\ge1$, there exists a nonnegative even function $\Psi_R\in C^2(\mathbb R)$ with the following forms in the three regions:
\begin{enumerate}[leftmargin=2em]
\item For $|z|\le\tfrac12$, the function is quadratic:
\begin{equation}
\label{eq:uniform-lyapunov-pieces}
\Psi_R(z)=z^2.
\end{equation}
\item For $\tfrac12<|z|<R$, the function smoothly interpolates between the quadratic and exponential pieces and satisfies
\[
z^2\le\Psi_R(z)\le e^{|z|}.
\]
\item For $|z|\ge R$, the function is exponential:
\[
\Psi_R(z)=e^{|z|}.
\]
\end{enumerate}
In particular, the following bounds hold globally:
\begin{equation}
\label{eq:uniform-lyapunov-envelope}
z^2\le\Psi_R(z)\le4z^2e^{|z|},
\qquad z\in\mathbb R.
\end{equation}
\end{lemma}

\begin{proof}
For $0\le y\le1$, let
\[
h(y)=10y^3-15y^4+6y^5.
\]
Its first two derivatives are
\[
h'(y)=30y^2(1-y)^2, \qquad h''(y)=60y(1-y)(1-2y).
\]
In particular, we have $ h(0)=0, h(1)=1, $ and $ h'(0)=h'(1)=h''(0)=h''(1)=0. $ Since $h'(y)\ge0$, we also have $0\le h(y)\le1$ on $[0,1]$.

For $r\ge0$, define
\[
\psi_R(r)=
\begin{cases}
r^2, & 0\le r\le\tfrac12,\\[2mm]
(1-h(y))r^2+h(y)e^r,\quad y=\dfrac{r-\tfrac12}{R-\tfrac12},
& \tfrac12\le r\le R,\\[2mm]
e^r, & r\ge R,
\end{cases}
\qquad
\Psi_R(z)\coloneqq\psi_R(|z|).
\]
The definition gives the required quadratic and exponential forms. At $r=\tfrac12$ and $r=R$, the endpoint conditions on $h,h',h''$ make the values and the first two derivatives of the adjacent pieces agree. Thus $\psi_R$ is $C^2$ across both boundaries. Since $\Psi_R(z)=z^2$ near zero, the even extension $\Psi_R(z)=\psi_R(|z|)$ belongs to $C^2(\mathbb R)$. It is nonnegative because each piece is nonnegative.

On the connecting interval, $\psi_R(r)$ is a convex combination of $r^2$ and $e^r$. Since $e^r\ge r^2$ for $r\ge0$, we have
\[
r^2\le\psi_R(r)\le e^r\le e^R.
\]
Finally, the construction gives $\Psi_R(z)\ge z^2$ for any $z\in\mathbb R$. For $|z|\ge\tfrac12$, the connecting and tail regions satisfy
\[
\Psi_R(z)\le e^{|z|}\le4z^2e^{|z|};
\]
for $|z|\le\tfrac12$, $\Psi_R(z)=z^2\le4z^2e^{|z|}$. This proves \eqref{eq:uniform-lyapunov-envelope}.
\end{proof}

\begin{lemma}[Foster--Lyapunov drift estimate]
\label{lem:uniform-bounded-noise}
Let $\varepsilon_{\max}^2$ be the fixed noise bound in \eqref{eq:uniform-noise-cap}. There exist a Lyapunov function $\mathcal V$ of the form constructed in Lemma~\ref{lem:uniform-lyapunov-function} and constants $c_L>0$, $C\ge1$ such that
\[
L_\varepsilon\mathcal V(z) \le -c_L\mathcal V(z) + C\varepsilon^2
\]
for any $z\in\mathbb R$ and $0\le\varepsilon^2\le \varepsilon_{\max}^2$.
\end{lemma}

\begin{proof}
By the tail limits \eqref{eq:smoothed-sign-tail-limits} in Lemma~\ref{lem:smoothed-sign-properties}, we may choose $R\ge1$ sufficiently large for the bound $\varepsilon_{\max}^2$ in \eqref{eq:uniform-noise-cap} so that
\[
m(z)\operatorname{sgn}(z)\ge\frac34,
\qquad
\frac{\varepsilon_{\max}^2}{2}\left(1-m(z)^2\right)\le\frac14
,\qquad |z|\ge R.
\]
Set $\mathcal V=\Psi_R$, using the function constructed in Lemma~\ref{lem:uniform-lyapunov-function}. We estimate its generator in the local, tail, and connecting regions.

For $|z|\le\tfrac12$, we have $\mathcal V(z)=z^2$, so
\[
L_\varepsilon z^2 = -2z m(z)+\varepsilon^2(1-m(z)^2).
\]
Using the local lower bound \eqref{eq:smoothed-sign-local-bounds} in Lemma~\ref{lem:smoothed-sign-properties} and $1-m(z)^2\le1$, we obtain
\[
L_\varepsilon z^2\le-cz^2+\varepsilon^2.
\]
Thus the desired drift inequality holds in the local region.

On the tail region $|z|\ge R$, we have $\mathcal V(z)=e^{|z|}$. Hence
\[
\mathcal V'(z) = \operatorname{sgn}(z)\mathcal V(z), \qquad \mathcal V''(z)=\mathcal V(z).
\]
Therefore, for any $\varepsilon^2\le \varepsilon_{\max}^2$,
\[
\begin{aligned}
  L_\varepsilon\mathcal V(z)
  &=
  \mathcal V(z)
  \left[
    -m(z)\operatorname{sgn}(z)
    +
    \frac{\varepsilon^2}{2}(1-m(z)^2)
  \right]
  \le
  \mathcal V(z)
  \left[
    -\frac34+\frac14
  \right]
  =
  -\frac12\mathcal V(z).
\end{aligned}
\]
On the connecting region $\tfrac12\le|z|\le R$, write $r=|z|$ and $y=(r-\tfrac12)/(R-\tfrac12)$. Differentiating the interpolation in Lemma~\ref{lem:uniform-lyapunov-function} gives
\[
\mathcal V'(z)\operatorname{sgn}(z)
=(1-h(y))2r+h(y)e^r
+\frac{h'(y)}{R-\tfrac12}(e^r-r^2)\ge1,
\]
since $2r\ge1$, $e^r\ge1$, and $h'(y)\ge0$. Smoothness on this compact region gives $|\mathcal V''(z)|\le b_R$ for some finite $b_R>0$. By the oddness and strict monotonicity in Lemma~\ref{lem:smoothed-sign-properties}, we have $m(z)\operatorname{sgn}(z)\ge m_*\coloneqq m(\tfrac12)>0$ on this region, so
\[
\begin{aligned}
L_\varepsilon\mathcal V(z)
&=-m(z)\mathcal V'(z)
+\frac{\varepsilon^2}{2}(1-m(z)^2)\mathcal V''(z)\\
&\le-m_*+\frac{b_R}{2}\varepsilon^2.
\end{aligned}
\]
Choose
\[
c_L\coloneqq
\min\left\{c,\frac12,\frac{m_*}{2e^R}\right\}>0,
\qquad
C\coloneqq\max\left\{1,\frac{b_R}{2}\right\}.
\]
Because $\mathcal V(z)\le e^R$ on the connecting region, we have
\[
-m_*\le-c_L\mathcal V(z).
\]
Together with the local and tail estimates, these choices give
\[
L_\varepsilon\mathcal V(z)
\le-c_L\mathcal V(z)+C\varepsilon^2
\]
for any $z\in\mathbb R$ and $0\le\varepsilon^2\le \varepsilon_{\max}^2$.
\end{proof}

\begin{corollary}[Second-moment upper bound]
\label{cor:uniform-second-moment-upper}
Let $\mathcal V$, $c_L$, and $C$ be as in Lemma~\ref{lem:uniform-bounded-noise}. If $Z$ solves
\[
\,\mathrm d Z(\tau) = -m(Z(\tau))\,\mathrm d\tau + \varepsilon(\tau) \sqrt{1-m(Z(\tau))^2} \,\mathrm d W_\tau,
\]
with deterministic initial value $Z(0)$ and $\sup_{\tau\ge0}\varepsilon(\tau)^2\le \varepsilon_{\max}^2$, then
\[
\EE[Z(\tau)^2] \le C e^{-c_L\tau}\mathcal V(Z(0)) + C \int_0^\tau e^{-c_L(\tau-r)} \varepsilon(r)^2\,\mathrm d r.
\]
\end{corollary}

\begin{proof}
The bounded drift and diffusion in the scalar SDE give exponential moments of every order, uniformly on each finite time interval. This ensures integrability in Dynkin's formula for $\mathcal V$ and its derivatives. Using the drift estimate in Lemma~\ref{lem:uniform-bounded-noise}, the upper-bound conclusion \eqref{eq:abstract-lyapunov-bound} of Lemma~\ref{lem:generator-dynkin} with $V=\mathcal V$, $c_V=c_L$, and $r(\tau,z)=C\varepsilon(\tau)^2$ gives
\[
\EE[\mathcal V(Z(\tau))] \le e^{-c_L\tau}\mathcal V(Z(0)) + C \int_0^\tau e^{-c_L(\tau-r)} \varepsilon(r)^2\,\mathrm d r.
\]
The bound $z^2\le\mathcal V(z)$ in \eqref{eq:uniform-lyapunov-envelope} and $C\ge1$ give the claimed second-moment upper bound.
\end{proof}

\subsection{Control of the burn-in process}
\label{app:uniform-modal-comparison}

The following argument controls the burn-in phase for modes with large normalized initial values. Recall that $|Z_j(0)|=k_j|\theta_j^*|$. A large $|Z_j(0)|$ places the mode in the saturated region at initialization. Therefore, we require the intrinsic time $t$ to be sufficiently large relative to $\sqrt B\,|\theta_j^*|$ to go through this phase.

We compare the signal learning terms in Corollary~\ref{cor:uniform-second-moment-lower} and Corollary~\ref{cor:uniform-second-moment-upper}, which involve $Z_j(0)^2$ and $\mathcal V(Z_j(0))$, respectively. For modes with $|Z_j(0)|\le\tfrac12$, the identity $\mathcal V(Z_j(0))=Z_j(0)^2$ gives a matching initialization in both bounds. For modes with large $|Z_j(0)|$, the factor $\mathcal V(Z_j(0))$ cannot be bounded by $Z_j(0)^2$ up to a uniform constant, so we need to control the effect of burn-in. Each modal signal learning term decays exponentially in time, allowing a sufficiently large $t$ to suppress this extra factor. For these modes, the global envelope \eqref{eq:uniform-lyapunov-envelope} gives
\[
\mathcal V(Z_j(0))\le4Z_j(0)^2e^{|Z_j(0)|}.
\]
The next lemma uses \eqref{eq:uniform-absorption-conditions} to absorb $e^{|Z_j(0)|}$ into the exponential decay. To separate the modes according to their initial values, set $\zeta\coloneqq1+q s$ and define
\begin{equation}
\label{eq:uniform-initial-cutoff}
J_B\coloneqq
\max\left\{1,\left\lceil(4w_+B/h_-)^{1/\zeta}\right\rceil\right\}.
\end{equation}
The cutoff satisfies $J_B\eqsim B^{1/\zeta}=B^{1/(1+q s)}$. Its purpose is to separate two cases in the signal learning bound. For $j\ge J_B$, the bounds on $w_j$ and $\lambda_j/a_j^2$ give
\[
Z_j(0)^2=B(\lambda_j/a_j^2)w_j\le\tfrac14,
\qquad \mathcal V(Z_j(0))=Z_j(0)^2.
\]
These modes already have a matching initialization in both bounds. For $j<J_B$, the initial value may grow with $B$. The conditions in \eqref{eq:uniform-absorption-conditions} ensure that $t$ exceeds the burn-in scale for these modes by a sufficient constant factor, allowing the extra factor to be absorbed into the exponential decay in $t$.

\begin{lemma}[Burn-in control for small indices]
\label{lem:uniform-burn-in}
Suppose
\begin{equation}
\label{eq:uniform-absorption-conditions}
t\ge C_1\sqrt B,\qquad
t\ge C_3B^{1/(2s+2/q)},
\end{equation}
where $C_1,C_3>0$ are sufficiently large constants depending only on the fixed model parameters. Then, for any integer $1\le j<J_B$,
\begin{equation}
\label{eq:uniform-initial-absorption}
e^{-c_L k_jt/\sqrt B}\mathcal V(Z_j(0))
\le 4Z_j(0)^2 e^{-c_L k_jt/(2\sqrt B)}.
\end{equation}
\end{lemma}

\begin{proof}
Fix an integer $1\le j<J_B$. Applying the global envelope \eqref{eq:uniform-lyapunov-envelope} to the signal learning term gives
\[
e^{-c_L k_jt/\sqrt B}\mathcal V(Z_j(0))
\le4Z_j(0)^2e^{|Z_j(0)|-c_L k_jt/\sqrt B}.
\]
Our goal is to bound this expression by $4Z_j(0)^2e^{-c_L k_jt/(2\sqrt B)}$. It therefore suffices to show
\[
|Z_j(0)|\le \frac{c_L k_jt}{2\sqrt B}.
\]
Since $|Z_j(0)|=k_j|\theta_j^*|$ and $k_j>0$, this condition reduces to
\begin{equation}
\label{eq:uniform-absorption-target}
t\ge\frac{2\sqrt B\,|\theta_j^*|}{c_L}.
\end{equation}
The target \eqref{eq:uniform-absorption-target} compares the observation time with the burn-in scale $\sqrt B\,|\theta_j^*|$. We next estimate this scale for $j<J_B$. The power-law assumptions give
\[
|\theta_j^*|=\sqrt{\frac{w_j}{\lambda_j}}
\lesssim j^{(q-\zeta)/2}.
\]
The sign of $q-\zeta$ determines whether this bound grows with $j$. If $\zeta\ge q$, it is bounded by a constant $C_\theta$ independent of $B$. Taking $C_1\ge2C_\theta/c_L$, the first condition in \eqref{eq:uniform-absorption-conditions} gives
\[
t\ge C_1\sqrt B\ge\frac{2C_\theta\sqrt B}{c_L}\ge\frac{2\sqrt B\,|\theta_j^*|}{c_L}.
\]
If $\zeta<q$, the bound grows with $j$, so we use $j<J_B$ and $J_B\eqsim B^{1/\zeta}$ to obtain
\[
|\theta_j^*|\le C_\theta B^{(q-\zeta)/(2\zeta)}.
\]
Thus we need $t$ to dominate $\sqrt B$ times this polynomial in $B$. This is precisely what the second condition in \eqref{eq:uniform-absorption-conditions} ensures: choosing $C_3\ge2C_\theta/c_L$ gives
\[
t\ge C_3B^{q/(2\zeta)}
\ge\frac{2C_\theta\sqrt B}{c_L}B^{(q-\zeta)/(2\zeta)}
\ge\frac{2\sqrt B\,|\theta_j^*|}{c_L}.
\]
Thus, \eqref{eq:uniform-absorption-target} holds in both cases, which proves the claim.
\end{proof}

Combining this control of the burn-in process with the normalized second-moment estimates gives the following bounds, with signal learning terms that depend quadratically on $Z_j(0)$ on both sides.

\begin{proposition}[Uniform two-sided modal bounds]
\label{prop:uniform-modal-two-sided}
Under the condition \eqref{eq:uniform-absorption-conditions}, there is a constant $C\ge1$, independent of $B,t$ and the learning rate schedule, such that, for any $j\ge1$ and $\widehat t_j=k_jt/\sqrt B$,
\begin{equation}
\label{eq:uniform-modal-final-lower}
\EE[Z_j(\widehat t_j)^2]\ge Z_j(0)^2e^{-\omega\widehat t_j}
+\int_0^{\widehat t_j} e^{-\omega(\widehat t_j-r)}\varepsilon_j(r)^2\,\dd r,
\end{equation}
and
\begin{equation}
\label{eq:uniform-modal-final-upper}
\EE[Z_j(\widehat t_j)^2]\le C\left[
Z_j(0)^2e^{-c_L\widehat t_j/2}
+\int_0^{\widehat t_j} e^{-c_L(\widehat t_j-r)/2}\varepsilon_j(r)^2\,\dd r
\right].
\end{equation}
Here $\omega$ and $c_L$ are as in Lemma~\ref{lem:uniform-second-moment-lower} and Lemma~\ref{lem:uniform-bounded-noise}, respectively.
\end{proposition}

\begin{proof}
The lower bound follows directly from Corollary~\ref{cor:uniform-second-moment-lower}. For the upper bound, apply Corollary~\ref{cor:uniform-second-moment-upper} to $Z_j$. When $j<J_B$, Lemma~\ref{lem:uniform-burn-in} bounds the signal learning term by $Z_j(0)^2e^{-c_L\widehat t_j/2}$ up to a constant factor. When $j\ge J_B$, the identity $\mathcal V(Z_j(0))=Z_j(0)^2$ in \eqref{eq:uniform-lyapunov-pieces} gives the same bound. Since $\varepsilon_j(r)^2\ge0$, replacing $e^{-c_L(\widehat t_j-r)}$ by $e^{-c_L(\widehat t_j-r)/2}$ bounds the noise integral from above. These estimates give \eqref{eq:uniform-modal-final-upper}.
\end{proof}

\subsection{Completion of the proof}
\label{app:uniform-proof-completion}

Recall from \eqref{eq:excess-risk-Z-uniform} that
\[
\EE[\cE(\vtheta_t)]
=\frac12\sum_{j=1}^{\infty}\frac{\lambda_j}{k_j^2}\EE[Z_j(\widehat t_j)^2].
\]
We next prove the upper bound for the expected excess risk. The lower bound follows by the same argument. The key step is to sum the exponentially decaying modal terms and apply Lemma~\ref{lem:uniform-characteristic-summation} to obtain the power laws. Choose $C_1,C_3$ as in Lemma~\ref{lem:uniform-burn-in}. The training-time condition in Theorem~\ref{thm:main-fsl-sign} then implies \eqref{eq:uniform-absorption-conditions}, so Proposition~\ref{prop:uniform-modal-two-sided} applies.

For the signal term, the exact risk weight gives $(\lambda_j/k_j^2)Z_j(0)^2=w_j$. For the noise term, substitute \eqref{eq:epsilon-j-def-uniform} and change variables $r=k_ju/\sqrt B$. The factors $k_j$ cancel, giving
\[
\frac{\lambda_j}{k_j^2}\int_0^{\widehat t_j}e^{-c_L(\widehat t_j-r)/2}\varepsilon_j(r)^2\,\dd r
=\lambda_j\int_0^t e^{-c_L k_j(t-u)/(2\sqrt B)}n_B(u)\,\dd u.
\]
Substituting these identities into \eqref{eq:uniform-modal-final-upper} and summing over $j$, we obtain
\begin{equation}
\label{eq:uniform-risk-upper}
\begin{aligned}
\EE[\cE(\vtheta_t)]&\lesssim
\sum_{j=1}^{\infty} w_je^{-c_L k_jt/(2\sqrt B)}
+\int_0^t\left(\sum_{j=1}^{\infty}\lambda_je^{-c_L k_j(t-u)/(2\sqrt B)}\right)n_B(u)\,\dd u.
\end{aligned}
\end{equation}
Since $w_j\eqsim j^{-1-qs}$ and $k_j/\sqrt B=\lambda_j/a_j\eqsim\sqrt{\lambda_j}\eqsim j^{-q/2}$, Lemma~\ref{lem:uniform-characteristic-summation} with weights $w_j$, rates $k_j/\sqrt B$, and $Q=c_Lt/2$ gives
\begin{equation}
\label{eq:uniform-signal-power}
\sum_{j=1}^{\infty} w_je^{-c_L k_jt/(2\sqrt B)}\eqsim(1+c_Lt/2)^{-2s}\eqsim(1+t)^{-2s}.
\end{equation}
Here in the last step we used the fact that $c_L$ is a fixed positive constant. For the noise accumulation term, apply the same lemma with $b_j=\lambda_j$, the same rates $k_j/\sqrt B$, $r=q$, $\theta=q/2$, and $Q=c_L(t-u)/2$. This gives
\begin{equation}
\label{eq:uniform-kernel-power}
\sum_{j=1}^{\infty}\lambda_je^{-c_L k_j(t-u)/(2\sqrt B)}
\eqsim(1+t-u)^{-(2-2/q)},
\qquad 0\le u\le t.
\end{equation}
The noise accumulation estimate also holds at $u=t$, where $Q=0$. Substituting \eqref{eq:uniform-signal-power} and \eqref{eq:uniform-kernel-power} into \eqref{eq:uniform-risk-upper} gives the required upper bound.

For the lower bound, apply the same steps to \eqref{eq:uniform-modal-final-lower}, taking $Q=\omega t$ and $Q=\omega(t-u)$ in Lemma~\ref{lem:uniform-characteristic-summation}. Since $\omega$ is also a fixed positive constant, the same signal and noise weights $w_j$ and $\lambda_j$ give the same power laws. Together, the two bounds prove
\[
\EE[\cE(\vtheta_t)]
\eqsim
(1+t)^{-2s}
+\int_0^t
(1+t-r)^{-(2-2/q)}
n_B(r)\,\dd r.
\]
Since $n_B(r)=\sigma_B\varphi(r)$, this establishes the FSL~\eqref{eq:fsl-sign} with the bounded-time profiles
\[
S(t)\eqsim(1+t)^{-2s},\qquad K(t)\eqsim(1+t)^{-(2-2/q)}.
\]
These profiles have the large-time decay rates stated in Theorem~\ref{thm:main-fsl-sign}, while the shift by $1$ keeps the forgetting kernel finite at zero lag. All comparison constants are independent of $B,t$ and the particular schedule $\varphi$. This completes the proof of Theorem~\ref{thm:main-fsl-sign}.

\subsection{Discussion: Initialization and relaxation across spectral modes}
The initial cutoff in \eqref{eq:uniform-initial-cutoff} and the characteristic frontier from Section~\ref{subsec:deriving-fsl} satisfy
\[
J_B\eqsim B^{1/(1+qs)},\qquad j_t\eqsim t^{2/q}.
\]
Here $J_B$ separates modes by their initial normalized error, with $|Z_j(0)|\le1/2$ for $j\ge J_B$. The frontier $j_t$ compares the observation time with the characteristic relaxation time:
\[
\tau_j\eqsim\frac{\sqrt B}{k_j}\eqsim j^{q/2},\qquad \tau_{j_t}\eqsim t.
\]
Thus, $J_B$ is fixed at a given batch size, while $j_t$ moves to larger indices as training proceeds.

The bound $t\gtrsim B^{1/(2s+2/q)}$ in \eqref{eq:uniform-absorption-conditions} gives $J_B\lesssim j_t$. Burn-in control also uses $t\gtrsim\sqrt B$, which is the stronger bound when $s+1/q>1$.

\begin{figure}[H]
\centering
\captionsetup{skip=3pt}
\includegraphics[width=0.8\linewidth]{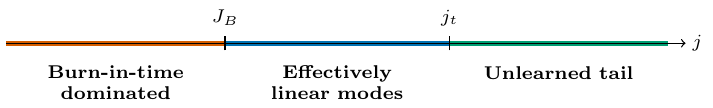}
\caption{\small \textbf{Regime separation for different modes when $J_B<j_t$.} The left region identifies modes that may require initial burn-in, controlled by Lemma~\ref{lem:uniform-burn-in}. The middle region contains modes with small initial normalized errors and relaxation times of order $t$ or shorter. The right region contains slower modes whose initial signal remains partly unlearned. The boundaries indicate characteristic scales.}
\label{fig:modal-index-regimes}
\end{figure}

For the ordering $J_B<j_t$ shown in \figref{fig:modal-index-regimes}, the three regions have the following meanings.
\begin{itemize}[leftmargin=*,topsep=3pt,itemsep=3pt,parsep=0pt]
\item \textbf{Burn-in-time dominated ($j<J_B$).} These modes may start with a large normalized error. When $|Z_j(0)|\gg1$, the saturated drift gives an initial burn-in scale of order $\sqrt B\,|\theta_j^*|$, which can exceed $\tau_j$. Lemma~\ref{lem:uniform-burn-in} controls the effect of this initial transient under the training-time requirement.
\item \textbf{Effectively linear modes ($J_B\le j<j_t$).} These modes start with $|Z_j(0)|\le1/2$ and have $\tau_j\lesssim t$. Their initial signal decays substantially when $j\ll j_t$, as reflected in the exponential terms of the modal bounds.
\item \textbf{Unlearned tail ($j>j_t$).} These modes have $\tau_j\gtrsim t$, so modes with $j\gg j_t$ retain most of their initial signal. Summing over this tail gives a signal contribution of order $j_t^{-qs}\eqsim t^{-2s}$.
\end{itemize}
These regions describe initial conditions and relaxation time scales. The ``unlearned'' label refers to the initial signal, while update noise contributes through the convolution term in the FSL.

\section{Data scaling analysis}
\label{app:data-scaling}

In this appendix, we analyze the data scaling behavior of SignSGD and SGD. Throughout, we fix the batch size at \(B=1\), so the number of iterations equals the data budget, \(N=D/B=D\), where \(D\) denotes the total number of samples.

At $B=1$, the two intrinsic clocks coincide: $\Gamma(z)\coloneqq\Gamma_1(z)=\Gamma_{\sgd}(z)=\int_0^z\eta(u)\,\dd u$. We use $\Gamma$ in the main-text data-scaling analysis. Similarly, $\varphi=\varphi_{\sgn}=\varphi_{\sgd}$ and $n_1(t)=\varphi(t)$. The final SignSGD risk is $\cE_D^{\sgn}=\EE[\cE(\vtheta_{\Gamma(D)})]$.

\subsection{Power decay schedules}
\label{app:power-decay}

The goal of the data scaling analysis is to characterize the best rate achievable by SignSGD or SGD through an appropriate choice of the learning rate schedule. To this end, we introduce the power decay schedule \citep{li2026optimal} and derive the rate it achieves in the following analysis. In Appendix~\ref{app:optimal-power-decay} we show that this schedule indeed gives the optimal rate. For $k\in\{0,1,\dots,N-1\}$, the schedule is defined as
\begin{equation}
\label{eq:app-power-decay}
    \eta_k
    =
    \eta_{\peak}\left(1-\frac{k}{N}\right)^p.
\end{equation}
It is determined by two hyperparameters: the peak learning rate \(\eta_{\peak}\) and the decay rate \(p>0\). In the continuous modeling, the power decay schedule takes the form
\[
    \eta(z)
    =
    \eta_{\peak}\left(1-\frac{z}{N}\right)^p,
\]
where $z\in[0,N]$ is the continuous training step. The intrinsic time is given by
\[
t = \Gamma(z)
=
\int_0^z \eta_{\peak}\left(1-\frac{u}{N}\right)^p
    \dd u
=
\frac{\eta_{\peak}N}{p+1}
\left[
    1-\left(1-\frac{z}{N}\right)^{p+1}
\right].
\]
Thus, the total intrinsic time of the entire training process is
\[
    T_D = \Gamma(D)
    =
    \frac{\eta_{\peak}D}{p+1}
    \eqsim
    \eta_{\peak}D.
\]
The learning rate schedule can be expressed in intrinsic time as
\begin{equation}
    \varphi(t)
    =
    \eta(\Gamma^{-1}(t))
    =
    \eta_{\peak}
    \left(1-\frac{t}{T_D}\right)^{\frac{p}{p+1}}.
    \label{eq: intrinsic power decay}
\end{equation}

\paragraph{Intuition behind the power decay schedule.}

We now provide an intuition for the choice of power decay. In our FSL, the loss is determined by two competing terms: the signal term, which prefers a large total intrinsic time, and the noise term, which prefers sufficiently fast decay near the end of training so that the accumulated noise can be forgotten. Since both SGD and SignSGD have power-law forgetting kernels (see Section~\ref{sec:functional_scaling_law_for_signsgd}), a power decay tail is the natural schedule shape for matching the kernel.

\subsection{Data scaling for SignSGD (Proof of Theorem \ref{thm:sign-data-scaling})}
\label{app:data-scaling-sign}

We first prove the data scaling result for SignSGD. Let $\cE^{\sgn}_D\coloneqq\EE[\cE(\vtheta_{T_D})]$ denote the final-step excess risk of SignSGD where \(T_D=\Gamma(D)\) is the total intrinsic time. The SignSGD FSL from Theorem~\ref{thm:main-fsl-sign} gives
\footnote{
Here and throughout the data-scaling analysis, we work in the
regime \(T_D\gg1\). Accordingly, for the signal learning term we have
\((1+T_D)^{-2s}\eqsim T_D^{-2s}\) for SignSGD and
\((1+T_D)^{-s}\eqsim T_D^{-s}\) for SGD, and we use the latter form to simplify the derivations.
}
\begin{equation}
\cE_D^{\sgn}
\eqsim
T_D^{-2s}
+
\int_0^{T_D}
(1+T_D-t)^{-(2-2/q)}
\varphi(t)\,\dd t.
\label{eq:app sign final loss}
\end{equation}
The following lemma simplifies the FSL under the power decay schedule. 

\begin{lemma}[SignSGD FSL under power decay schedules]
\label{lem:fsl-sign-power}
Under the assumptions of Theorem~\ref{thm:main-fsl-sign}, let $B=1$ and use the power-decay schedule~\eqref{eq: intrinsic power decay}. Let $\alpha_{\sgn}=\min\{\tfrac{q}{2},p+1\}$, where $p$ is the power decay exponent. Then the final excess risk satisfies
\begin{equation}
\label{eq:fsl-sign-power}
    \cE_D^{\sgn}
    \eqsim
    (\eta_{\peak}D)^{-2s}
    +
    \eta_{\peak}
    (\eta_{\peak}D)^{-(1-\frac{1}{\alpha_{\sgn}})}
\end{equation}
up to a logarithmic factor when
\(p+1=\tfrac{q}{2}\).
\end{lemma}

\begin{proof} 
For the power decay schedule, the total intrinsic time satisfies $T_D \eqsim \eta_{\peak} D$. By the SignSGD FSL, it remains to estimate the noise term \( \int_0^{T_D} (1+T_D-t)^{-(2-2/q)} \varphi(t)\,\dd t . \) Write \(u=T_D-t\). Then, under the power decay schedule \eqref{eq: intrinsic power decay}, \[ \varphi(T_D-u) = \eta_{\peak} \left(\frac{u}{T_D}\right)^{\frac{p}{p+1}} . \] Hence the noise term is \[ \eta_{\peak}T_D^{-\frac{p}{p+1}} \int_0^{T_D} (1+u)^{-(2-2/q)} u^{\frac{p}{p+1}} \dd u . \] Let \(a=\frac{p}{p+1}\). Since \( (1+u)^{-(2-2/q)}u^a \eqsim u^{a-(2-2/q)} \) for large $u$, we have \[ \int_0^{T_D} (1+u)^{-(2-2/q)} u^a \dd u \eqsim \begin{cases} 1, & a-(2-2/q)<-1,\\[2mm] \log T_D, & a-(2-2/q)=-1,\\[2mm] T_D^{a-(2-2/q)+1}, & a-(2-2/q)>-1. \end{cases} \] 
The boundary condition $a-(2-2/q) = -1$ is equivalent to $p+1=\frac{q}{2}$. 
Therefore, the noise term satisfies 
\[ \int_0^{T_D} (1+T_D-t)^{-(2-2/q)} \varphi(t)\,\dd t \eqsim \eta_{\peak} T_D^{-\left(1-\frac{1}{\alpha_{\sgn}}\right)}\]
up to a logarithmic factor at the boundary, where \(\alpha_{\sgn} = \min\{\frac{q}{2},p+1\}\). Hence we have \[ \cE_D^{\sgn} \eqsim (\eta_{\peak}D)^{-2s} + \eta_{\peak} (\eta_{\peak}D)^{-\left(1-\frac{1}{\alpha_{\sgn}}\right)} . \]
\end{proof}

Optimizing \eqref{eq:fsl-sign-power} over the peak learning rate gives the data-scaling result. The following theorem is a restatement of Theorem \ref{thm:sign-data-scaling}.

\begin{theorem}[Data scaling for SignSGD]
\label{thm:sign-data-scaling-appendix}
Under the assumptions of Theorem~\ref{thm:main-fsl-sign}, let $B=1$ and use the power-decay schedule~\eqref{eq: intrinsic power decay}. Then the following rates hold.
\begin{itemize}[leftmargin=2em]
    \item If \(s\ge \tfrac12-\tfrac1q\), then choosing
    $
        p>\tfrac{q}{2}-1,
        \eta_{\peak}
        \eqsim
        D^{-\frac{2s-1+2/q}{2s+2/q}}
    $
    yields
    \[
        \cE_D^{\sgn}
        \eqsim
        D^{-\frac{sq}{sq+1}}.
    \]
    \item If \(0<s<\tfrac12-\tfrac1q\), then choosing
    $
        p>\tfrac{2s}{1-2s},
        \eta_{\peak}\eqsim 1
    $
    yields
    \[
        \cE_D^{\sgn}
        \eqsim
        D^{-2s}.
    \]
\end{itemize}
\end{theorem}

\begin{proof}
First consider \(s\ge \tfrac12-\tfrac1q\). Choosing
\(p>\tfrac{q}{2}-1\) gives
\(\alpha_{\sgn}=\tfrac{q}{2}\). Then, applying Lemma~\ref{lem:fsl-sign-power}, we have
\[
    \cE_D^{\sgn}
    \eqsim
    (\eta_{\peak}D)^{-2s}
    +
    \eta_{\peak}(\eta_{\peak}D)^{-(1-2/q)}.
\]
Balancing the two terms yields
\[
    \eta_{\peak}
    \eqsim
    D^{-\frac{2s-1+2/q}{2s+2/q}}.
\]
Substituting this into the loss formula gives
\[
    \cE_D^{\sgn}
    \eqsim
    D^{-\frac{sq}{sq+1}}.
\]

If \(0<s<\tfrac12-\tfrac1q\), the choice of \(\eta_{\peak}\) above grows with \(D\), which violates the stability condition $\eta_{\peak} \lesssim 1$. We therefore take
\(\eta_{\peak}\eqsim1\). 
Choosing \(p>\tfrac{2s}{1-2s}\) gives $\alpha_{\sgn} = \min\{\tfrac{q}{2},p+1\} > \frac{1}{1-2s}$. 
The noise term is $o(D^{-2s})$, including the logarithmic factor when $p+1=q/2$. Hence
\[
    \cE_D^{\sgn}
    \eqsim
    D^{-2s}.
\]
\end{proof}

\subsection{Data scaling of SGD}
\label{app:data-scaling-sgd}

We now review the data scaling result for SGD. The SGD FSL yields the following achievable rates for power decay~\citep{li2026optimal}. Proposition~\ref{prop:sign-hard-achievable} supplies the matching lower bounds for all admissible schedules.

For SGD, let $\vtheta_t$ denote the process in~\eqref{eq:sde-sgd} and $\cE_D^{\sgd}\coloneqq\EE[\cE(\vtheta_{T_D})]$ its final expected excess risk at $T_D=\Gamma(D)$. The SGD FSL from \citet{li2025functional} gives
\begin{equation} 
\cE_D^{\rm sgd} \eqsim T_D^{-s} + \int_0^{T_D} (1+T_D-t)^{-(2-1/q)} \varphi(t)\,\dd t.
\label{eq:app sgd final loss}
\end{equation} 
The following lemma summarizes the SGD FSL under power decay schedules.

\begin{lemma}[SGD FSL under power decay schedules; \citet{li2026optimal}] 
\label{lem:fsl-sgd-power}
Consider the SGD FSL \eqref{eq:app sgd final loss} with the power decay schedule \eqref{eq: intrinsic power decay}. Let \( \alpha_{\rm sgd} = \min\{q,p+1\}, \) where \(p\) is the power decay exponent. Then the final excess risk satisfies \begin{equation} \label{eq:fsl-sgd-power} \cE_D^{\rm sgd} \eqsim (\eta_{\peak}D)^{-s} + \eta_{\peak} (\eta_{\peak}D)^{-\left(1-\frac{1}{\alpha_{\rm sgd}}\right)} \end{equation} up to a logarithmic factor when \(p+1=q\). \end{lemma}

\begin{proof} The proof is the same as that of Lemma~\ref{lem:fsl-sign-power}, with the SignSGD signal exponent \(2s\) replaced by the SGD signal exponent \(s\), and the effective capacity exponent \(q/2\) replaced by \(q\). Therefore, with \( \alpha_{\rm sgd} = \min\{q,p+1\}, \) the claimed result follows from the same calculation. \end{proof}

Optimizing \eqref{eq:fsl-sgd-power} over the peak learning rate gives the data-scaling result:

\begin{theorem}[Data scaling for SGD; \citet{li2026optimal}]
\label{thm:sgd-data-scaling}
Consider the intrinsic time SDE for SGD~\eqref{eq:sde-sgd} with the
power decay schedule~\eqref{eq: intrinsic power decay}. Then we have:
\begin{itemize}[leftmargin=2em]
    \item If \(s\ge 1-\tfrac1q\), then choosing
    $
        p>q-1,
        \eta_{\peak}
        \eqsim
        D^{-\frac{s-1+1/q}{s+1/q}}
    $
    yields
    \[
        \cE_D^{\rm sgd}
        \eqsim
        D^{-\frac{sq}{sq+1}}.
    \]
    \item If \(0<s<1-\tfrac1q\), then choosing
    $
        p>\tfrac{s}{1-s},
        \eta_{\peak}\eqsim 1
    $
    yields
    \[
        \cE_D^{\rm sgd}
        \eqsim
        D^{-s}.
    \]
\end{itemize}
\end{theorem}

\begin{proof}
First consider \(s\ge1-\tfrac1q\). Choosing \(p>q-1\) gives
\(\alpha_{\rm sgd}=q\), then applying Lemma \ref{lem:fsl-sgd-power}, we have
\[
    \cE_D^{\rm sgd}
    \eqsim
    (\eta_{\peak}D)^{-s}
    +
    \eta_{\peak}(\eta_{\peak}D)^{-(1-1/q)}.
\]
Balancing the signal and noise terms gives \(\eta_{\peak}\eqsim D^{-\frac{s-1+1/q}{s+1/q}}\) and the
minimax rate \(D^{-sq/(sq+1)}\).

Now consider \(0<s<1-\tfrac1q\), the choice of \(\eta_{\peak}\) above grows with \(D\), which violates the stability condition $\eta_{\peak} \lesssim 1$. We therefore take
\(\eta_{\peak}\eqsim1\). 
Choosing \(p>\tfrac{s}{1-s}\) gives $\alpha_{\sgd} = \min\{q,p+1\} > \frac{1}{1-s}$. 
The noise term is $o(D^{-s})$, including the logarithmic factor when $p+1=q$. Hence
\[
    \cE_D^{\sgd}
    \eqsim
    D^{-s}.
\]
\end{proof}

\subsection{Optimality of power decay schedules in data scaling}
\label{app:optimal-power-decay}

We compare all admissible learning rate schedules at the same sample budget $D$. A lower bound on accumulated noise limits the benefit of increasing the total intrinsic time, while the peak learning rate bounds the time available for learning the signal.

\begin{proposition}[Optimal rates over bounded learning rate schedules]
\label{prop:sign-hard-achievable}
Consider the SignSGD and SGD models of Appendix~\ref{app:data-scaling-sign} and Appendix~\ref{app:data-scaling-sgd} with $B=1$. Let $\mathcal A_D$ consist of all nonnegative schedules on $[0,D]$ bounded by a fixed admissible peak $C_0>0$, independent of $D$. Then
\[
\inf_{\eta\in\mathcal A_D}\cE_D^{\sgn}\eqsim D^{-\min\{2s,\,sq/(sq+1)\}},
\qquad
\inf_{\eta\in\mathcal A_D}\cE_D^{\sgd}\eqsim D^{-\min\{s,\,sq/(sq+1)\}}.
\]
\end{proposition}

\begin{proof}
Fix $\eta\in\mathcal A_D$ and write $T=\Gamma(D)$. For SignSGD, summing the global moment lower bound in Corollary~\ref{cor:uniform-second-moment-lower} by Lemma~\ref{lem:uniform-characteristic-summation}, as in Appendix~\ref{app:uniform-proof-completion}, bounds the risk below by the sum of the signal and noise terms for every $T\ge0$. For SGD the corresponding lower bound follows from~\eqref{eq:app sgd final loss} when $T\ge1$. For $T<1$, the first coordinate of the mean SGD error retains the factor $e^{-\lambda_1 T}$ from~\eqref{eq:sde-sgd}, so its risk is bounded below by a positive constant. The SignSGD signal lower bound gives the same conclusion. It suffices to consider $T\ge1$.

For a forgetting kernel $(1+u)^{-\beta}$, write the noise contribution in training steps. Cauchy--Schwarz on $[0,D]$ gives
\[
\begin{aligned}
\int_0^D(1+T-\Gamma(z))^{-\beta}\eta(z)^2\,\dd z
&\ge \frac{1}{D}\left[\int_0^D(1+T-\Gamma(z))^{-\beta/2}\eta(z)\,\dd z\right]^2\\
&=\frac{1}{D}\left[\int_0^T(1+u)^{-\beta/2}\,\dd u\right]^2.
\end{aligned}
\]
The equality uses $\Gamma'(z)=\eta(z)$ almost everywhere and also holds on schedules with zero learning rates, including entire intervals. With $\beta=2-2/q$ for SignSGD and $\beta=2-1/q$ for SGD, adding the signal lower bounds yields
\[
\cE_D^{\sgn}\gtrsim T^{-2s}+\frac{T^{2/q}}{D},
\qquad
\cE_D^{\sgd}\gtrsim T^{-s}+\frac{T^{1/q}}{D}.
\]
The two terms balance at $T=D^{1/(2s+2/q)}$ for SignSGD and at $T=D^{1/(s+1/q)}$ for SGD, giving a lower bound of order $D^{-sq/(sq+1)}$ in both cases. The cap gives $T\le C_0D$, so the signal terms also bound the respective risks below by constant multiples of $D^{-2s}$ and $D^{-s}$. Taking the larger of the two lower bounds for each optimizer gives the asserted exponents. The power decay schedules in Theorem~\ref{thm:sign-data-scaling-appendix} and Theorem~\ref{thm:sgd-data-scaling}, with the peak prefactors chosen to satisfy the cap, attain these rates.
\end{proof}

When SignSGD attains $D^{-sq/(sq+1)}$, its exponent agrees with the statistical minimax benchmark discussed in Section~\ref{subsec:data-scaling}. The proposition establishes optimality over schedules within the continuous models.

\section{Batch-size scaling analysis}
\label{app:critical-batch-size}
We prove Theorem~\ref{cor:batch-size-range} by comparing the intrinsic time needed to preserve the optimal small-batch risk with the time available at batch size $B$. This gives one argument for all task regimes in the theorem. Write $\alpha=s+1/q$ and $N=D/B$.

\begin{proof}
\textbf{The small-batch target.} Let $T=\Gamma_{\sgd}(N)$ and $U=\Gamma_B(N)$ denote the final intrinsic times of SGD and SignSGD. The data-scaling analysis in Appendix~\ref{app:data-scaling} identifies the optimal $B=1$ time scales
\[
T_*=\min\{D,D^{1/\alpha}\},\qquad
U_*=\min\{D,D^{1/(2\alpha)}\}.
\]
By Proposition~\ref{prop:sign-hard-achievable}, the corresponding optimal risks satisfy
\[
\cE_\star^{\sgd}(D,1)\eqsim T_*^{-s},\qquad
\cE_\star^{\sgn}(D,1)\eqsim U_*^{-2s}.
\]
Thus the question is how large $B$ can be while still reaching these time scales.

\textbf{The time available at batch size $B$.} Since there are only $N=D/B$ updates, the peak learning rate bounds $\eta_{\peak}^{\sgd}\lesssim1$ and $\sqrt B\,\eta_{\peak}^{\sgn}\lesssim1$ imply
\[
T\lesssim\frac DB,\qquad U\lesssim\frac DB.
\]
The signal lower bounds for SGD and Corollary~\ref{cor:uniform-second-moment-lower} for SignSGD hold for every schedule, including before burn-in. Consequently, for $1\le B\le D$,
\[
\cE_\star^{\sgd}(D,B)\gtrsim\left(\frac BD\right)^s,\qquad
\cE_\star^{\sgn}(D,B)\gtrsim\left(\frac BD\right)^{2s}.
\]
Comparing with the small-batch risks above, preserving a fixed risk ratio requires, respectively,
\[
B\lesssim\frac{D}{T_*}=\max\{1,D^{1-1/\alpha}\},\qquad
B\lesssim\frac{D}{U_*}=\max\{1,D^{1-1/(2\alpha)}\}.
\]
These are the claimed CBS upper bounds.

\textbf{Reaching the target times.} Use the power-decay schedule with fixed $p>q-1$ and the peak learning rates stated after Theorem~\ref{cor:batch-size-range}. Within the respective batch-size ranges above, they give $T\eqsim T_*$ and $U\eqsim U_*$. For SignSGD, we also need the training-time conditions
\[
U\gtrsim\sqrt B,\qquad U\gtrsim B^{1/(2\alpha)}.
\]
If $\alpha\le1/2$, the claimed batch range is bounded and $U\eqsim D$, so both conditions hold for large $D$. If $1/2<\alpha<3/2$, substituting $B\lesssim D^{1-1/(2\alpha)}$ and $U\eqsim D^{1/(2\alpha)}$ verifies both conditions. The power-decay estimates from Appendix~\ref{app:data-scaling} therefore give
\[
\begin{aligned}
\cE^{\sgd}(D,B;\eta)
&\lesssim T_*^{-s}+\frac{T_*^{1/q}}{D}\eqsim T_*^{-s},\\
\cE^{\sgn}(D,B;\eta)
&\lesssim U_*^{-2s}+\frac{U_*^{2/q}}{D}\eqsim U_*^{-2s},
\end{aligned}
\]
where $\cE^{\sgd}(D,B;\eta)$ and $\cE^{\sgn}(D,B;\eta)$ denote the final risks under these schedules. Hence both methods retain their respective small-batch risk throughout the claimed ranges. If the claimed CBS is of constant order, the matching lower bound already follows from $B_{\mathrm{crit}}(D)\ge1$.

\textbf{The fixed comparison factor.} Choose common constants $0<c_-\le C_+$ such that each optimizer's $B=1$ optimal risk is at least $c_-$ times its rate above, and the constructed schedules have risk at most $C_+$ times that rate. These constants do not depend on $D$ or $B$. For each optimizer, the constructed schedule therefore satisfies
\[
\frac{\cE(D,B;\eta)}{\cE_\star(D,1)}\le\frac{C_+}{c_-}
\]
throughout its claimed batch-size range. Fix the comparison factor $1+\delta$ in Definition~\ref{def:critical-batch-size} sufficiently large to absorb this ratio, independently of $D$ and $B$. The constructed schedules then give the CBS lower bounds for this fixed comparison factor. Together with the upper bounds, this proves~\eqref{eq:cbs-scaling}.
\end{proof}

\paragraph{The super-easy regime.}
Along the balanced SignSGD schedule, $U\eqsim D^{1/(2\alpha)}$. The burn-in requirement $U\gtrsim\sqrt B$ gives $B\lesssim D^{1/\alpha}$, while the peak learning rate cap gives $B\lesssim D^{1-1/(2\alpha)}$. At $s=3/2-1/q$ both bounds have order $D^{2/3}$, and the constants determine which one is reached first. For $s>3/2-1/q$, the burn-in condition restricts this schedule at a smaller batch scale. Beyond that scale the two-sided FSL for the balanced schedule no longer gives the optimal-risk comparison in Theorem~\ref{cor:batch-size-range}.

\section{Heuristic extension to RMSprop and Adam}
\label{app:adaptive-extension}

This appendix develops a heuristic extension of the SignSGD analysis to RMSprop and Adam. We work in the diagonal model and approximate the second-moment normalizer by the coordinate-wise gradient-noise variance after burn-in. For Adam, we additionally require a separation between its momentum memory and the coordinate relaxation time. These assumptions yield effective dynamics with the same FSL exponents and batch-size dependence as SignSGD. These results are supported by our experiments involving Adam and RMSprop in Appendix~\ref{app:experiments-toy}.

Ignoring the $\epsilon$ in the denominator and bias correction, the Adam update is
\[
\bmm_{k+1} = \beta_1 \bmm_k + (1-\beta_1)\bg_k,\; \bv_{k+1} = \beta_2 \bv_k + (1-\beta_2)\bg_k^2,\quad \btheta_{k+1} = \btheta_k - \eta_k \bmm_{k+1}/\sqrt{\bv_{k+1}},
\]
where all operations are coordinate-wise, with $\bmm_0=\bv_0=\bzero$. RMSprop corresponds to the case $\beta_1=0$. To expose the main mechanism, we first focus on RMSprop and then discuss Adam as a momentum-filtered variant.

\subsection{Second moments and coordinate curvatures}
Consider one coordinate with curvature $\lambda$ and write the stochastic gradient as $g_{k} = \mu_{k} + \xi_k$ with $\mu_k$ being the full-batch gradient. By Assumption~\ref{ass:curvature-aligned-noise}, in the label-noise-dominated regime, $\EE[\xi_k]=0$ and $\EE[\xi_k^2]\eqsim \sigma_B^2\lambda$, where $\sigma_B=1/\sqrt B$ is the batch noise scale from Section~\ref{sec:preliminaries}. For the following heuristic scaling calculation, we suppress fixed noise--curvature comparison constants and use $\sigma_B^2\lambda$ as a representative variance. In the noise-dominated regime: $\mu_k^2\ll \EE[\xi_k^2]$,  we have $v_{k+1}=(1-\beta_2)\sum_{j=0}^k\beta_2^{k-j}g_{j}^2\approx (1-\beta_2)\sum_{j=0}^k\beta_2^{k-j}\xi_j^2$. Approximating the noise across steps by independent draws with fixed variance, for $k(1-\beta_2)\gg 1$:
\[
\begin{aligned}
\EE[v_{k+1}] &\approx (1-\beta_2^{k+1})\sigma_B^2\lambda\approx \sigma_B^2\lambda,\\
\var[v_{k+1}] &\approx\frac{(1-\beta_2)(1-\beta_2^{2(k+1)})}{1+\beta_2}\var[\xi^2]\eqsim (1-\beta_2)\sigma_B^4\lambda^2,
\end{aligned}
\]
where we use $\EE[\xi^4]\lesssim (\EE[\xi^2])^2$ for Gaussian noise. Hence, $ v_{k+1} \eqsim \sigma_B^2\lambda \left[1+O_{\PP}\left(\sqrt{1-\beta_2}\right)\right]. $ When $\beta_2$ is close to $1$, the EMA self-averages and for $k(1-\beta_2)\gg 1$,
\[
v_{k+1}\eqsim \sigma_B^2\lambda.
\]
Thus, in the noise-dominated regime, RMSprop and Adam estimate a coordinate-wise gradient-noise variance proportional to curvature.

\subsection{The effective RMSprop dynamics}
Let $u_k = \theta_k - \theta^*$ be the  deviation along the coordinate with curvature $\lambda$. Using the representative normalization $\sqrt{v_{k+1}}\approx\sigma_B\sqrt{\lambda}$, the RMSprop update in the noise-dominated regime becomes
\begin{equation}\label{eqn: rmsprop-1}
u_{k+1} \approx u_{k} -\eta_k \frac{\lambda u_k+\xi_k}{\sigma_B\sqrt{\lambda}} = u_k - \eta_k\left(\sigma_B^{-1}\sqrt{\lambda}u_k + \zeta_k\right),
\end{equation}
where $\zeta_k=\xi_k/(\sigma_B\sqrt{\lambda})$ is the normalized noise satisfying $\EE[\zeta_k^2]\eqsim 1$. In the continuous model, let $u_j(t)$ be the $j$-th coordinate deviation at intrinsic time $t=\Gamma_B(z)=\sigma_B^{-1}\int_0^z\eta(v)\,\dd v$. We use $\varphi(t)=\varphi_{\sgn}(t)=\eta(\Gamma_B^{-1}(t))$ and $n_B(t)=\varphi(t)/\sqrt B$, as in the main text. With the fixed normalized noise-variance factor also suppressed, the representative effective dynamics become
\begin{equation}\label{eqn: coordiante-dynamics}
\dd u_j(t) = -\sqrt{\lambda_j}u_j(t)\dd t + \sqrt{\frac{\varphi(t)}{\sqrt B}}\dd W_j(t).
\end{equation}
Therefore, the second moment of the $j$-th coordinate is given by
\begin{equation}\label{eqn: coordiante-loss-dynamics}
\EE[u_j(t)^2] = u_j(0)^2e^{-2\sqrt{\lambda_j}t} + \frac1{\sqrt B}\int_0^t e^{-2\sqrt{\lambda_j}(t-t')}\varphi(t')\dd t'.
\end{equation}
\begin{remark}[Characteristic time]
The $j$-th coordinate dynamics has characteristic time $\tau_j\eqsim\lambda_j^{-1/2}$ in intrinsic time. In contrast, the characteristic time of SGD is $\lambda_j^{-1}$. This means the coordinate-wise normalization introduces acceleration by  compressing the characteristic time from $\lambda^{-1}$ to $\lambda^{-1/2}$, which is curvature adaptive. Batch size accelerates the intrinsic clock through the factor $\sqrt B$ in its definition.
\end{remark}

\subsection{The FSL of RMSprop}
We sum the coordinate-wise dynamics over the infinite power-law spectrum to identify the signal-learning and noise-forgetting profiles.

Multiplying~\eqref{eqn: coordiante-loss-dynamics} by $\lambda_j$ and summing over coordinates gives the dynamics of total loss $\frac12\EE[\bu_t^\top\bH\bu_t]$:
\begin{align}
    \frac12\EE[\bu_t^\top\bH\bu_t]\eqsim \sum_j\lambda_j u_j(0)^2e^{-2\sqrt{\lambda_j}t} + \frac1{\sqrt B}\int_0^t \sum_j \lambda_je^{-2\sqrt{\lambda_j}(t-t')}\varphi(t')\dd t'.
\end{align}
Hence, the loss dynamics of RMSprop admit the FSL with noise injection intensity $n_B(t)=\varphi(t)/\sqrt B$ and the signal learning and forgetting kernel given by
\[
S_{\rm rms}(t)=\sum_j\lambda_j u_j(0)^2e^{-2\sqrt{\lambda_j}t},\quad K_{\rm rms}(t) = \sum_j \lambda_je^{-2\sqrt{\lambda_j}t}.
\]
\paragraph{The origin of power-law scaling.} Under the power-law assumption (Assumption~\ref{ass:power-law}): $\lambda_j\eqsim j^{-q}$ and $w_j\coloneqq\lambda_j u_j(0)^2\eqsim j^{-(1+sq)}$,  we have
\[
S_{\rm rms}(t)\eqsim (1+t)^{-2s},\qquad K_{\rm rms}(t)\eqsim (1+t)^{-(2-2/q)}.
\]
We now provide a physical-style derivation of why they follow the above power-law decay. The coordinate with curvature $\lambda_j$ has characteristic time $\tau_j\eqsim\lambda_j^{-1/2}$. Take the signal-learning term as an example, $S(t)=\sum_j w_j e^{-2\sqrt{\lambda_j}t}$. At time $t$, the modes with $\tau_j\lesssim t$ have essentially been learned, while the modes with $\tau_j\gtrsim t$ are still largely unlearned. Since $\lambda_j\eqsim j^{-q}$, the moving frontier $j_t$ is determined by $\tau_{j_t}\eqsim t$, yielding $j_t\eqsim t^{2/q}$, as in Section~\ref{subsec:deriving-fsl}. For $t\gtrsim1$,
\begin{align}
S_{\rm rms}(t) &= \sum_{j\le j_t}w_j \underbrace{e^{-2\sqrt{\lambda_j}t}}_{\approx 0} + \sum_{j>j_t}w_j \underbrace{e^{-2\sqrt{\lambda_j}t}}_{\approx 1} \approx \sum_{j>j_t}w_j \eqsim (j_t)^{-sq} \eqsim t^{-2s}.
\end{align}
\Needspace{14\baselineskip}
\begingroup
\setlength{\intextsep}{3pt}
\begin{wrapfigure}[12]{r}{0.28\textwidth}
    \centering
    \includegraphics[width=\linewidth]{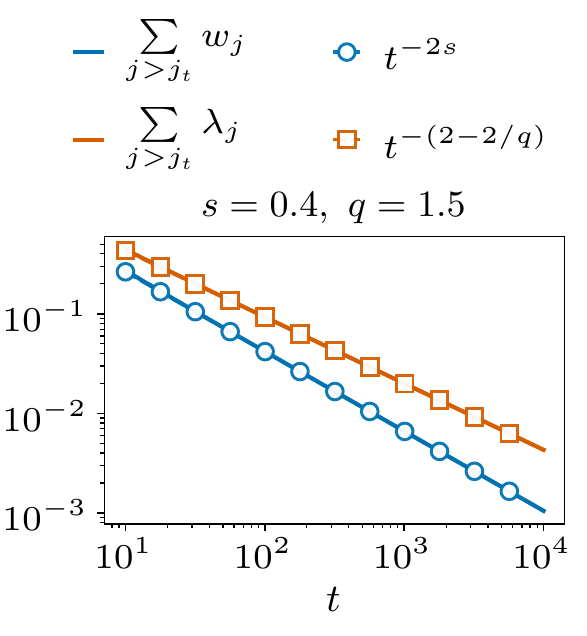}
    \captionsetup{font=small,skip=3pt,width=\linewidth,justification=centering}
    \caption{\small Tail-sum scaling.}
    \label{fig:tail-sum-scaling}
\end{wrapfigure}
The power-law scaling of the forgetting kernel follows from the same moving-cutoff argument.
\begin{remark}[Tail dominates scaling law]
The above derivation suggests a simple picture of learning in power-law data. The expected excess risk is a sum over spectral modes, and each mode has its own characteristic time $\tau_j$. At time $t$, the moving frontier $j_t$ is determined by $ \tau_{j_t}\eqsim t . $ Modes with $\tau_j\lesssim t$ have already been learned and contribute little to the remaining signal error, whereas modes with $\tau_j\gtrsim t$ are still largely unlearned. Thus the signal-learning term is essentially the tail mass of the unlearned modes. Analogously, the power-law decay of $S(t)$ therefore comes from how fast this moving frontier advances through the spectrum. In summary,
\begin{equation}
S(t)\approx \sum_{j\gtrsim j_t}w_j, \qquad K(t)\approx \sum_{j\gtrsim j_t}\lambda_j,\quad\quad \text{with } \tau_{j_t}\eqsim t.
\end{equation}
\figref{fig:tail-sum-scaling} demonstrates that the above moving-tail approximation indeed yields the right scaling for both signal-learning and noise-forgetting terms.
\end{remark}
\endgroup

\subsection{The FSL of Adam}
Adam differs from RMSprop only through the first-moment accumulator in the numerator. For a complementary analysis of how momentum affects data scaling and the critical batch size in one-pass kernel regression, see \citet{wang2026momentum}. Under the same denominator approximation $v_{k+1}\approx \sigma_B^2\lambda$, the first-moment EMA has an effective memory window $M_1\eqsim(1-\beta_1)^{-1}$ steps. After burn-in, $k\gg M_1$, its expansion is
\[
m_{k+1}=(1-\beta_1)\sum_{j=0}^{k}\beta_1^j\mu_{k-j}+(1-\beta_1)\sum_{j=0}^{k}\beta_1^j\xi_{k-j}.
\]
Assume that the learning rate varies little across this window and that $\sqrt B\,\eta_kM_1\ll\tau_\lambda$, where $\tau_\lambda\eqsim\lambda^{-1/2}$ is the intrinsic relaxation time. Recall from~\eqref{eqn: rmsprop-1} that $\zeta_k=\xi_k/(\sigma_B\sqrt{\lambda})$ is the normalized gradient noise. Its first-moment EMA, denoted by $\widetilde\zeta_k$, gives the noise contribution to the normalized Adam update. The drift contribution is then approximately {\it quasi-static}, giving
\begin{equation}\label{eqn: quasi-static}
u_{k+1}\approx u_k-\eta_k\left(\sigma_B^{-1}\sqrt{\lambda}u_k+\widetilde\zeta_k\right),\qquad \widetilde\zeta_k\coloneqq(1-\beta_1)\sum_{j=0}^{k}\beta_1^j\zeta_{k-j}.
\end{equation}
Assume that the normalized noises $\zeta_k$ are independent and identically distributed, with mean zero and variance $\sigma_\zeta^2\coloneqq\var[\zeta_k]\eqsim1$. In the stationary approximation after burn-in,
\[
\var[\widetilde\zeta_k]\approx (1-\beta_1)^2\sum_{j=0}^{\infty}\beta_1^{2j}\sigma_\zeta^2=\frac{1-\beta_1}{1+\beta_1}\sigma_\zeta^2,\quad \operatorname{Cov}(\widetilde\zeta_k,\widetilde\zeta_{k+h})\approx\beta_1^h\var[\widetilde\zeta_k],\quad h\geq0.
\]
The first equation gives the single-step variance reduction. The second equation shows that different steps remain positively correlated because their moving averages share past noise. For a block of $L\gg M_1$ steps, the variance of the accumulated noise therefore satisfies
\[
\frac1L\var\!\left[\sum_{r=0}^{L-1}\widetilde\zeta_{k+r}\right]\approx \sigma_\zeta^2\frac{1-\beta_1}{1+\beta_1}\left[1+2\sum_{h=1}^{L-1}\left(1-\frac hL\right)\beta_1^h\right]\approx \sigma_\zeta^2\frac{1-\beta_1}{1+\beta_1}\left(1+\frac{2\beta_1}{1-\beta_1}\right)=\sigma_\zeta^2.
\]
Momentum spreads the noise of each step across its memory window. For $L\gg M_1$, these contributions accumulate within the block, giving the same leading noise variance $L\sigma_\zeta^2$ as the unfiltered noise despite smoother individual updates.

Now choose a block with $M_1\ll L\ll\tau_\lambda/(\sqrt B\,\eta_k)$, and assume that the learning rate remains approximately constant over this block. The intrinsic time elapsed over this block and the accumulated parameter-noise variance are
\[
\Delta t\approx\sqrt B\,L\eta_k,\qquad
\var\!\left[\sum_{r=0}^{L-1}\eta_{k+r}\widetilde\zeta_{k+r}\right]\approx L\eta_k^2\sigma_\zeta^2\approx\frac{\sigma_\zeta^2\eta_k}{\sqrt B}\Delta t.
\]
Together with the drift in~\eqref{eqn: quasi-static}, this yields the effective intrinsic-time SDE
\[
\dd u(t)=-\sqrt{\lambda}u(t)\,\dd t+\sqrt{\frac{\sigma_\zeta^2\varphi(t)}{\sqrt B}}\,\dd W_t.
\]
This is the same form as the RMSprop model~\eqref{eqn: coordiante-dynamics}, with $\sigma_\zeta^2\eqsim1$ absorbed into the constants. Therefore, following the same argument as for RMSprop, we obtain the FSL for Adam with noise injection intensity $n(t)\eqsim n_B(t)=\varphi(t)/\sqrt B$ and
\begin{equation}
S_{\rm adam}(t)\eqsim(1+t)^{-2s},\qquad K_{\rm adam}(t)\eqsim(1+t)^{-(2-2/q)}.
\end{equation}

\section{Experimental details and additional results}
\label{app:additional-experiments}

\subsection{Fitting and prediction using FSL}
\label{app:fsl-detail}

Although derived from a kernel regression model, the FSL framework is expected to model loss dynamics in more general training scenarios, as the decomposition into ``irreducible, signal, and noise'' components carries a broader physical interpretation. Inspired by the functional form of \eqref{eq:fsl-sign}, we heuristically propose a parametrized ansatz:
\begin{equation}\label{eqn: fsl-parametrized-app}
\begin{aligned}
L_{\Theta}(t;\varphi, B) ={}& \underbrace{L_\infty}_{\text{irreducible loss}} + \underbrace{C_S(A_S+t)^{-\alpha}}_{\text{signal-learning term}}\\
&+\underbrace{\frac{C_K}{\sqrt B}\int_0^t \bigl(A_K+t-r\bigr)^{-\nu}\varphi(r)\,\dd r}_{\text{noise-forgetting term}},
\end{aligned}
\end{equation}
Here $t=\Gamma_B(z)=\sqrt B\int_0^z\eta(u)\,\dd u$ denotes intrinsic time, $B$ is the batch size, $\varphi(t)=\eta(\Gamma_B^{-1}(t))$ is the learning rate schedule in intrinsic time, and $\Theta=(L_\infty,C_S,A_S,\alpha,C_K,A_K,\nu)$ denotes fitting parameters. Thus $\varphi=\varphi_{\sgn}$ and $n_B(t)=\varphi(t)/\sqrt B$. The noise-forgetting term in \eqref{eqn: fsl-parametrized-app} is equivalently $C_K\int_0^t(A_K+t-r)^{-\nu}n_B(r)\,\dd r$. We use this base form for synthetic experiments and extend its learning-rate dependence for language model training as described in Appendix~\ref{app:llm}.

\paragraph{Discrete FSL.}

To obtain the left-endpoint form used in the main text, define
\[
\begin{aligned}
t_k&=\sqrt B\sum_{j=0}^{k-1}\eta_j,\qquad I_j=[t_j,t_{j+1}),\\
\Delta t_j&=t_{j+1}-t_j=\sqrt B\,\eta_j.
\end{aligned}
\]
On each nonempty interval $I_j$, take $\varphi(r)=\eta_j$. The left-endpoint rectangle rule gives
\[
\begin{aligned}
\frac1{\sqrt B}\int_0^{t_k}K(t_k-r)\varphi(r)\,\dd r
&\approx\frac1{\sqrt B}\sum_{j=0}^{k-1}K(t_k-t_j)\varphi(t_j)\Delta t_j\\
&=\sum_{j=0}^{k-1}K(t_k-t_j)\eta_j^2.
\end{aligned}
\]
Steps with $\eta_j=0$ contribute zero. With $K(u)=(A_K+u)^{-\nu}$, this yields the parameterized form in~\eqref{eqn:fsl-parametrized-discrete}.

\paragraph{Numerical convention used in the reported fits.}

We discretize the noise integral on the cumulative-learning-rate grid
\[
\tau_k=\sum_{j=0}^{k-1}\eta_j,\qquad
\tau_0=0,\qquad
\Delta\tau_j=\tau_{j+1}-\tau_j=\eta_j.
\]
Here $k$ is the number of completed updates, so $\ell_k$ is the loss at $\btheta_k$ and $t_k=\sqrt B\,\tau_k$ is the same intrinsic clock as in the main text. The first stored loss corresponds to $k=1$. Thus the signal term is $C_S(A_S+\sqrt B\,\tau_k)^{-\alpha}$.

To discretize the noise integral in \eqref{eqn: fsl-parametrized-app}, set $r=\sqrt B\,v$ and write $\psi(v)=\varphi(\sqrt B\,v)$ for the learning rate expressed in cumulative-learning-rate time. Then the noise term at $t_k$ is
\[
\begin{aligned}
\frac{C_K}{\sqrt B}\int_0^{t_k}
K(t_k-r)\varphi(r)\,\dd r
&=C_K\int_0^{\tau_k}
K\!\left(\sqrt B(\tau_k-v)\right)\psi(v)\,\dd v,
\end{aligned}
\]
where $K(u)=(A_K+u)^{-\nu}$. On each interval $I_j=(\tau_j,\tau_{j+1}]$, take $\psi(v)=\eta_j$. The right-endpoint rectangle rule gives
\[
\begin{aligned}
\int_0^{\tau_k}K\!\left(\sqrt B(\tau_k-v)\right)\psi(v)\,\dd v
&\approx\sum_{j=0}^{k-1}
K\!\left(\sqrt B(\tau_k-\tau_{j+1})\right)\eta_j\Delta\tau_j\\
&=\sum_{j=0}^{k-1}
K\!\left(\sqrt B(\tau_k-\tau_{j+1})\right)\eta_j^2.
\end{aligned}
\]
Steps with $\eta_j=0$ have zero interval length and contribute zero. The resulting predictor is
\begin{equation}\label{eqn:fsl-discrete}
\begin{aligned}
\widehat L_{\Theta}(k;\bfeta,B)
={}&L_\infty+C_S(A_S+\sqrt B\,\tau_k)^{-\alpha}\\
&+C_K\sum_{j=0}^{k-1}
\bigl(A_K+\sqrt B(\tau_k-\tau_{j+1})\bigr)^{-\nu}\eta_j^2.
\end{aligned}
\end{equation}
The discretization step is $\Delta\tau_j=\eta_j$. Batch size enters the signal and forgetting arguments through $\sqrt B$. Equation~\eqref{eqn:fsl-discrete} records the right-endpoint convention used for the reported fits and predictions. The main-text expression~\eqref{eqn:fsl-parametrized-discrete} uses left endpoints. Both discretize the continuous FSL~\eqref{eqn: fsl-parametrized-app}, but they need not agree at finite step sizes.

\paragraph{Fitting the discrete FSL.}

For a loss trajectory with learning rate schedule $\bfeta=(\eta_0,\ldots,\eta_{N-1})$, batch size $B$, and losses $(\ell_1,\ldots,\ell_N)$, the parameters $\Theta$ are estimated by minimizing the mean squared error (MSE):
\begin{equation}
\min_{\Theta}\; \frac{1}{|\mathcal I|} \sum_{k\in\mathcal I} \bigl(\widehat L_{\Theta}(k;\bfeta,B)-\ell_k\bigr)^2,
\label{eq:fit-loss}
\end{equation}
where $\mathcal I=\{N_{\mathrm{skip}}+1,\ldots,N\}$ denotes the set of step indices used for fitting (the first $N_{\mathrm{skip}}$ steps are excluded). We optimize $\Theta$ using a two-stage procedure: 2000 steps of Adam followed by 30 steps of L-BFGS. The optimizer configurations are shown in \figref{lst:fsl-optimizer-config}.

\begin{figure}[t]
\centering
\small

\begin{minipage}[t]{0.48\linewidth}

\begin{lstlisting}[style=cleanpython,basicstyle=\ttfamily\small]
adam_optimizer = torch.optim.Adam(
    model.parameters(),
    lr=0.05,
    weight_decay=1e-6,
    betas=(0.9, 0.95),
    eps=1e-8,
)
\end{lstlisting}
\end{minipage}
\hfill
\begin{minipage}[t]{0.48\linewidth}

\begin{lstlisting}[style=cleanpython,basicstyle=\ttfamily\small]
lbfgs_optimizer = torch.optim.LBFGS(
    model.parameters(),
    lr=0.1,
    max_iter=50,
    tolerance_grad=1e-20,
    tolerance_change=1e-20,
    history_size=100,
    line_search_fn="strong_wolfe",
)
\end{lstlisting}
\end{minipage}

\caption{\small Optimizer configurations used in the two-stage FSL fitting procedure. \textbf{(Left)} Stage 1: Adam. \textbf{(Right)} Stage 2: L-BFGS.}
\label{lst:fsl-optimizer-config}
\end{figure}

  In some experiments, we fit multiple loss trajectories simultaneously. The fitting objective is the average MSE:
  \[
\min_{\Theta} \frac{1}{M}\sum_{m=1}^M \frac{1}{|\mathcal I^{(m)}|} \sum_{k\in\mathcal I^{(m)}} \bigl(\widehat L_\Theta(k;\bfeta^{(m)},B^{(m)})-\ell_k^{(m)}\bigr)^2,
  \]
  where $m \in \{1,\cdots,M\}$ denotes the index of the loss trajectory. Each trajectory uses its own cumulative-learning-rate grid $\tau_k^{(m)}=\sum_{j=0}^{k-1}\eta_j^{(m)}$. Its intrinsic time is $\sqrt{B^{(m)}}\tau_k^{(m)}$.

After fitting the parameters $\Theta$, we use the discrete FSL \eqref{eqn:fsl-discrete} to predict loss trajectories for held-out learning rate schedules or batch sizes.

\subsection{PLK regression experiments}
\label{app:experiments-toy}

The loss-curve and data-scaling experiments in Section~\ref{subsec:synthetic} use $(s,q)=(0.3,4)$, which satisfies
\[
\frac12-\frac1{q} \le s < 1-\frac1{q} .
\]
As discussed in Section~\ref{subsec:data-scaling}, this is the regime where SignSGD attains the statistically optimal data scaling rate, whereas SGD does not.

In all experiments, we set dimension $d=512$, eigenvalues $\lambda_j = j^{-q}$, and target magnitudes $|\theta_j^*| = j^{-\frac12 (sq-q+1)}$ for $j=1,\ldots,d$. We fix the noise variance $\sigma^2=1.0$, and initialize at $\btheta_0 =\bzero$. Unless otherwise stated, the default peak learning rate and batch size are $\eta_{\peak} = 0.1$ and $B=1$, respectively. We report the population excess risk after $k$ updates,
\[
\ell_k=\mathcal E(\btheta_k)=\frac12\sum_{j=1}^d \lambda_j(\theta_{k,j}-\theta_j^*)^2 .
\]

Across the synthetic experiments, we use SGD, SignSGD, RMSprop and Adam; the optimizer subset depends on the experiment. For RMSprop, we set $\beta_2=0.9$ and $\epsilon=10^{-8}$. For Adam, we set $(\beta_1,\beta_2)=(0.9,0.95)$ and $\epsilon=10^{-8}$. For the loss-curve and data-scaling experiments, we report the excess risk averaged over 200 independent runs unless otherwise stated.

\subsubsection{Risk dynamics across learning rate schedules}
\label{app:plk-risk-dynamics}

\begin{figure}[!ht]
  \centering
  \captionsetup{skip=3pt}
  \includegraphics[width=0.48\linewidth]{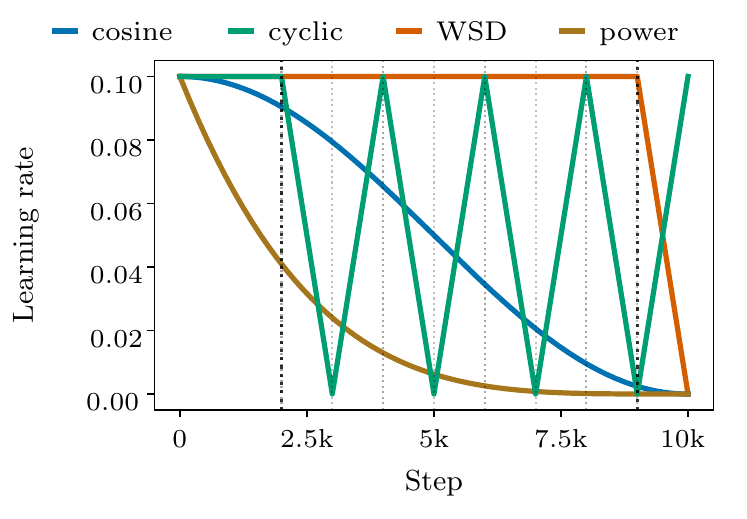}
  \caption{\small Learning rate schedules in synthetic experiments.}
  \label{fig:toy lrs}
\end{figure}

\paragraph{Fitting and prediction for different schedules.}

We provide experimental details for risk-curve fitting and prediction. For each optimizer, we fit the FSL on cosine~\citep{loshchilov2017sgdr} and WSD~\citep{hu2024minicpm} schedules and evaluate its predictive performance on the held-out cyclic schedule~\citep{smith2017cyclical}. See \figref{fig:toy lrs} for a visualization of these schedules. For SignSGD in \figref[left]{fig:story-overview}, we set $(s,q,d)=(0.3,4,512)$, $B=1$, and $\sigma^2=1$, initialize at zero, and train for $10{,}000$ updates with peak learning rate $0.1$. For each schedule, we report the unsmoothed population excess risk averaged over $200$ independent runs. We fit six shared parameters in \eqref{eqn:fsl-discrete} and keep them fixed for the held-out cyclic prediction. We use the fitting procedure of Appendix~\ref{app:fsl-detail}, consisting of $2000$ Adam steps followed by $30$ L-BFGS steps, with $L_\infty=0$ for excess risk and $N_{\text{skip}}=500$ unless specified otherwise.

For Adam in \figref[left]{fig:adaptive-extension}, we set $N_{\text{skip}}=1000$ and fit piecewise-linear interpolants of the risk curves at zero-based stored indices $1000$ through $9999$.

\paragraph{Decomposition and intrinsic-time collapse across schedules.}
\figref{fig:intrinsic-time-collapse} uses the same experimental settings and fitted FSL parameters as \figref[left]{fig:story-overview}. \figref[a]{fig:intrinsic-time-collapse} shows the signal-learning and noise-forgetting terms for the three LR schedules. \figref[b]{fig:intrinsic-time-collapse} plots the signal-learning terms against intrinsic time $t_k=\sqrt B\,\tau_k$, showing their collapse across schedules.

\subsubsection{Data scaling and critical batch size}
\label{app:plk-scaling}

\paragraph{Data scaling.}

We provide details for the SGD/SignSGD data-scaling comparison in \figref[middle]{fig:story-overview} and the adaptive-optimizer comparison in \figref[left]{fig:adaptive-scaling}. Across these comparisons, we run SGD, SignSGD, RMSprop, and Adam using the power decay schedule~\citep{li2026optimal}:
\[
\eta_k=\eta_{\peak}\left(1-\frac{k}{N}\right)^{p}, \qquad k=0,1,\dots,N-1.
\]
Here $N$ is the number of updates and $D=BN$. We choose $p=4$, which satisfies the power-decay conditions in Theorem~\ref{thm:sign-data-scaling} for SignSGD and in Appendix~\ref{app:data-scaling-sgd} for SGD at the experimental parameters. We use the same schedule family for RMSprop and Adam to compare their data-scaling behavior.

All runs use batch size $B=1$, with $\eta_{\peak}$ tuned at each budget. We use $D\in\{1000,2000,4000,8000,16000,32000\}$, evaluate SignSGD on 64 independent runs after tuning, and compute SGD risk by exact second-moment recursion. The four-optimizer comparison in \figref[left]{fig:adaptive-scaling} uses the five budgets from 2000 to 32000, gradient mean $2\bH(\btheta-\btheta^*)$, the same noise covariance, and 200 evaluation runs per setting.

\paragraph{Critical batch size.}\leavevmode
\figref[right]{fig:story-overview} compares SGD and SignSGD. \figref[right]{fig:adaptive-scaling} additionally reports Adam and RMSprop. For these batch-size experiments, we use the common PLK setup with $(s,q)=(1,4)$. We use the budgets and schedule
\[
\begin{aligned}
D&\in\{2000,4000,8000,16000,32000,64000,128000\},\qquad N=\lceil D/B\rceil,\\
\eta_k&=\eta_{\peak}(1-k/N)^4,\qquad k=0,\ldots,N-1.
\end{aligned}
\]
The final batch consumes the remaining samples, so every run uses exactly $D$ samples. For every optimizer and $(D,B)$, we tune $\eta_{\peak}$ within $(0,1]$. RMSprop uses decay $0.9$ and Adam uses $(\beta_1,\beta_2)=(0.9,0.95)$ with bias correction, both with $\epsilon=10^{-8}$ and no weight decay. Each risk is normalized by the same optimizer's independently evaluated, optimized $B=1$ risk at the same $D$. The CBS is the last tested batch in the passing prefix whose normalized risk is at most $1.2$, ending at the first tested failure even if later batches pass again. We refine the batch grid adaptively near this first crossing. SGD risks are computed by an exact second-moment recursion. For the normalized optimizers, learning-rate selection and final risk evaluation use independent random streams, and stratified Monte Carlo includes all $512$ coordinates. The dashed curves in the CBS plot in Section~\ref{subsec:synthetic} are ordinary least-squares fits in log--log coordinates over all displayed data budgets.

\paragraph{Batch-size experiment.}
\figref[right]{fig:story-overview} uses the diagonal Gaussian-gradient model with $D=32{,}000$, $(s,q,d)=(1,4,512)$, unit noise, and zero initialization. At every batch size, we run $N=D/B$ updates with power decay $p=4$ and retune the peak learning rate up to one. SignSGD risks are averaged over $512$ runs independent of tuning, while SGD risks are computed exactly. We take the empirical CBS as the last tested batch before the mean risk exceeds $1.2$ times its own $B=1$ value. This gives $2000$ for SignSGD and $16$ for SGD. Adding the SignSGD cap $\eta_{\peak}\leq B^{-1/2}$ gives $256$ versus $16$. These finite-grid results illustrate the batch-size advantage. The common-cap experiment does not directly test the theoretical CBS exponent.

\subsubsection{Additional results for Adam and RMSprop}
\label{app:adaptive-results}

\begin{figure}[!ht]
\centering
\captionsetup{skip=3pt}
\includegraphics[width=0.80\linewidth]{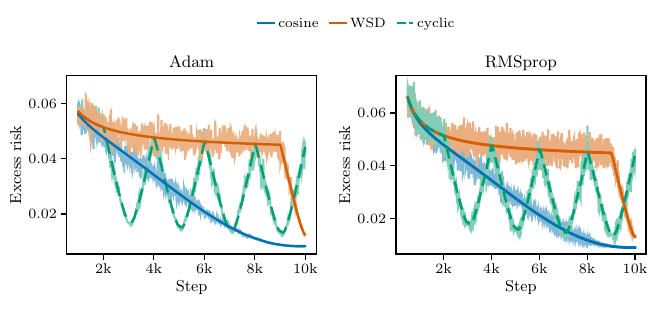}
\caption{\small \textbf{FSL predictions for Adam and RMSprop.} Risk dynamics for Adam and RMSprop, respectively. Light lines show the mean excess risk over 200 runs. Dark solid lines are FSL fits for cosine and WSD, and dashed lines predict the held-out cyclic schedule.}
\label{fig:adaptive-extension}
\label{fig:adaptive-extension-rmsprop}
\end{figure}

\begin{figure}[!ht]
\centering
\captionsetup{skip=3pt}
\includegraphics[width=0.80\linewidth]{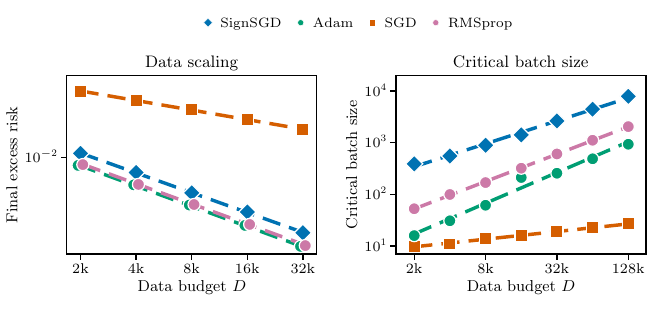}
\caption{\small \textbf{Data scaling and critical batch size.} \textbf{(Left)} Final excess risk versus data budget for SGD, SignSGD, RMSprop and Adam. \textbf{(Right)} Estimated CBS versus data budget for the same optimizers. Dashed lines in both panels denote power-law fits. The CBS uses a $20\%$ risk-increase tolerance relative to each optimizer's $B=1$ baseline.}
\label{fig:adaptive-scaling}
\end{figure}

\figref{fig:adaptive-extension} shows that the FSL captures risk dynamics for both Adam and RMSprop, which also exhibit data-scaling and batch-size advantages over SGD (\figref{fig:adaptive-scaling}). In \figref[right]{fig:adaptive-extension}, the held-out cyclic prediction has a root mean squared error of $1.90\times10^{-3}$ against the mean excess risk over 200 runs. On the same stored entries (zero-based indices $500,\ldots,9999$), the fitting errors are $1.60\times10^{-3}$ for cosine and $2.46\times10^{-3}$ for WSD.

\subsubsection{Structural-condition experiments}
\label{app:structural-conditions}
\figref{fig:structural-conditions} uses quadratic objectives with additive Gaussian gradient noise and the spectral and target-magnitude parameterization above. The target signs are fixed as in \hyperref[app:rotation-construction]{the rotation construction below}. We set $(s,q,d)=(0.3,4,512)$, initialize at zero, and use $B=1$ and $D=32{,}000$. All conditions have the same total gradient noise energy. Each optimizer and setting uses a separately tuned peak learning rate in $[10^{-5},1]$ and the schedule $\eta_k=\eta_{\peak}(1-k/N)^4$. Each setting and rotation path is evaluated on $64$ held-out stochastic runs. The left panel shows mean risk trajectories, with isotropic-noise controls (iso.) and one fixed rotated geometry (rot.).

For the middle panel, the Hessian remains diagonal and the noise covariance varies as
\[
\bSigma_\gamma=\frac{\operatorname{tr}(\bH)}{\operatorname{tr}(\bH^\gamma)}\bH^\gamma,\qquad \gamma\in\{0,0.1,\ldots,1\}.
\]
The endpoints give isotropic and curvature-aligned noise, respectively. The right panel uses the joint rotations described in \hyperref[app:rotation-construction]{the rotation construction below}, preserving the Hessian spectrum and target energy in each eigenmode. Each positive rotation strength averages three fixed rotation paths, with one shared unrotated control. Both panels report the ratio of the mean final SGD excess risk to the mean final SignSGD excess risk, so values above one favor SignSGD.

\paragraph{Rotation construction.}\leavevmode
\label{app:rotation-construction}
We use the spectral and target parameterization above, with $(s,q,d)=(0.3,4,512)$, $\lambda_j=j^{-q}$, and $|\theta_j^*|=j^{-\frac12(sq-q+1)}$. The target has independent random signs sampled once and then fixed. Following the matrix-exponential rotation construction of \citet{xie2025adamgeometry}, we draw a matrix $\bM$ with independent standard Gaussian entries and set
\[
\begin{aligned}
\bA&=\bM-\bM^\top,\qquad \bR_\rho=\exp(\rho\bA),\\
\bH_0&=\diag(\lambda_1,\ldots,\lambda_d),\qquad
\bH_\rho=\bR_\rho^\top\bH_0\bR_\rho,\qquad
\btheta_\rho^*=\bR_\rho^\top\btheta^*.
\end{aligned}
\]
No normalization is applied to $\bA$. We use three independent generator matrices with seeds $21161024$, $21161025$, and $21161026$, each held fixed across all $\rho$. A pilot experiment sets the six positive rotation strengths $\rho\in\{10^{-6},10^{-5},10^{-4},10^{-3},10^{-2},10^{-1}\}$. We also evaluate the exactly diagonal control $\rho=0$ using the same implementation and Gaussian seed protocol.

\paragraph{Training and evaluation.}\leavevmode
For each fixed $\rho$, we initialize $\btheta_0=\bzero$ and set $B=\sigma^2=1$. The gradient oracle and population excess risk are
\[
\begin{gathered}
\bg_k=\bH_\rho(\btheta_k-\btheta_\rho^*)+\bxi_k,\qquad
\bxi_k=\bR_\rho^\top\bH_0^{1/2}\bz_k,\qquad
\bz_k\overset{\mathrm{iid}}\sim\cN(\bzero,I),\\
\ell_k=\cE(\btheta_k)=\tfrac12(\btheta_k-\btheta_\rho^*)^\top\bH_\rho(\btheta_k-\btheta_\rho^*).
\end{gathered}
\]
Thus $\bxi_k\sim\cN(\bzero,\bH_\rho)$. This additive Gaussian oracle isolates the noise geometry used in the mechanism analysis, rather than simulating the full state-dependent noise of stochastic regression gradients.

\subsection{Language model experiments}
\label{app:llm}

\paragraph{Training configuration.}

We train a 0.1B-parameter LLaMA-style Transformer and track the validation loss throughout training. Unless otherwise specified, each run uses 10B tokens with sequence length $2048$, batch size $128$, peak learning rate $5\times10^{-4}$, and gradient clipping threshold $1.0$.

We train with SignSGD and Adam. For Adam, we set $(\beta_1, \beta_2) = (0.9, 0.95)$ and $\epsilon = 10^{-8}$. The schedule-comparison runs use decoupled weight decay with $\lambda=0.1$ for both optimizers~\citep{loshchilov2018decoupled}, giving the update
\[
\btheta_{k+1} = (1-\eta_k\lambda)\btheta_k - \eta_k \bu_k ,
\]
where $\bu_k$ is the optimizer-specific update at step $k$.

\paragraph{Learning rate schedules.}

\begin{figure}[!ht]
  \centering
  \captionsetup{skip=3pt}
  \includegraphics[width=0.50\linewidth]{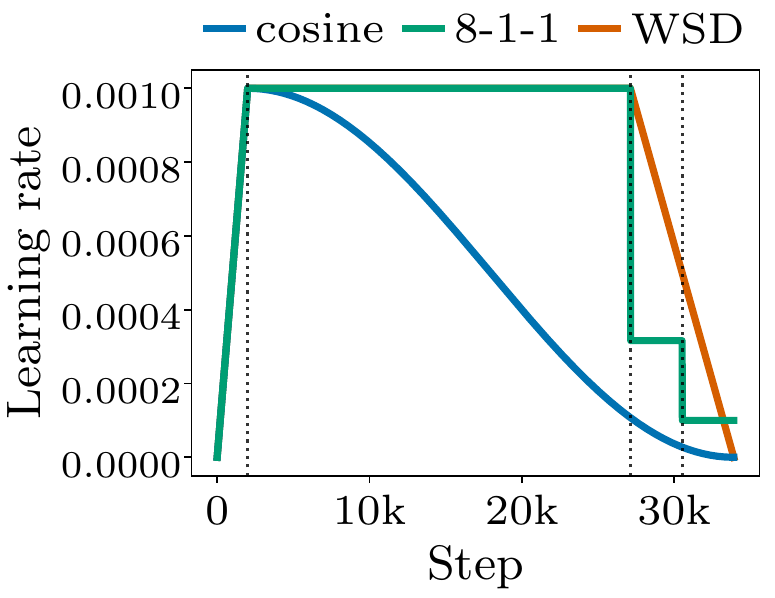}
  \caption{\small Learning rate schedules for language model training. All schedules use a warmup of $2000$ steps and the same peak learning rate.}
  \label{fig:lrs llm}
\end{figure}

\figref{fig:lrs llm} visualizes the learning rate schedules in our experiments: cosine~\citep{loshchilov2017sgdr}, WSD~\citep{hu2024minicpm}, and 8-1-1~\citep{bi2024deepseek}. All schedules incorporate a $2000$-step linear warmup from $0$ to the peak learning rate $\eta_{\peak}$. Following warmup, both WSD and 8-1-1 maintain a constant learning rate of $\eta_{\peak}$ until step $0.8N$ (where $N$ denotes the total training steps), after which WSD linearly decays to zero by the end of training, while the 8-1-1 schedule adopts a piecewise constant profile: dropping to $\eta_{\peak}/\sqrt{10}$ at step $0.8N$, and then to $\eta_{\peak}/10$ at step $0.9N$.

\paragraph{Details of FSL fitting.} We use the fitting procedure described in Appendix~\ref{app:fsl-detail} with $N_{\text{skip}}=2000$.

\paragraph{Loss curve fitting and prediction for different schedules.}

For \figref[a]{fig:llm-fit-predict}, we fit the FSL to the cosine and $8$-$1$-$1$ loss curves and predict the held-out WSD curve. In the fitting objective~\eqref{eq:fit-loss}, we use a 100-step moving average of the loss,
\[
\ell_k^{\mathrm{smooth}}=\frac{1}{2H}\sum_{j=k-H}^{k+H-1}\widetilde\ell_j,\qquad H=50,
\]
where $\widetilde\ell_j=\ell_{\min\{N,\max\{1,j\}\}}$ extends the first and last observed losses beyond the endpoints.

\paragraph{Joint fitting across batch sizes under a cosine schedule.}

We next consider a nonconstant learning rate.
\figref[left]{fig:llm-batch-additional} reports joint fitting
of Adam cosine-schedule runs across all three batch sizes
$B\in\{64,128,256\}$, with peak learning rate $5\times10^{-4}$.
The FSL closely tracks all three curves, indicating that the
same square-root batch-size clock remains effective in this
setting. Since all three curves are included in this fit,
this experiment tests joint batch-size consistency rather than
held-out batch-size extrapolation.

\paragraph{Extrapolation of batch sizes and data size.}

In \figref[b]{fig:llm-fit-predict}, we demonstrate that FSL can extrapolate to held-out batch sizes. We now test the simultaneous extrapolation of both batch size and data size. The batch-size comparisons at peak learning rate $5\times10^{-4}$ use the first 80\% of the 10B-token runs (about 8B tokens) for $B\in\{64,128,256\}$. SignSGD uses a constant rate throughout. The Adam curves cover the warmup and plateau of the WSD schedule up to the start of decay. We fit a single FSL model using only
\begin{itemize}[leftmargin=2em]
  \item the initial $50\%$ of the $B=64$ curve,
  \item the initial $50\%$ of the $B=128$ curve.
\end{itemize}
We then evaluate the predictive performance on the remaining
parts of the two curves and on the entire held-out $B=256$
curve. As shown in \figref[right]{fig:llm-batch-additional},
FSL predicts both the continuation of the fitted batch-size
curves and the fully held-out $B=256$ curve, with small
deviations over the longest continuation horizons.

\begin{figure}[!ht]
    \centering
\captionsetup{skip=3pt}
    \begin{minipage}[t]{0.44\linewidth}
      \centering
      \includegraphics[width=\linewidth]{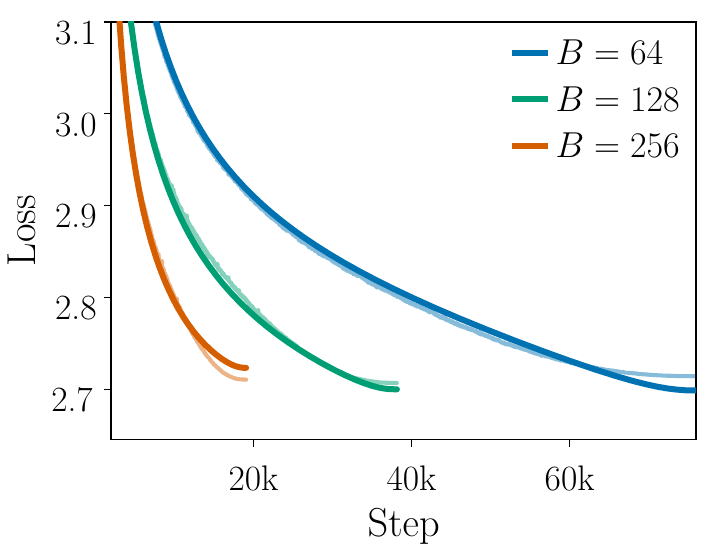}
    \end{minipage}%
    \hspace{0.02\linewidth}%
    \begin{minipage}[t]{0.44\linewidth}
      \centering
      \includegraphics[width=\linewidth]{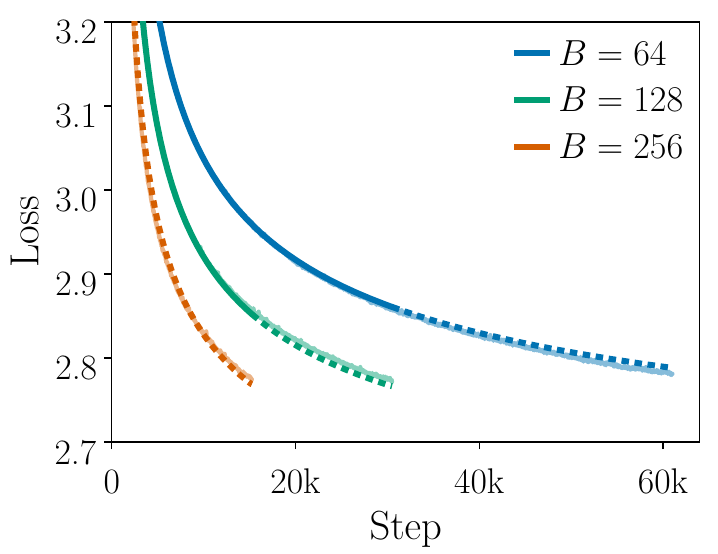}
    \end{minipage}
    \caption{\small
    \textbf{Joint batch-size fitting and extrapolation for Adam.}
    \textbf{(Left)} Under a cosine schedule with peak learning
    rate $5\times10^{-4}$, the parameters
    $\Theta$ are jointly fitted to all three loss
    curves for $B\in\{64,128,256\}$.
    \textbf{(Right)} Under a constant learning rate
    $\eta=5\times10^{-4}$, FSL is fitted using the first half
    of the $B=64$ and $B=128$ trajectories, then predicts
    their remaining parts and the entire held-out $B=256$
    trajectory.
    Solid portions denote FSL fits, and dashed portions denote
    predictions on held-out horizons.}
    \label{fig:llm-batch-additional}
\end{figure}

\paragraph{Robustness to a larger learning rate.}

\begin{figure}[!ht]
\centering
\captionsetup{skip=3pt}
\begin{minipage}[t]{0.44\textwidth}
  \centering
  \includegraphics[width=\linewidth]{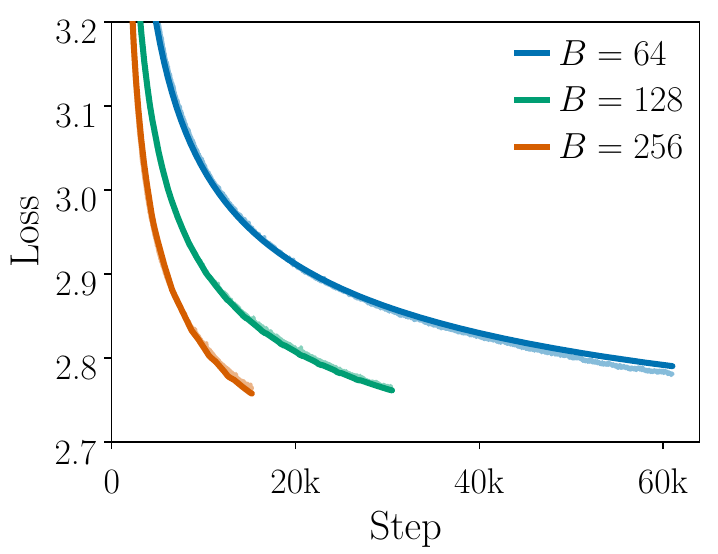}
\end{minipage}%
\hspace{0.02\linewidth}%
\begin{minipage}[t]{0.44\textwidth}
  \centering
  \includegraphics[width=\linewidth]{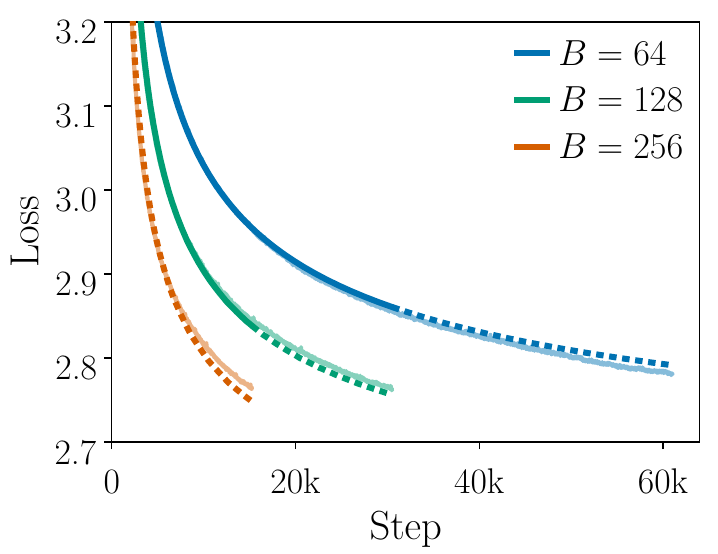}
\end{minipage}
\caption{\small\textbf{Fitting and extrapolation under a larger learning rate.} We repeat the Adam experiments in \figref[b]{fig:llm-fit-predict} and \figref[right]{fig:llm-batch-additional} with the learning rate increased from $5\times10^{-4}$ to $10^{-3}$. Solid portions denote FSL fitting results, and dashed portions represent FSL predictions on held-out horizons. \textbf{(Left)} Joint fitting across batch sizes. \textbf{(Right)} Fitting and extrapolation to held-out batch sizes and training horizons. FSL remains accurate for joint fitting and the held-out batch size, while the longer-horizon continuation predictions show modest deviations.}
\label{fig:peak lr 0.001 results}
\end{figure}

We extend the Adam experiments in \figref[b]{fig:llm-fit-predict} and \figref[right]{fig:llm-batch-additional} by increasing the learning rate from $5\times10^{-4}$ to $10^{-3}$. The model, data, batch sizes, and training horizon are kept unchanged. As shown in \figref{fig:peak lr 0.001 results}, FSL continues to capture the loss dynamics under this larger learning rate.

\end{document}